%% file: main.tex
\documentclass{article} %
\usepackage{iclr2025_conference,times}

\input{math_commands.tex}

\usepackage{graphicx}
\usepackage{colortbl}
\usepackage{caption}
\usepackage{subcaption}
\usepackage{textcomp}
\usepackage{gensymb}
\usepackage{xspace}
\usepackage{enumitem}
\usepackage{tabularx}
\usepackage{hhline}
\usepackage{booktabs}
\usepackage{wrapfig}
\usepackage{soul}
\usepackage{multirow}
\usepackage{ctable}
\usepackage{lipsum}
\usepackage{overpic}   %

\makeatletter
\DeclareRobustCommand\onedot{\futurelet\@let@token\@onedot}
\def\@onedot{\ifx\@let@token.\else.\null\fi\xspace}

\def\eg{\emph{e.g}\onedot} 
\def\ie{\emph{i.e}\onedot} 
\def\cf{\emph{c.f}\onedot}

\makeatother

\definecolor{tabgreen}{HTML}{81C784}
\definecolor{tabyellow}{HTML}{AED581}
\definecolor{tabyellowlight}{HTML}{DCE775}

\newcommand{\comparisonwidth}{0.112}
\newcommand{\doublecomparisonwidth}{0.224}

\definecolor{Gray}{gray}{0.95}
\newcolumntype{a}{>{\columncolor{Gray}}c}

\usepackage{hyperref}
\usepackage{cleveref}
\usepackage{url}

\newcommand\FramedBox[3]{%
  \setlength\fboxsep{0pt}
    \fboxrule=1pt%
  \fcolorbox{green}{white}{\parbox[t][#1][c]{#2}{\centering\tiny #3}}}

\title{CubeDiff: Repurposing Diffusion-Based Image Models for Panorama Generation}

\author{Nikolai Kalischek \\
ETH Z\"urich, Google\\
\And
Michael Oechsle \\
Google \\
\And
Fabian Manhardt \\
Google \\
\And
Philipp Henzler \\
Google \\
\And
Konrad Schindler \\
ETH Z\"urich\\
\And
Federico Tombari \\
Google \\
}

\iclrfinalcopy %
\begin{document}

\maketitle

\input{sections/abstract}
\input{sections/intro}
\input{sections/related}
\input{sections/method}
\input{sections/eval}

\input{sections/conclusion}

\bibliography{iclr2025_conference}
\bibliographystyle{iclr2025_conference}

\input{sections/appendix}

\end{document}

%% file: math_commands.tex
\usepackage{amsmath,amsfonts,bm}

\def\Figref#1{Figure~\ref{#1}}

\def\eqref#1{equation~\ref{#1}}

\def\1{\bm{1}}

\DeclareMathAlphabet{\mathsfit}{\encodingdefault}{\sfdefault}{m}{sl}
\SetMathAlphabet{\mathsfit}{bold}{\encodingdefault}{\sfdefault}{bx}{n}

%% file: sections/abstract.tex
\begin{abstract}
We introduce a novel method for generating 360° panoramas from text prompts or images. Our approach leverages recent advances in 3D generation by employing multi-view diffusion models to jointly synthesize the six faces of a cubemap. Unlike previous methods that rely on processing equirectangular projections or autoregressive generation, our method treats each face as a standard perspective image, simplifying the generation process and enabling the use of existing multi-view diffusion models. We demonstrate that these models can be adapted to produce high-quality cubemaps without requiring correspondence-aware attention layers. Our model allows for fine-grained text control, generates high resolution panorama images and generalizes well beyond its training set, whilst achieving state-of-the-art results, both qualitatively and quantitatively. Project page: \url{https://cubediff.github.io/}
\end{abstract}

%% file: sections/intro.tex
\begin{figure}[ht!]
    \centering
    \includegraphics[width=\linewidth]{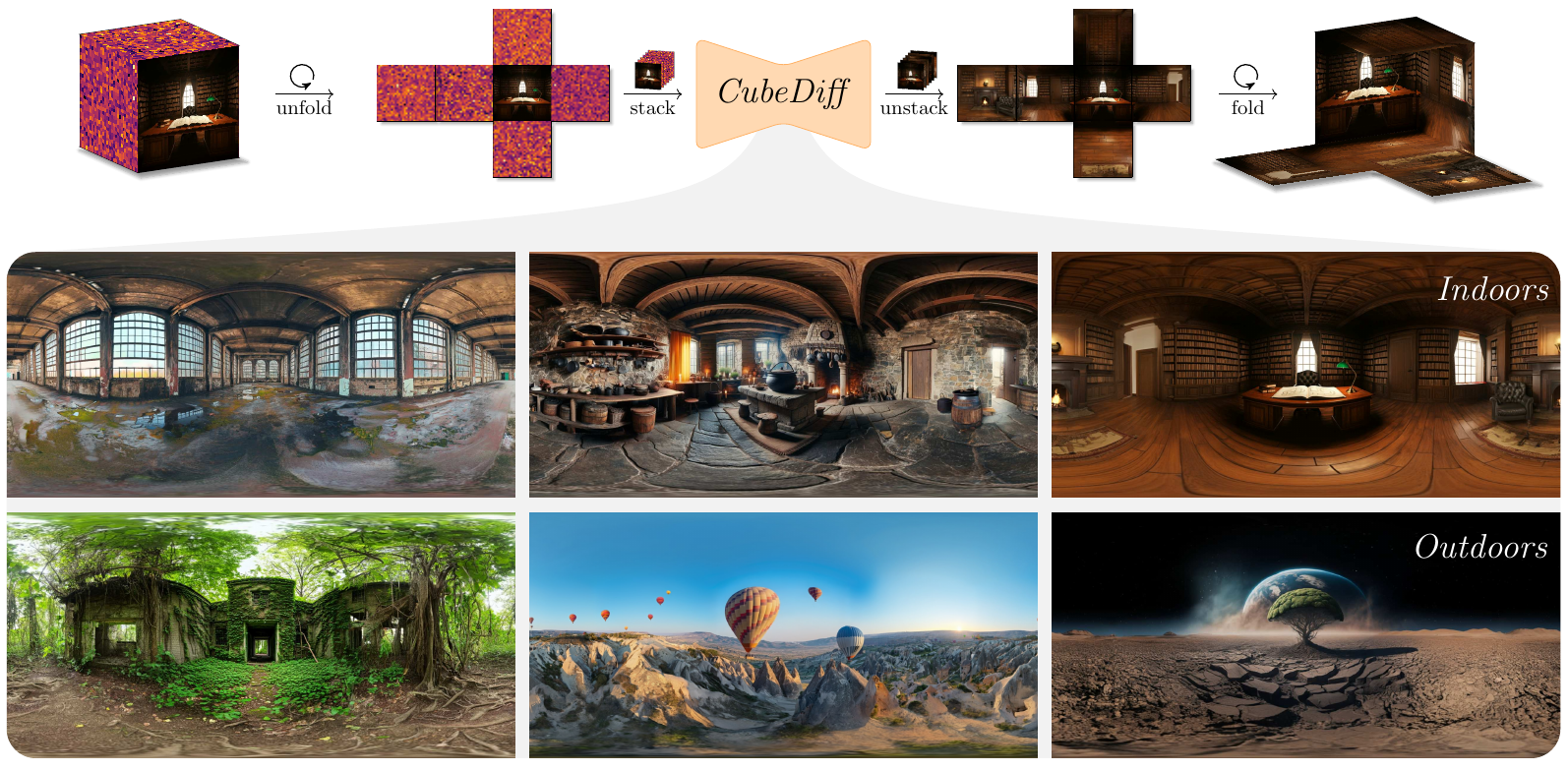}
    \caption{\emph{CubeDiff} leverages cubmaps to represent 360° panoramas and denoises all faces together in a single pass. In contrast to other works, \emph{Cubediff} does not need to consider distortions, since it operatkes on common 90° FOV perspective images, maing it possible to directly utilize the internet-scale image prior of the underlying diffusion model.}
    \label{fig:teaser}
\end{figure}
\vspace*{-1em}
\section{Introduction}

Recent advances in diffusion-based generative models have seen tremendous progress over the last two years, enabling a wide range of applications from artistic expression and product design to personalized content creation. Beyond generating realistic and diverse images based on text-to-image models \citep{Rombach2022Stable, saharia2022photorealistic}, these models are now capable of more complex tasks such as 3D asset creation (e.g., \citep{kalischek2022tetradiffusion, wang2024prolificdreamer, mohammad2022clip, poole2022dreamfusion}), estimating scene properties such as depth or semantics \citep{ke2024repurposing, baranchuk2021label}, illumination changes \citep{jin2024neural, zhao2024illuminerf, zeng2024dilightnet}, and generation of multi-view consistent images \citep{gao2024cat3d, Tang2023mvdiffusion}.

The latter is particularly interesting in virtual reality, gaming and entertainment, where 3D consistency is crucial for fully immersive experiences and thus user satisfaction. However, synthesizing high-quality, visually coherent panoramas presents unique challenges. First, capturing sufficient panoramic data is tedious and costly, as specialized cameras and/or additional processing are needed to remove stitching artifacts. Consequently, models must be trained in a low-data regime making them prone to overfitting, this limiting their generalization capabilities. Exemplary, a lot of models are restricted to indoor environments only~\citep{wu2023panodiffusion, song2023roomdreamer}. Second, panoramas must fulfill additional constraints compared to perspective images. Most notably, the image borders must align to allow a seamless wrap-around. But there are also more intricate, semantic constraints, e.g., the viewing frustum must cover the entire scene. Hence, when generating a panorama of a bedroom, it must contain \emph{exactly one} bed, \emph{at least one} door, etc. On the other hand, outdoor panoramas should maintain realistic spatial relationships between elements.

To satisfy those requirements, prior work had to introduce complex additional model components \citep{gao2024opa, Tang2023mvdiffusion, yang2024dreamspace}, or employ dedicated mechanisms such as autoregressive outpainting from a perspective view (causing artifacts like content drift and the Janus effect \citep{wang2023360}, and circular padding to enforce consistent wrap-around \citep{feng2023diffusion360, wu2023panodiffusion}. 

We introduce a simple yet highly effective solution: we generate panoramas using a fine-tuned multi-view diffusion model, following recent line of work \citep{gao2024cat3d, Tang2023mvdiffusion, zhang2023diffcollage}. This approach leverages the inherent properties of cubemaps, where a 360° $\times$ 180° panorama is represented by six perspective images on the faces of a cube. This allows us to fully recycle a pretrained text-to-image model, enabling generalization far beyond the limited training data. Contrary to existing methods, the architectural modifications we require to ensure consistency between cube faces are minimal: all attention layers are inflated by one additional dimension to enable crosstalk between the six faces. This simple modification, combined with fine-tuning on panorama data, achieves state-of-the-art results with significant visual and semantic coherence. Additionally, the model allows for fine-grained text control by training with face-specific image-text pairs, easily generated by prompting an LLM to produce per face text descriptions.

Our key insight is that existing, generative image models can be easily extended to generate high-resolution panoramas, by performing diffusion in cubemap space and adding attention mechanisms to other faces within the cubemap, see ~\Figref{fig:teaser}. The resulting model 
\begin{itemize}[noitemsep,topsep=0pt,leftmargin=4mm]
    \item enables consistent image generation across all cubemap faces and preserves the internet-scale image prior of the underlying diffusion model to generalize beyond the training panoramas;
    \item delivers state-of-the-art results on panorama generation, both qualitatively and quantitatively, and outperforms previous methods in terms of visual fidelity and coherence;
    \item enables efficient high-resolution synthesis, benefiting from current and future advances in off-the-shelf image diffusion models;
    \item allows for novel fine-grained text control, enabling users to guide the generation with detailed textual descriptions.
\end{itemize}

%% file: sections/related.tex
\section{Related Work}

Similar to 3D generative modelling, training data for panorama generation is scarce and much effort has been spent on how to repurpose standard perspective image priors for panoramas. The prevalent approach has been to autoregressively outpaint panoramas, more recently multi-view diffusion models have attracted interest. We now discuss relevant works and differences to our approach. 

\subsection{Panorama generation.}

Most panorama generators operate in equirectangular projection, thus having to deal with it severe nonlinear distortions (especially near the poles). Previous methods either autoregressively outpaint the panorama \citep{gao2024opa, lu2024autoregressive, wang2023360} or generate the entire equirectangular image in one shot \citep{feng2023diffusion360, wu2023panodiffusion}. They are commonly conditioned on either a single narrow field-of-view image \citep{akimoto2022diverse} or solely on a text prompt \citep{chen2022text2light}. The state of the art are diffusion models, which have gradually replaced adversarial approaches \citep{akimoto2022diverse, somanath2021hdr}. \cite{feng2023diffusion360} fine-tune a latent diffusion model on a panorama dataset and apply a circular blending strategy in the denoising and decoding stages to enforce consistent wrap-around. Similarly, \cite{wu2023panodiffusion} stitches the right part of the image to the left part in latent space in each denoising step. Such blending improves the results, but encumbers the inference step. In our method it is not required. \cite{lu2024autoregressive} propose to autoregressively outpaint a panorama with a complex architecture of submodules for panorama-aware visual guidance, NFoV guidance and panorama-aware geometric guidance. In \cite{wang2023360}, the authors extend the outpainting task to ingest multiple NFoV images of the same scene. A two-stage network predicts their relative rotations, then a diffusion model with ControlNet~\citep{zhang2023adding} outpaints the panorama based on the projected inputs. Recently \citep{voynov2023curved} introduce a diffusion model with control over the rendering geometry, including panoramic outputs. \cite{gao2024opa} additionally incorporate a state space model to aggregate global information into cross-attention layers of the diffusion model, building up the panorama by inpainting empty regions. The present work demonstrates that, with the right representation, high-quality panoramas can be obtained without inflating the complexity and brittleness of the architecture. Related to panorama generation is the more modest strategy to alter existing panoramas by injecting a user-defined style, in either equirectangular or cubemap projection \citep{yang2024dreamspace, song2023roomdreamer}.

\subsection{Multi-view diffusion}

Multi-view diffusion models offer a compelling alternative to equirectangular or autoregressive panorama generators. \cite{zhang2023diffcollage} introduces a compositional diffusion scheme that enables the generation of large-scale content, leveraging models trained on smaller constituent parts. That work is based on factor graphs, and demonstrates how the cubemap can be turned into a factor graph in order to train a diffusion model conditioned on segmentation maps. The work most closely related to ours is \cite{Tang2023mvdiffusion}. It aims to generate cylindrical panoramas (i.e., 360\degree horizontal field of view, but restricted vertical view angle). The authors propose a sophisticated correspondence-aware cross-attention between local neighborhoods of eight perspective feature maps spaced at 45\degree angles. Recently, \cite{gao2024cat3d} and \cite{shi2023mvdream} discovered that expanded attention layers that connect not only features within an image but also across multiple images, are beneficial when handling multiple object-centric views. Our approach turns this setup inside-out and extends a pretrained text-to-image (T2I) model in a similar manner for panorama generation. We instead do not require camera pose or 3D information, due to the fixed viewing geometry of the cubemap.

%% file: sections/method.tex
\section{Panorama Representations}

Panoramic images aim to capture a complete 360° $\times$ 180° view of a scene from a fixed view point. There exist several different panorama representations in literature, each with its own advantages and drawbacks. This section briefly discusses the most prominent ones. 

\paragraph{Spherical projection} maps a 360° view onto a sphere, preserving the geometric relationships between points in the scene. Points are generally defined using longitude and latitude. While conceptually intuitive, directly utilizing a spherical representation for image processing is challenging due to difficulties in representing a sphere on a flat image plane, which often leads to distortions and non-uniform sampling densities in practical implementations.

\paragraph{Equirectangular projection} projects the spherical panorama onto a 2D rectangle. To this end, latitude and longitude coordinates on the sphere are mapped to vertical and horizontal coordinates on a rectangle. While widely used due to its simplicity, equirectangular projection suffers from significant distortions, especially near the poles where horizontal stretching becomes extreme. This distortion affects both visual quality and the performance of algorithms processing equirectangular panoramas, as most existing T2I models process images with NFoV images. %

\paragraph{Cubemaps} offer an alternative representation where a 360° view is projected onto the six faces of a cube. Each face captures a 90°
field of view, resulting in six perspective images that can be seamlessly stitched together. This representation avoids the polar distortions inherent to equirectangular projections, providing more uniform sampling, making it highly applicable to existing diffusion models trained on vast amount of perspective images. However, note that cubemaps introduce discontinuities at the edges of the cube faces, which needs to be handled carefully.

\section{Method}

\begin{figure}
    \centering
    \includegraphics[width=\linewidth]{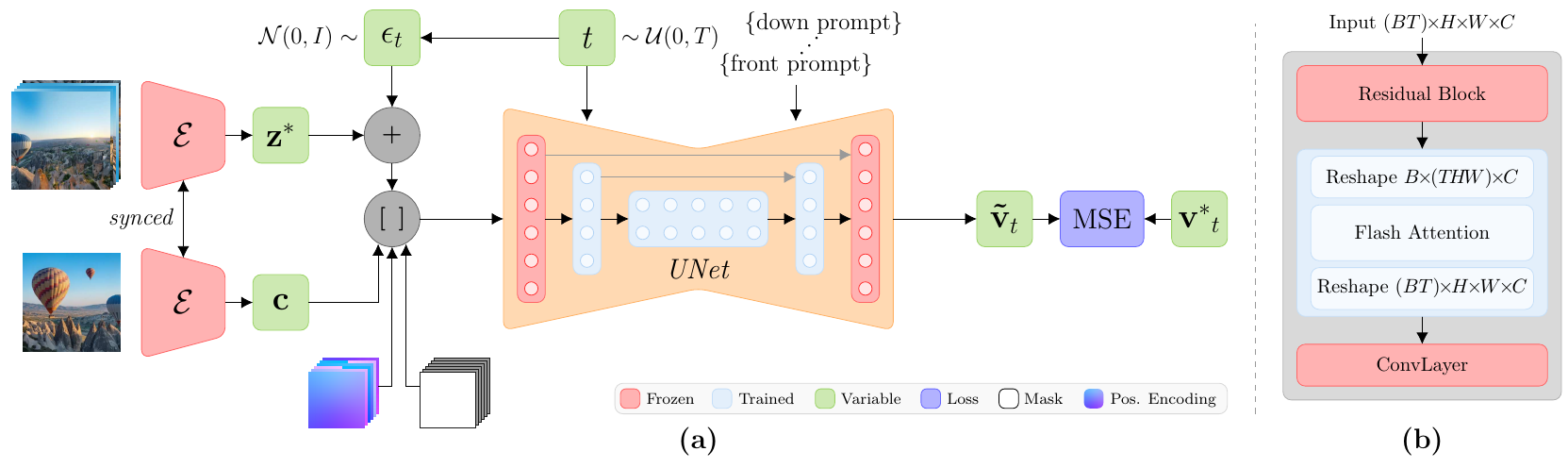}
     \caption{ \textbf{An overview of our training pipeline and panorama model.} \textbf{(a)} We project all training panoramas onto a cubmap and feed the faces to our frozen VAE encoder with synchronized GroupNorm to obtain the respective latents and enrich them with panorama-specific positional encodings for explicit spatial awareness. \textbf{(b)} We only train the inflated attention layers to be cross-frame aware. \vspace{-0.5cm}
     }\label{fig:method}
\end{figure}

We introduce \emph{CubeDiff}, a novel approach for generating high-quality, consistent panoramas using the cubemap representation. %
\emph{CubeDiff} generates the six perspective views of a cubemap in parallel and context-aware manner, exploiting the strengths of pretrained T2I diffusion models. 
Below, we delve into the architectural choices that enable \emph{CubeDiff} to achieve high-quality and consistent panoramas, while retaining strong generalization capabilities inherited from the pretrained model. Similar to \cite{gao2024cat3d}, \emph{CubeDiff} comprises a variational autoencoder (VAE) and a latent diffusion model (LDM), mirroring the structure of conventional T2I diffusion models. However, we carefully adapt each component for effective multi-view panorama generation. 

\subsection{Model architecture} 
The latents produced by the VAE are used to fine-tune a pretrained LDM operating on a 128x128x8 latent space, initialized with weights from a model trained on a large-scale image dataset.
The pretrained LDM consists of an architecture similar to Stable Diffusion~\citep{Rombach2022Stable}, which is build with multiple convolutional, self-attention, and cross-attention layers.
To enable cross-view awareness and maintain global consistency, we inflate all existing 2D attention layers, \ie both self-attention and cross-attention for text conditioning. These layers, adapted from \citep{shi2023mvdream}, extend the attention mechanism across all six cube faces, allowing the model to learn relationships and dependencies between different viewpoints. Inflating of layers can be easily conducted by extending the token sequence length from $b\times (hw)\times l$ to $b\times(thw)\times l$, \eg for self-attention, where $b$ is the batch size, $hw$ the flattened spatial size and $t=6$ the cube length. While this is different to more sophisticated attention layers \citep{Tang2023mvdiffusion, huang2024epidiff}, it in turn enables us to retain the original pretrained attention weights, which reduces the risk of overfitting and thus greatly improves overall performance.

The LDM receives two conditioning signals. We incorporate text embeddings, either one common prompt or one prompt for each face, and a single conditional view of the scene (w.l.o.g. we assume the front face of the cube). During training, we concatenate the VAE latents of the conditioning views to the noisy latents of the target views, providing the LDM with complete context information. Furthermore, we incorporate a binary mask channel into the latent representations. This mask distinguishes between conditioning views (provided as clean latents) and target views (subjected to noise injection during training). We show an overview of our model architecture in \Figref{fig:method}.

\subsection{Synchronized GroupNorm}

Our VAE architecture incorporates synchronized group normalization, a crucial element for achieving consistent color tones across the generated panorama. Since our VAE processes the six faces of a cubemap as a batch of six individual images, standard group normalization can lead to subtle color inconsistencies among different views (\cf \cref{fig:syncedgn:a}). This occurs as feature statistics are computed and normalized independently for each image in the batch. Without synchronization, encoding and decoding a panorama results in noticeable shifts, particularly evident in the equirectangular projection. Synchronized group normalization addresses this issue by jointly normalizing feature activations across both spatial and inter-view dimensions. Consequently, synchronized group normalization contributes significantly to the generation of visually harmonious and coherent panoramas. Similar effects have been observed in~\citep{he2023scalecrafter}. We further discuss this in \Cref{sec:ablations} and compare synchronized and unsynchronized results in \Cref{fig:syncedgn:a}.

\subsection{Positional encoding}
To provide the LDM with explicit spatial awareness within the cubemap, we augment the latent representations with positional encodings derived from the 3D geometry of the cube. For each point on a cube face, we compute its corresponding UV coordinates on the unit cube, defined by:
\begin{equation}
    u = \arctan2(x, z) \quad,\quad v = \arctan2(y, \sqrt{x^2 + z^2}),
\end{equation}
where $(x, y, z)$ are the 3D coordinates of the point on the cube face, projected onto the unit cube. These UV coordinates are then normalized to $[0, 1]$ and concatenated as two additional channels to the (noisy) latents. This positional encoding scheme provides the model with information about the spatial location of each latent patch within its respective cube face, facilitating the generation of panoramas with consistent geometry and object relationships across views.

\subsection{Overlapping predictions}
To further enhance the geometric and color consistency across cube faces, we introduce overlapping predictions during both training and generation. Instead of generating each face with a 90° field of view (FoV), we enlarge the FoV by 2.5° on each side, resulting in an effective FoV of 95° per face. This means each generated face includes a small overlap with its neighboring ones. This overlapping generation strategy serves two purposes. During training, it encourages the model to learn consistent representations across adjacent faces, as the overlapping regions provide additional context and constraints. During panorama assembly, we discard these overlapping regions and only retain the central 90° portion of each generated face. This strategy effectively avoids the need for explicit blending operations at the cube face boundaries, which can sometimes introduce subtle artifacts. The overlaps can be seen at the boundaries of the cubemaps in \Cref{fig:conditionalresults} (\eg, the duplicated fireplace in the right and back views) and in the appendix.

\subsection{Classifier-free guidance}

We employ classifier-free guidance (CFG) \citep{ho2022classifier} on both the text and image conditions during training. Thereby, we randomly drop either the text prompt, the conditional image, or both. When the text prompt is dropped, it is replaced with null tokens in the cross-attention layers; when the conditional image is dropped, its corresponding tokens in the self-attention layers are masked out by setting them to negative infinity, effectively zeroing out their attention weights. 
This training procedure enables diverse panorama generation scenarios during inference. Users can provide both text and image conditions for maximum control and fidelity or drop both or either condition to explore unconditional generation modes.

%% file: sections/eval.tex
\section{Experiments}
This section details our experimental setup, followed by quantitative and qualitative evaluations. We compare the performance of \emph{~CubeDiff} against the state-of-the-art
and ablate our design choices.

\subsection{Evaluation protocol}

\subsubsection{Training and inference setup}

We finetune our model using Adam~\citep{Kingma2014Adam} and train for 30,000 iterations with batch size 64. The learning rate is ramped up to $8\times 10^{-5}$ in the first 10,000 steps. During training, we employ classifier-free guidance, dropping conditional signals 10\% of the time. We find it important to not only drop the text condition in the cross-attention layers but to also zero out the input condition in the self-attention layers. The diffusion model is finetuned using v-prediction~\citep{salimans2022progressive}. 
We employ DDIM sampling~\citep{song2020denoising} with 50 steps during inference.

\subsubsection{Datasets}

\paragraph{Training.} We train on a mixture of indoor and outdoor environments by combining multiple publicly available sources, including Polyhaven~\citep{polyhaven}, Humus~\citep{humus}, Structured3D~\citep{zheng2020structured3d} and Pano360~\cite{kocabas2021spec}, giving in total around 48000 panoramas for training. While Humus provides an explicit cubemap representations, all other datasets come with equirectangular panoramas. We thus first generate cubemaps from these panoramas using standard perspective projection, ensuring consistent overlap between adjacent faces. 
To further enable text-guided panorama generation, we infer textual descriptions for each panorama in the datasets using the publicly available Gemini model \citep{team2023gemini}. We explore two captioning strategies: (1) generating a single caption for the entire panorama by providing Gemini with all six cube faces as input and (2) generating individual captions for each face independently, enabling fine-grained text control. 

\paragraph{Testing.} We evaluate our method on the common Laval Indoor~\citep{gardner2017learning} and Sun360~\citep{xiao2012sun360} datasets. Laval Indoor consists of over 2100 high quality panorama captures of various indoor environments, Sun360 encompasses around 1000 panoramas including both -- indoor and outdoor scenes.
Note that we use those datasets only for evaluation, while Diffusion360 also uses Sun360 for training and OmniDreamer even leverages both datasets to train their models. Nonetheless, we decided to use these datasets for the sake of fairness and due to the lack of any proper overlapping test datasets.

\subsubsection{Metrics}

We use various metrics and modalities for evaluation -- including perceptual metrics, text alignment, and a user study.

\paragraph{Perceptual Metrics.} We use the very common Fréchet Inception Distance (FID) \citep{heusel2017gans} metric to measure the similarity between the distribution of real and generated images in a feature space derived from a pretrained Inception network. Lower FID scores indicate greater similarity and, thus, higher image realism; We additionally report the CLIP-FID \citep{kynkaanniemi2022role} metric, replacing the Inception network with CLIP \citep{radford2021learning} to leverage its semantic understanding capabilities through a joint image-text embedding space. This metric captures thus both -- visual fidelity and text-image alignment; Finally, we employ the kernel inception distance (KID)\citep{binkowski2018demystifying}. Similar to FID, KID uses features from a pre-trained network, however, it quantifies the difference between real and generated data distributions using the maximum mean discrepancy rather than the Fréchet distance.

\paragraph{Text Alignment.} To measure text alignment we refer to the common CLIP score~\citep{hessel2021clipscore} . The CLIP score computes the cosine similarity within the shared text-image embedding to measure the agreement between generated panoramas and their corresponding text prompts. Hence, a higher CLIP score indicates stronger semantic agreement between image and text.

\subsubsection{Competitors}
We compare \emph{CubeDiff} to various state-of-the-art panorama generation methods.
As for plain text to panorama generation, we employ Text2Light~\citep{chen2022text2light} and PanFusion~\citep{zhang2024taming} to serve as our main competitors. 
For single image conditioning, we respectively use OmniDreamer~\citep{lu2024autoregressive} and PanoDiffusion~\cite{wu2023panodiffusion} as representatives for autoregressive and direct panorama generation based approaches.
Finally, we compare against Diffusion360~\citep{feng2023diffusion360} and MVDiffusion~\citep{Tang2023mvdiffusion} for text and image conditioning based methods. Note that while Diffusion360 directly outputs panorama images, MVDiffusion instead employs multi-view diffusion models with a custom cross attention mechanism. Overall, the choice of baselines represents a variety of different generation techniques, covering various different  tasks. Please note that none of the existing methods besides MVDiffusion offers the possibility to condition specific parts of the panorama on individual text prompts.

\subsection{Qualitative evaluation}
In this section, we provide a qualitative evaluation of our method. We first present several conditional image generations of our method, before comparing \emph{CubeDiff} against the state-of-the-art. 

\subsubsection{Conditional image generation.}
In \Figref{fig:conditionalresults}, we show generated panoramas given text-image pairs as condition. We considered input conditions that cover a broad range of scenes, such as outdoor and indoor scenes, bright and dark settings as well as texture rich and uniformly colored areas. Note that we do not show the text conditions due to limited space, however, we provide them in the appendix. We see that our approach yields high quality results under these diverse input settings. We especially emphasize the level of detail and geometric consistency beyond the input image.
\input{figures/conditionalresults}

\subsubsection{Qualitative comparison.}
For visual comparison against the state-of-the-art, we show generated panoramas and their respective perspective projections in \Figref{fig:comparison}. To this end, we sample random image and text pairs from the LAVAL Indoor dataset. We further group the methods according to their input modalities. Compared to the text-only approach Text2Light, our method is able to produce much more complex panoramas with better details and visual appeal. As for image-only approaches, we see that \emph{CubeDiff} is capable of producing the most realistic panoramas. In particular, while OmniDreamer suffers from blurry regions, PanoDiffusion is not able to properly transfer the input image appearance across the whole panorama. Finally, also for text and image conditioning our method again produces the best results, especially in terms of geometry. For example, while MVDiffusion is indeed capable of generating high quality images, the method sometimes produces inaccurate geometries as, for example, some walls and hand rails exhibit bending artifacts after perspective projection. Similarly, Diffusion360 occasionally suffers from implausible indoor layouts. To summarize, despite of using different input modalities, \emph{CubeDiff} always generates high quality panoramas, surpassing all other state-of-the-art works in terms of visual appeal and geometric consistency.
\input{figures/qualitativecomparison}

\subsection{Quantitative evaluation}

\begin{wrapfigure}{R}{0.375\textwidth}
\vspace{-1em}
    \includegraphics[width=0.66\linewidth]{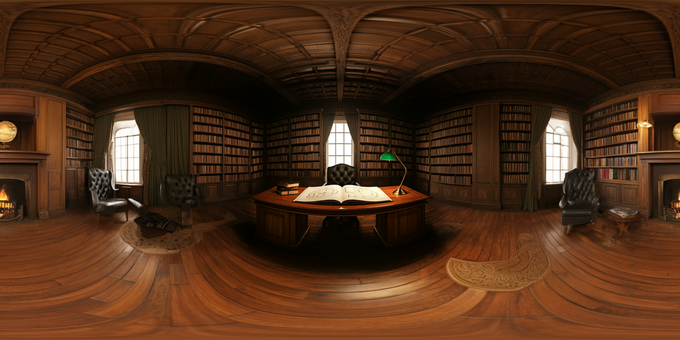}%
    \hfill
    \includegraphics[width=0.33\linewidth]{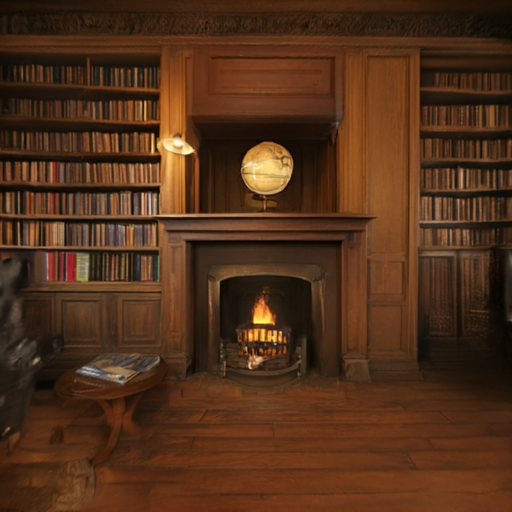}
    \includegraphics[width=0.66\linewidth]{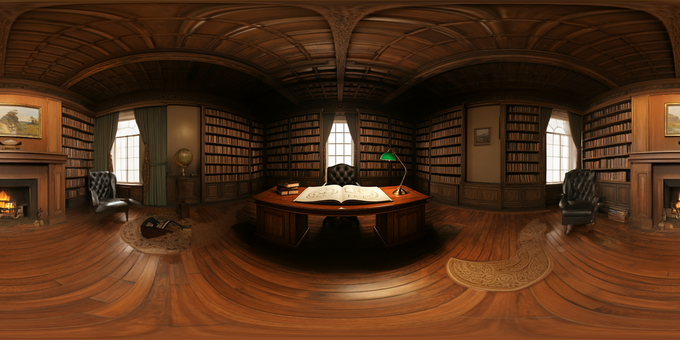}%
    \hfill
    \includegraphics[width=0.33\linewidth]{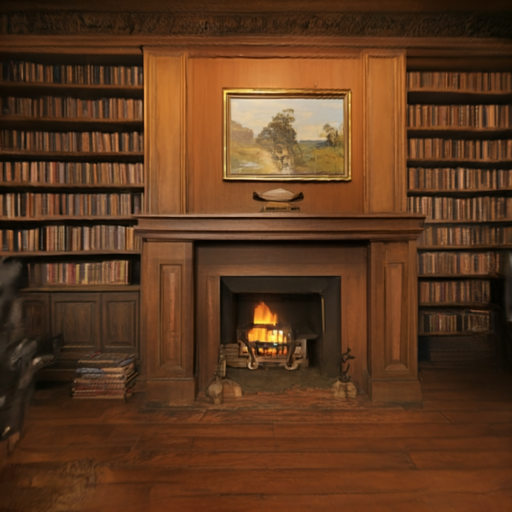}
    \includegraphics[width=0.66\linewidth]{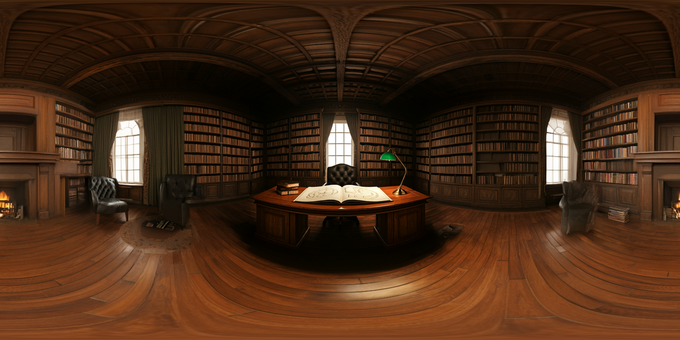}%
    \hfill
    \includegraphics[width=0.33\linewidth]{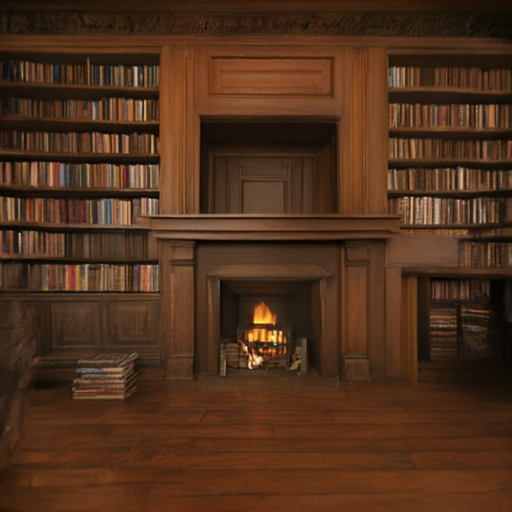}
    \includegraphics[width=0.66\linewidth]{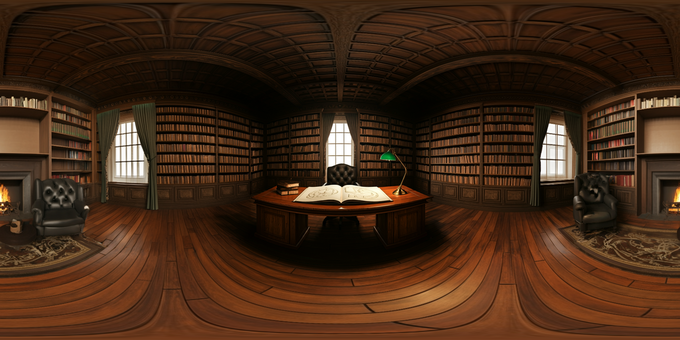}%
    \hfill
    \includegraphics[width=0.33\linewidth]{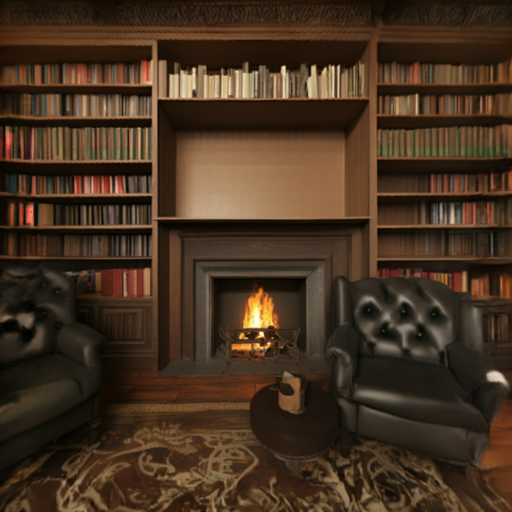}
    
    \captionof{figure}{\textbf{Fine-grained Text Control.} We show an example for fine-grained-text control of the back face. Our model is able to change details following the provided prompt. First, we add a golden globe above the fireplace; second, we place a picture above the fireplace; third, we leave the space above empty; last, we instead add a bookshelf above it.}
    \vspace{-2em}
    \label{fig:ablationtext}
\end{wrapfigure}
In this section, we provide the results of our quantitative evaluation on the Laval Indoor and the SUN360 dataset. 
We evaluate all methods on perceptual quality and consistency.

In \Cref{tab:perceptualeval} we provide quantitative results for visual quality. Our method outperforms all competitors significantly, regardless of input modalities. For example, we can report a FID score of 9.47 on Laval Indoor, which is a 270\% relative improvement compared to the second best performing method MVDIffusion, reporting a score of 25.7. Compared to works that use only image or text as input conditioning, the gap even widens with Text2Light and PanoDiffusion respectively reporting a FID of 28.3 and 58.6. This trend holds across all metrics. Interestingly, \emph{CubeDiff} performs similarly across different input modalities, demonstrating its strong generalizibility. 

However, the provided perceptual metrics can only evaluate the overall realism of the generated panoramas and are not capable of capturing consistency towards input. We next study the alignment to the input text prompt. To this end, we leverage the CLIP score to measure how well the generated panoramas align with the text input. As can be seen in the table our method surpasses the state-of-the-art again by a significant amount for all datasets and modalities, showing how precisely our model respects the textual input.

\begin{table}[!t]
\centering
\small
\setlength{\tabcolsep}{3pt} %
\renewcommand{\arraystretch}{1.1}

\resizebox{0.95\linewidth}{!}{
\begin{tabular}{lcccccccccccc}
        \specialrule{0.75pt}{0pt}{1pt}
        & \multicolumn{5}{c}{\textbf{LAVAL Indoor}} & \multicolumn{5}{c}{\textbf{SUN360}} \\
        \hhline{~----------}
        & FID $\downarrow$ & KID ($\times 10^2$)$\downarrow$ & Clip-FID $\downarrow$ & FAED $\downarrow$ & CS $\uparrow$ & FID$\downarrow$  & KID ($\times 10^2$)$\downarrow$  & Clip-FID$\downarrow$ & FAED $\downarrow$ & CS $\uparrow$ \\
        \hline
        Text2Light & 28.3 & 1.45 & \cellcolor{tabyellow}11.5 & 136.1 & 25.18 & 60.1 & 4.31 & 31.3 & 82.9 & 23.27\\
        PanFusion & 41.7 & 2.85 & 19.8 & 71.7 & 26.58 & \cellcolor{tabyellow}30.0 & \cellcolor{tabyellow}1.42 & \cellcolor{tabyellow}\phantom{0}7.8 & 44.5 & 25.28\\
        \hline
        OmniDreamer & 71.0 & 5.17 & 23.9 & \cellcolor{tabyellow}19.2& - & 92.3 & 8.89 & 51.7 & 30.4 & -\\
        PanoDiffusion & 58.6 & 4.08 & 26.6 & 106.8 & - & 52.9 & 3.51 & 28.9 & 98.0 & -\\
        \cellcolor{Gray}\textbf{Ours\textsubscript{img}} & \cellcolor{Gray}11.7 & \cellcolor{Gray}0.47 & \cellcolor{Gray}4.4 & \cellcolor{Gray}22.0 & \cellcolor{Gray}- & \cellcolor{Gray}27.4 & \cellcolor{Gray}1.35 & \cellcolor{Gray}11.5 & \cellcolor{Gray}\phantom{0}8.9 & \cellcolor{Gray}-\\
        \hline
        Diffusion360 & 33.1 & 2.07 & 16.9 & 23.7 & 26.38 & 45.4 & 3.73 & 18.5 & \cellcolor{tabyellow}12.6 & 22.89 \\
        \cellcolor{Gray}\textbf{Ours\textsubscript{img+txt}} & \cellcolor{tabgreen}\textbf{\phantom{0}9.5} & \cellcolor{tabgreen}\textbf{0.32} & \cellcolor{tabgreen}\textbf{\phantom{0}3.2} & \cellcolor{tabgreen}\textbf{18.4} & \cellcolor{Gray} 27.02 & \cellcolor{Gray}25.5 & \cellcolor{tabgreen}\textbf{1.33} & \cellcolor{Gray}\phantom{0}8.1 & \cellcolor{Gray}\phantom{0}7.6 & \cellcolor{Gray}25.00\\
        \hline
        MVDiffusion & \cellcolor{tabyellow}25.7 & \cellcolor{tabyellow}1.11 & 13.5 & - &\cellcolor{tabyellow} 27.44 & 50.9 & 3.71 & 15.4  & 32.3 & \cellcolor{tabyellow} 25.54\\
        \cellcolor{Gray}\textbf{Ours\textsubscript{img+multitxt}} & \cellcolor{Gray}10.0 & \cellcolor{Gray}0.35 & \cellcolor{Gray}\phantom{0}4.1 & \cellcolor{Gray}21.2 & \cellcolor{tabgreen} \textbf{30.17} & \cellcolor{tabgreen}\textbf{24.1} & \cellcolor{tabgreen}\textbf{1.33} & \cellcolor{tabgreen}\textbf{\phantom{0}7.0} & \cellcolor{tabgreen}\textbf{\phantom{0}5.7} & \cellcolor{tabgreen} \textbf{28.14} \\
\specialrule{0.75pt}{1pt}{1pt}
\specialrule{0.75pt}{1pt}{1pt}
\end{tabular}
}
\caption{\textbf{Quantitative Evaluation on the Laval Indoor and SUN360 dataset.} We provide a comparison to various competitors and different input modalities. The first block of rows are text-only methods, the second image-only, the third image and single text description and the last block are image and multi-caption methods. \emph{CubeDiff} provides the best perceptual quality having the best scores across all methods. Moreover, we find that the performance of \emph{CubeDiff} remains similar among different input modalities.}
\label{tab:perceptualeval}
\vspace{-1em}
\end{table}

\subsection{User study}
We conducted a user study with a two-alternative forced choice (2AFC) survey to evaluate our panorama generation method. Each of the 28 participants was shown 30 pairs of generated panoramas alongside the original conditioning image and asked to select their preferred option based on quality, composition, style, and alignment with the condition image.

Our method outperformed competitors statistically ($p < 0.1$, binomial test). Specifically, $16.9\%$, $17.3\%$, and $19.5\%$ of participants preferred our single-image, multi-image, and no-text variants, respectively. The no-text variant nearly matched the ground truth preference ($19.9\%$), demonstrating our method’s ability to generate realistic and accurate panoramas. In contrast, OmniDreamer, PanoDiffusion, MVDiffusion, and Diffusion360 had significantly lower preference rates of $1.7\%$, $5.3\%$, $7.0\%$, and $12.3\%$, respectively.

\subsection{Fine-grained Text Control}
Different to all competitors, our method enables complete fine-grained and per-face text control. For example in \Figref{fig:ablationtext}, we show results for providing different text descriptions for the back face. We can always generate visually appealing results, regardless of the object we place above the fireplace.

\subsection{Ablations}\label{sec:ablations}

\begin{figure}[t]
\centering
    \begin{subfigure}[b]{1.0\textwidth}
        \centering
        \includegraphics[width=\linewidth]{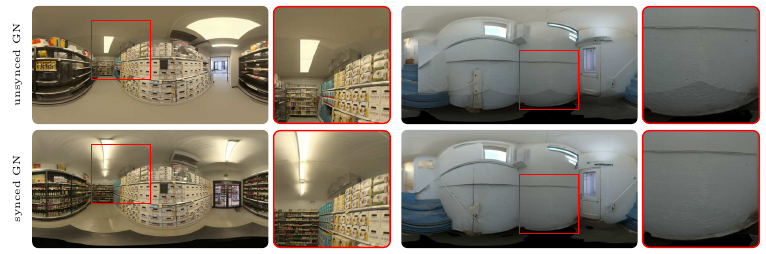} 
        \subcaption{}
        \label{fig:syncedgn:a}
    \end{subfigure}
    \begin{subfigure}[b]{1.0\textwidth}
        \centering
        \includegraphics[width=\linewidth]{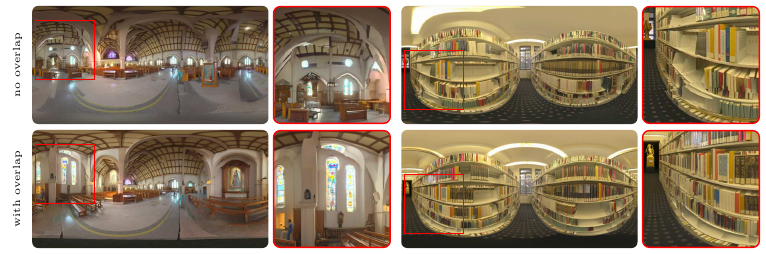} 
        \subcaption{}
        \label{fig:syncedgn:b}
    \end{subfigure}
    \caption{\textbf{Ablation on synchronized GN and overlap prediction.} \textbf{(a)} Top: Group normalization over the spatial dimension only. Bottom: Additional normalization over the frame dimension. \textbf{(b)} Top: Panoramas without overlapping cube faces. Bottom: Panoramas with our standard 2.5° overlap. Please zoom in to observe the differences.}
    \label{fig:syncedgn}
    \vspace{-1em}
\end{figure}

\paragraph{Synchronized Group Norm (GN)}
Synchronized GN ensures consistency across cube faces by normalizing over both spatial and frame dimensions, as shown in \Cref{fig:syncedgn:a}. Without it, models often exhibit color inconsistencies and artifacts at cube face boundaries. While metrics like FAED may not capture these subtle issues, synchronized GN significantly improves visual quality.

\paragraph{Overlapping Prediction}
Overlapping predictions mitigate discontinuities at cube face boundaries by introducing small overlaps, as illustrated in \Cref{fig:syncedgn:b}. This ensures seamless transitions, with non-overlapping regions cropped for the final panorama. The approach leverages global context from full attention, eliminating visible seams without additional VAE finetuning.

%% file: figures/conditionalresults.tex
\fboxsep=0mm%
\fboxrule=2pt%

\begin{figure}[t!]
    \centering
    \begin{minipage}[c]{0.497\linewidth}
        \centering
        \fcolorbox{green}{green}{\includegraphics[width=0.162\linewidth, height=0.162\linewidth]{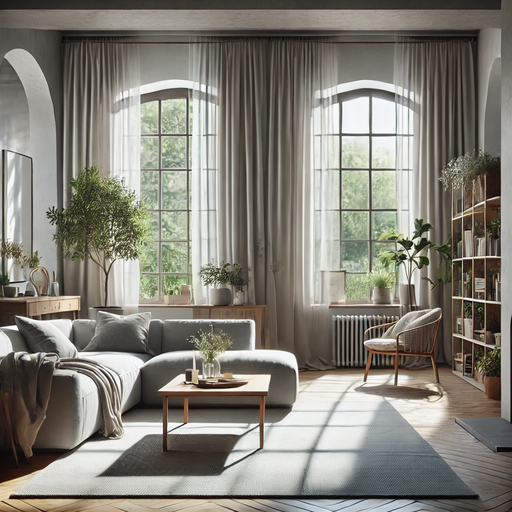}}%
        \hfill
        \includegraphics[width=0.162\linewidth, height=0.162\linewidth]{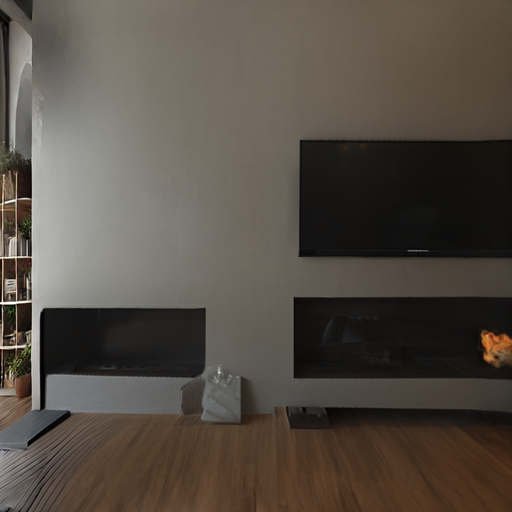}%
        \hfill
        \includegraphics[width=0.162\linewidth, height=0.162\linewidth]{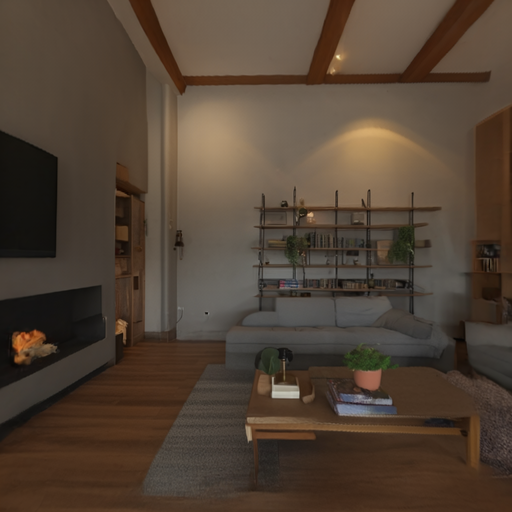}%
        \hfill
        \includegraphics[width=0.162\linewidth, height=0.162\linewidth]{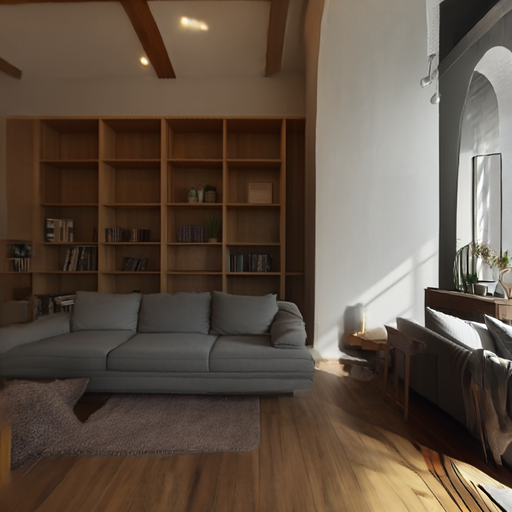}%
        \hfill
        \includegraphics[width=0.162\linewidth, height=0.162\linewidth]{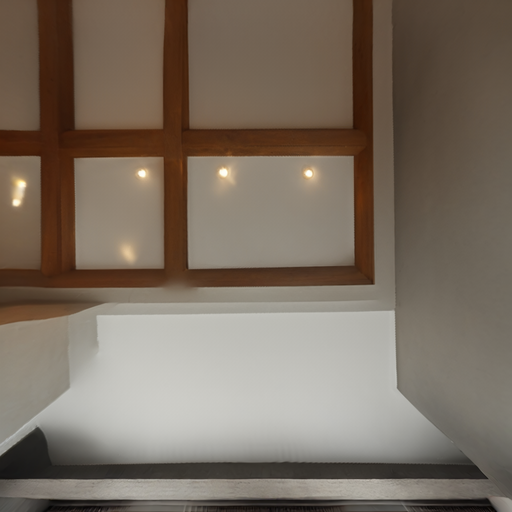}%
        \hfill
        \includegraphics[width=0.162\linewidth, height=0.162\linewidth]{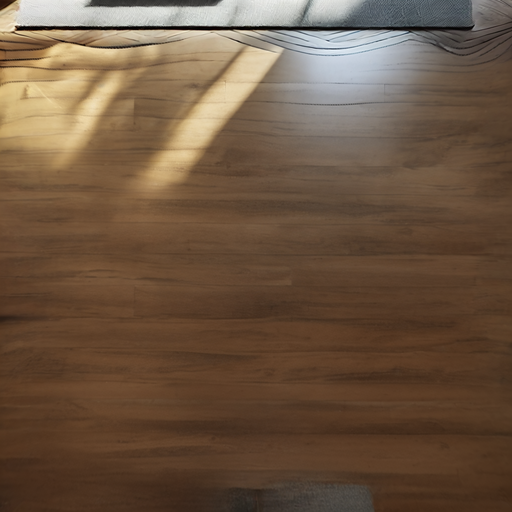}%
    \end{minipage}
    \hfill
    \begin{minipage}[c]{0.497\linewidth}
        \centering
        \fcolorbox{green}{green}{\includegraphics[width=0.162\linewidth, height=0.162\linewidth]{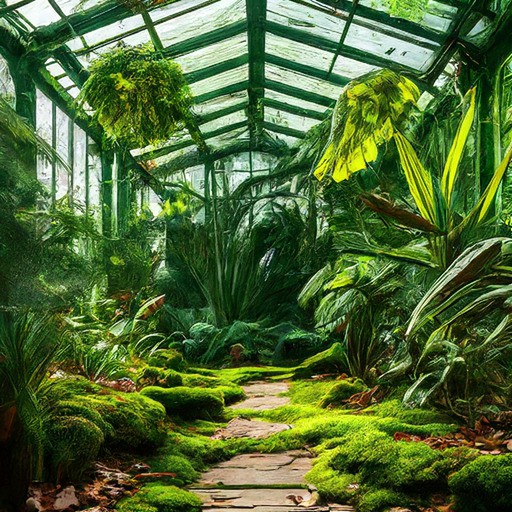}}%
        \hfill
        \includegraphics[width=0.162\linewidth, height=0.162\linewidth]{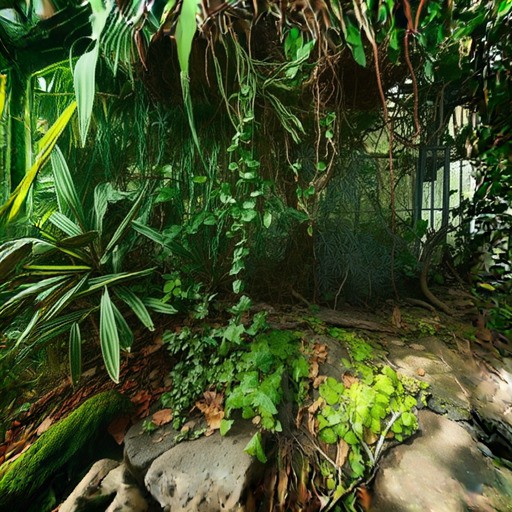}%
        \hfill
        \includegraphics[width=0.162\linewidth, height=0.162\linewidth]{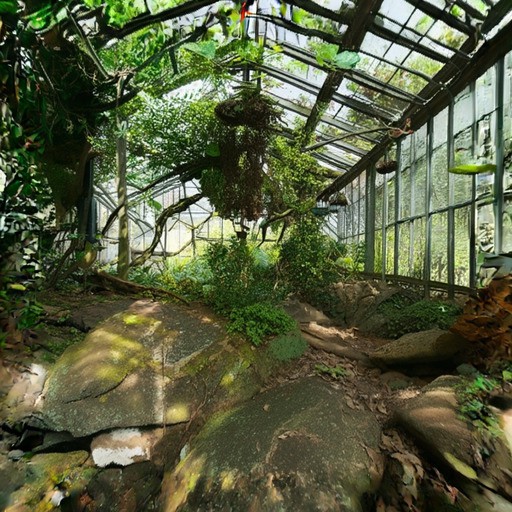}%
        \hfill
        \includegraphics[width=0.162\linewidth, height=0.162\linewidth]{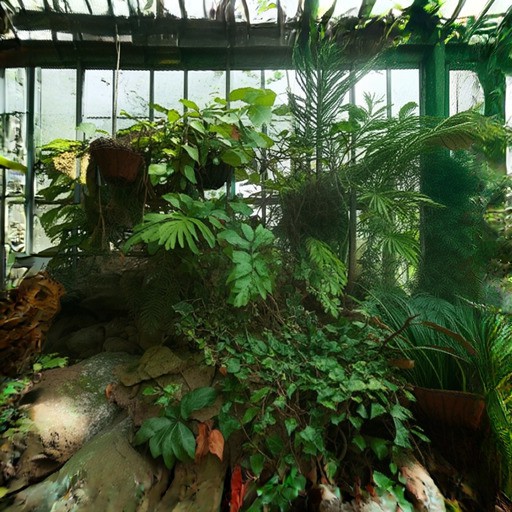}%
        \hfill
        \includegraphics[width=0.162\linewidth, height=0.162\linewidth]{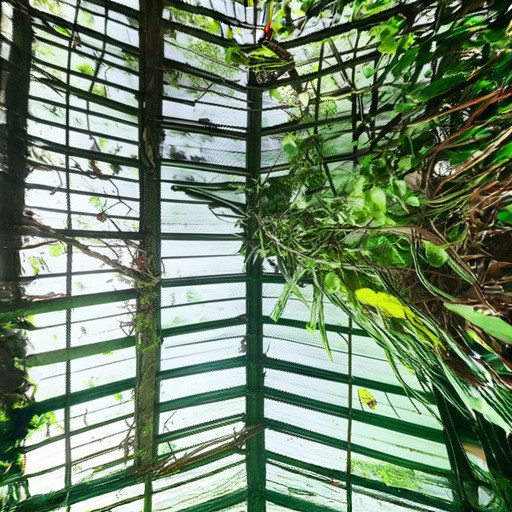}%
        \hfill
        \includegraphics[width=0.162\linewidth, height=0.162\linewidth]{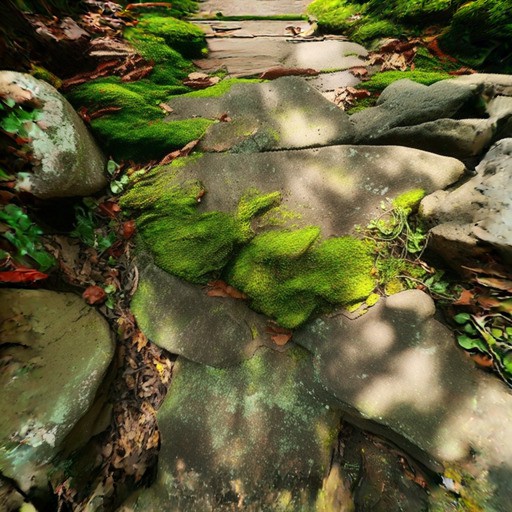}%
    \end{minipage}
    \begin{minipage}[c]{0.497\linewidth}
        \centering
        \includegraphics[width=\linewidth, height=0.5\linewidth]{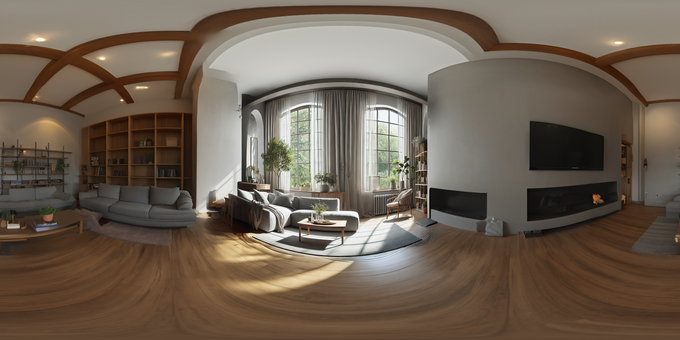}
    \end{minipage}
    \hfill
    \begin{minipage}[c]{0.497\linewidth}
        \centering
        \includegraphics[width=\linewidth, height=0.5\linewidth]{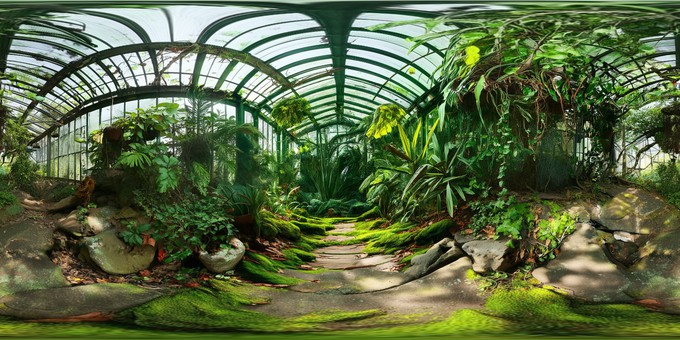}
    \end{minipage}

    \vspace{1.5mm}
    \begin{minipage}[c]{0.497\linewidth}
        \centering
        \fcolorbox{green}{green}{\includegraphics[width=0.162\linewidth, height=0.162\linewidth]{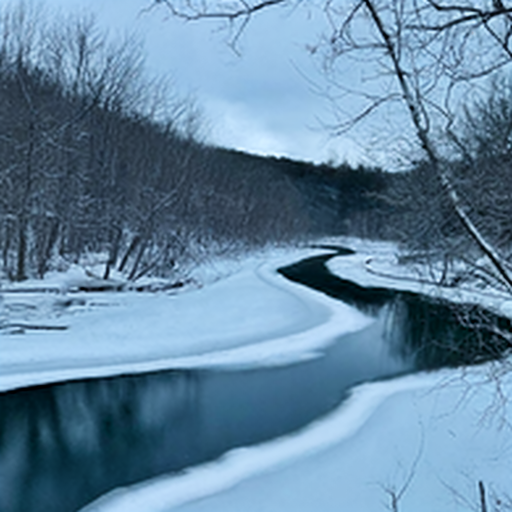}}%
        \hfill
        \includegraphics[width=0.162\linewidth, height=0.162\linewidth]{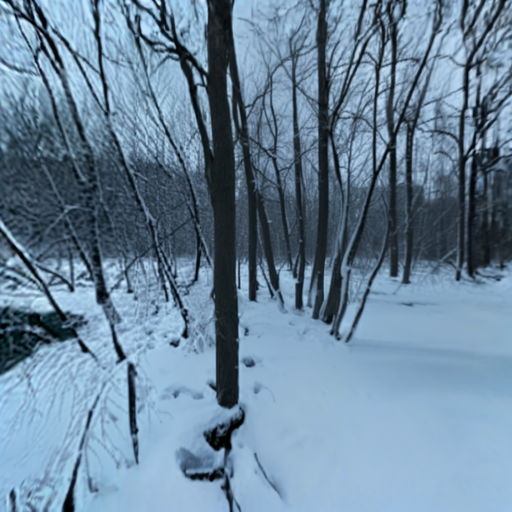}%
        \hfill
        \includegraphics[width=0.162\linewidth, height=0.162\linewidth]{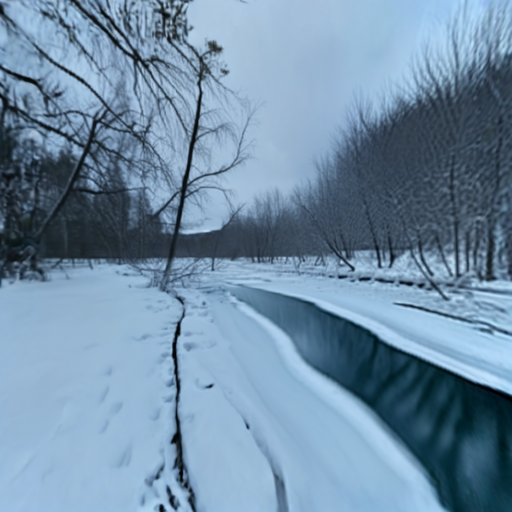}%
        \hfill
        \includegraphics[width=0.162\linewidth, height=0.162\linewidth]{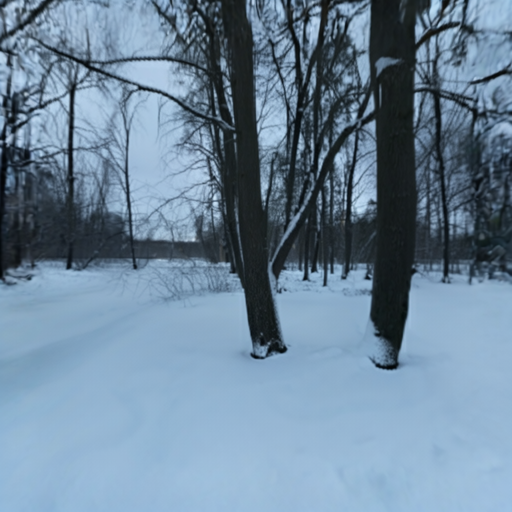}%
        \hfill
        \includegraphics[width=0.162\linewidth, height=0.162\linewidth]{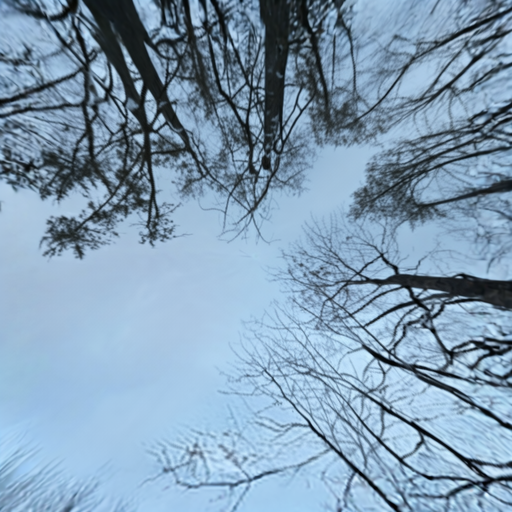}%
        \hfill
        \includegraphics[width=0.162\linewidth, height=0.162\linewidth]{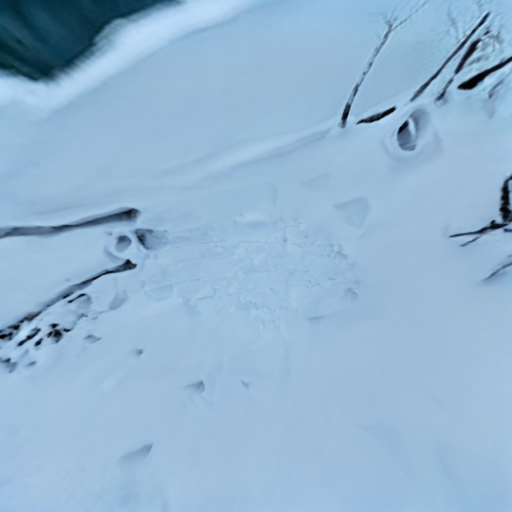}%
    \end{minipage}
    \hfill
    \begin{minipage}[c]{0.497\linewidth}
        \centering
        \fcolorbox{green}{green}{\includegraphics[width=0.162\linewidth, height=0.162\linewidth]{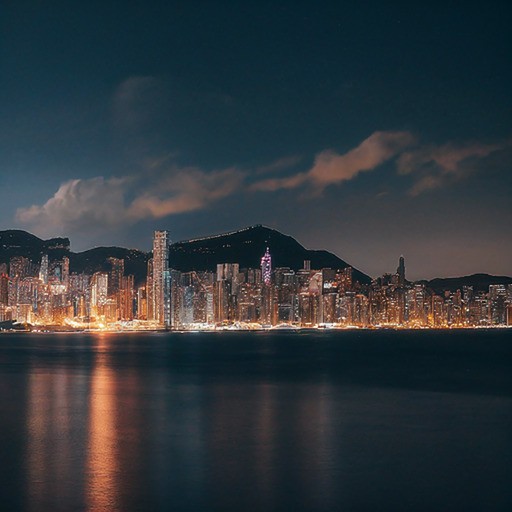}}%
        \hfill
        \includegraphics[width=0.162\linewidth, height=0.162\linewidth]{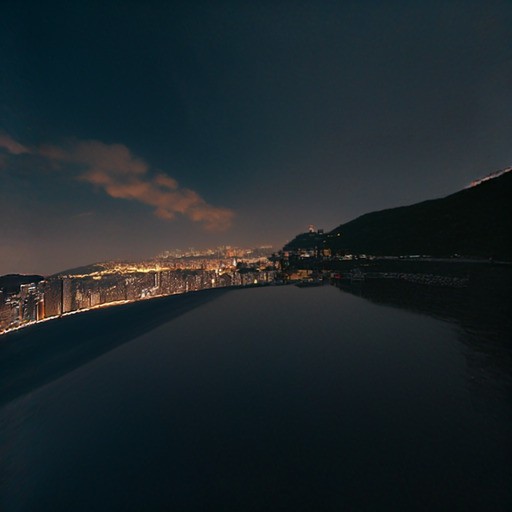}%
        \hfill
        \includegraphics[width=0.162\linewidth, height=0.162\linewidth]{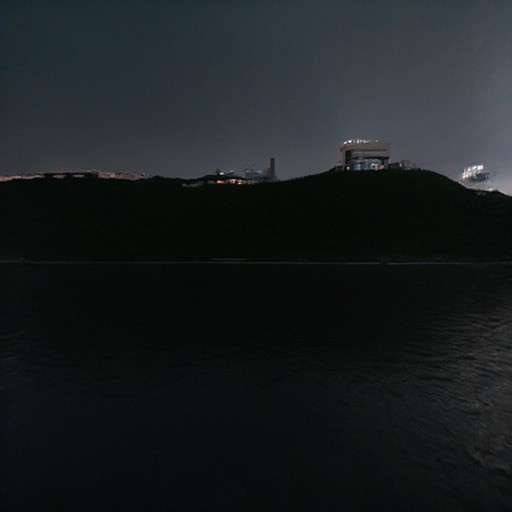}%
        \hfill
        \includegraphics[width=0.162\linewidth, height=0.162\linewidth]{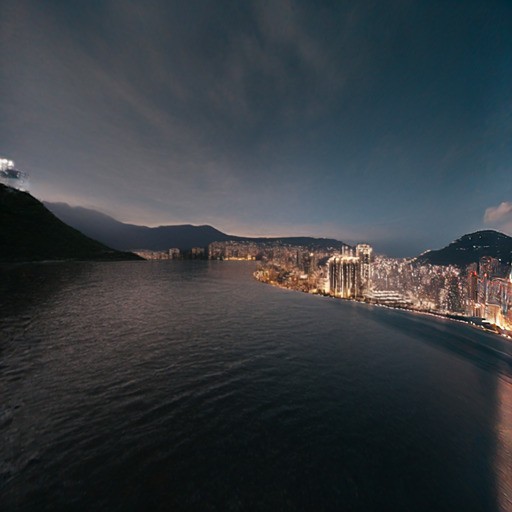}%
        \hfill
        \includegraphics[width=0.162\linewidth, height=0.162\linewidth]{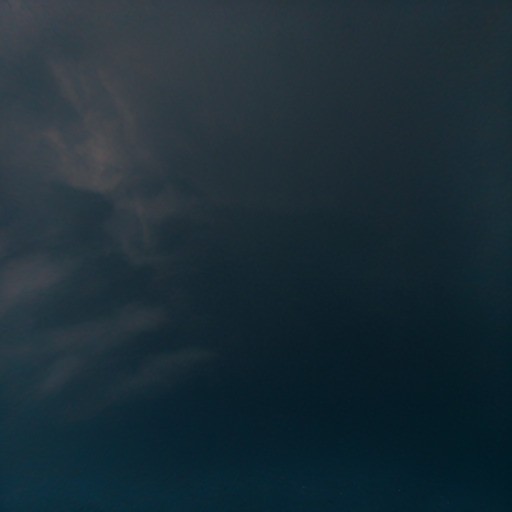}%
        \hfill
        \includegraphics[width=0.162\linewidth, height=0.162\linewidth]{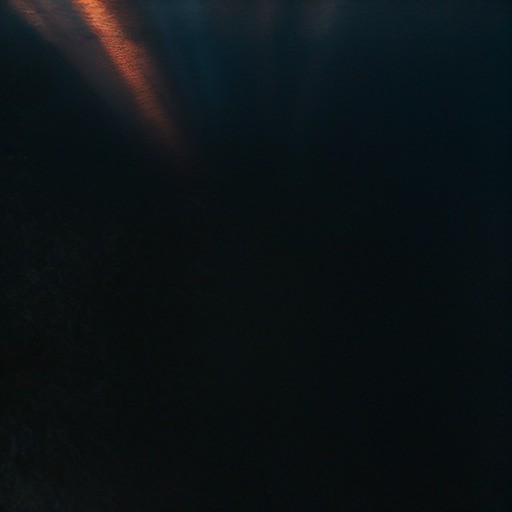}%
    \end{minipage}
    \begin{minipage}[c]{0.497\linewidth}
        \centering
        \includegraphics[width=\linewidth, height=0.5\linewidth]{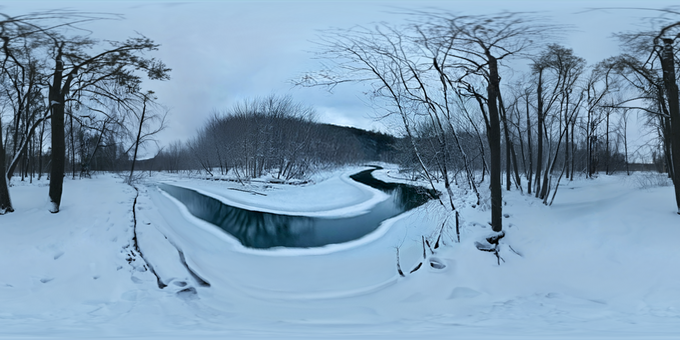}
    \end{minipage}
    \hfill
    \begin{minipage}[c]{0.497\linewidth}
        \centering
        \includegraphics[width=\linewidth, height=0.5\linewidth]{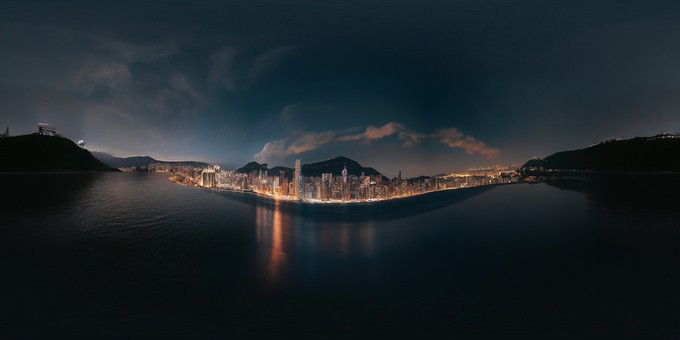}
    \end{minipage}

    \caption{\textbf{Cubemaps and panoramas generated by \emph{CubeDiff} with image and text condition.} We depict a diverse set of generated panoramas including indoor, outdoors, bright and dark scenes. In all settings, \emph{CubeDiff} produces high quality and realistic panoramas that align with the input image.}
    \vspace{-0.5cm}
    \label{fig:conditionalresults}
\end{figure}

%% file: figures/qualitativecomparison.tex
\begin{figure}[!t]
    \centering
    \begin{minipage}[c][0.05\linewidth]{0.015\linewidth}
        \centering
        \tiny
        \rotatebox{90}{Text-only}
        \vspace{-18mm}
    \end{minipage}
    \begin{minipage}[c][0.1\linewidth]{0.03\linewidth}
        \centering
        \tiny
        \rotatebox{90}{Text2Light}
    \end{minipage}
    \begin{minipage}[c]{\doublecomparisonwidth\linewidth}
        \centering
    \includegraphics[width=\linewidth, height=0.5\linewidth]{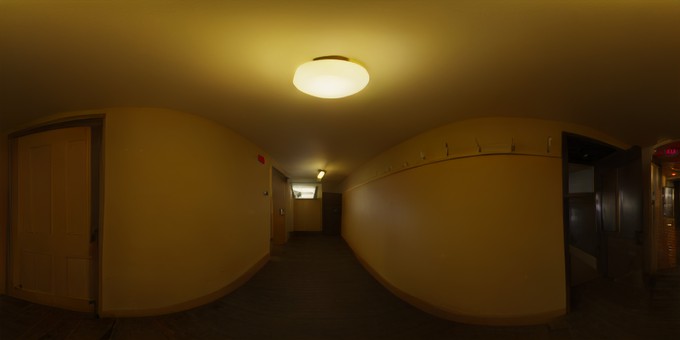}
    \end{minipage}
    \begin{minipage}[c]{\comparisonwidth\linewidth}
        \centering
    \includegraphics[width=\linewidth, height=\linewidth]{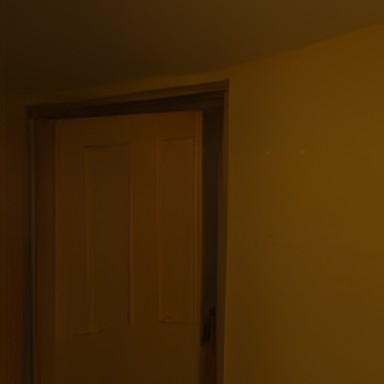}
    \end{minipage}
    \begin{minipage}[c]{\comparisonwidth\linewidth}
        \centering
    \includegraphics[width=\linewidth, height=\linewidth]{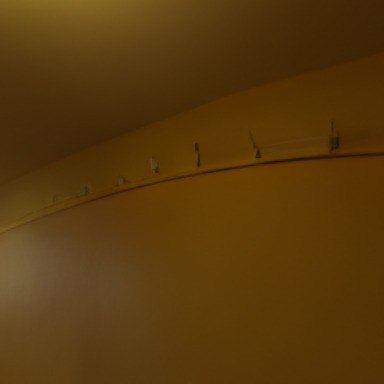}
    \end{minipage}
    \vspace{1mm}
    \begin{minipage}[c]{\doublecomparisonwidth\linewidth}
        \centering
    \includegraphics[width=\linewidth, height=0.5\linewidth]{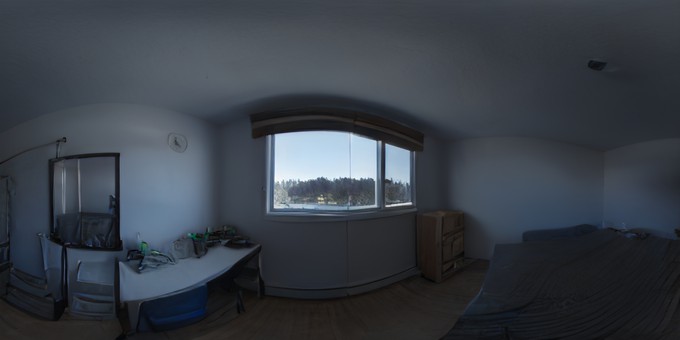}
    \end{minipage}
    \begin{minipage}[c]{\comparisonwidth\linewidth}
        \centering
    \includegraphics[width=\linewidth, height=\linewidth]{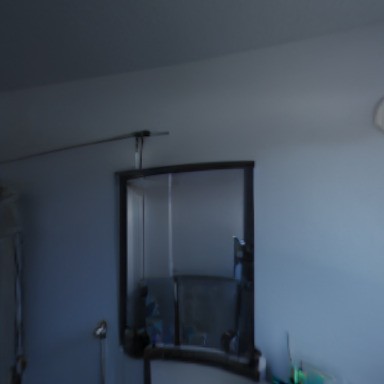}
    \end{minipage}
    \begin{minipage}[c]{\comparisonwidth\linewidth}
        \centering
    \includegraphics[width=\linewidth, height=\linewidth]{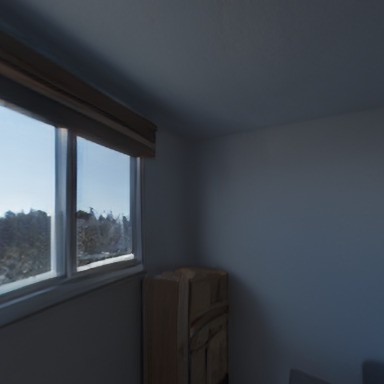}
    \end{minipage}
    \begin{minipage}[c][0.05\linewidth]{0.015\linewidth}
        \centering
        \tiny
        \phantom{\rotatebox{90}{Text-only}}
    \end{minipage}
    \begin{minipage}[c][0.1\linewidth]{0.03\linewidth}
        \centering
        \tiny
        \rotatebox{90}{PanFusion}
    \end{minipage}
    \begin{minipage}[c]{\doublecomparisonwidth\linewidth}
        \centering
    \includegraphics[width=\linewidth, height=0.5\linewidth]{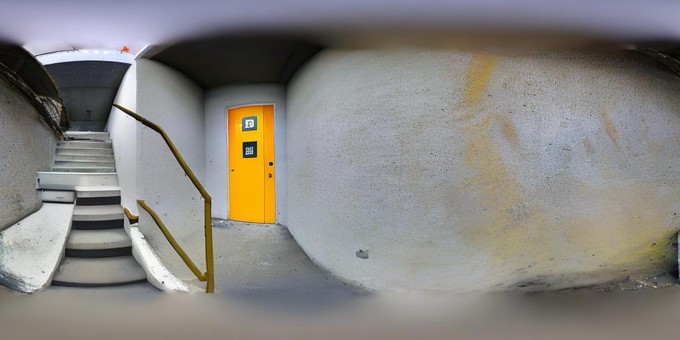}
    \end{minipage}
    \begin{minipage}[c]{\comparisonwidth\linewidth}
        \centering
    \includegraphics[width=\linewidth, height=\linewidth]{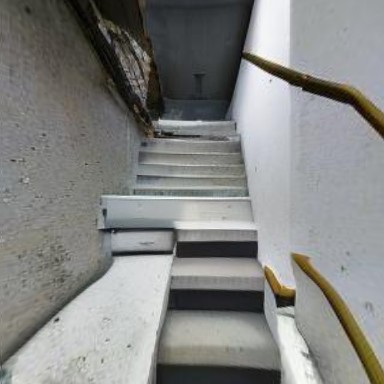}
    \end{minipage}
    \begin{minipage}[c]{\comparisonwidth\linewidth}
        \centering
    \includegraphics[width=\linewidth, height=\linewidth]{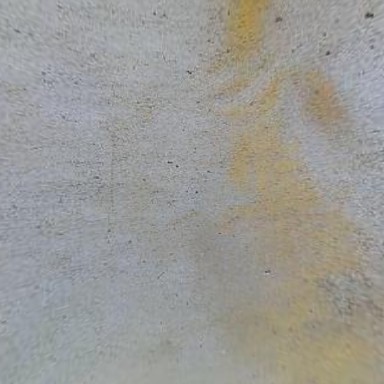}
    \end{minipage}
    \vspace{1mm}
    \begin{minipage}[c]{\doublecomparisonwidth\linewidth}
        \centering
    \includegraphics[width=\linewidth, height=0.5\linewidth]{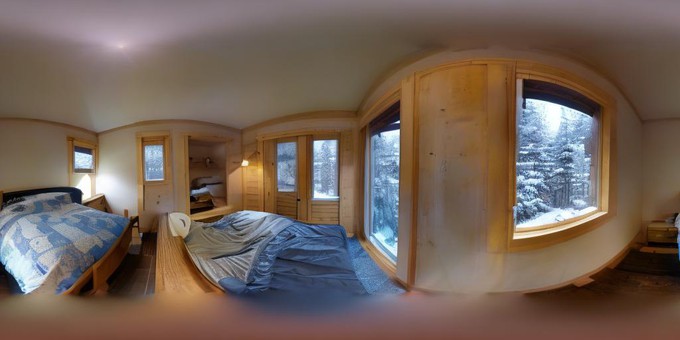}
    \end{minipage}
    \begin{minipage}[c]{\comparisonwidth\linewidth}
        \centering
    \includegraphics[width=\linewidth, height=\linewidth]{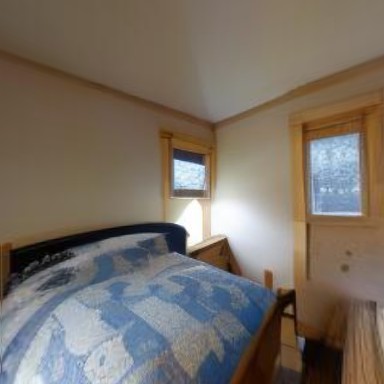}
    \end{minipage}
    \begin{minipage}[c]{\comparisonwidth\linewidth}
        \centering
    \includegraphics[width=\linewidth, height=\linewidth]{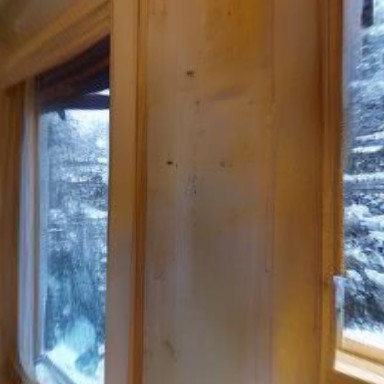}
    \end{minipage}
    \rule[0.5ex]{\linewidth}{0.5pt}
    \vspace{3pt}
    \begin{minipage}[c][0.05\linewidth]{0.015\linewidth}
        \centering
        \tiny
        \phantom{\rotatebox{90}{Text2Light}}
    \end{minipage}
    \begin{minipage}[c][0.1\linewidth]{0.03\linewidth}
        \centering
        \tiny
        \rotatebox{90}{OmniDreamer}
    \end{minipage}
    \begin{minipage}[c]{\doublecomparisonwidth\linewidth}
        \centering
    \includegraphics[width=\linewidth, height=0.5\linewidth]{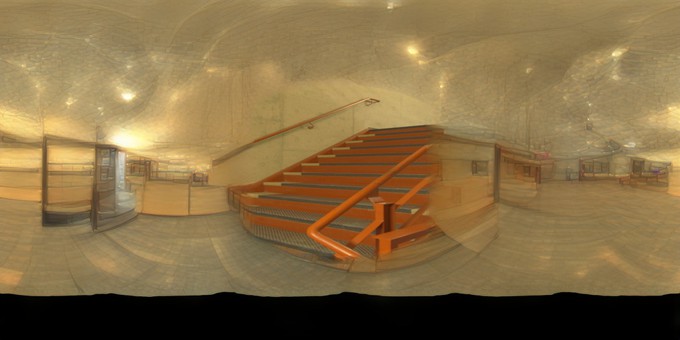}
    \end{minipage}
    \begin{minipage}[c]{\comparisonwidth\linewidth}
        \centering
    \includegraphics[width=\linewidth, height=\linewidth]{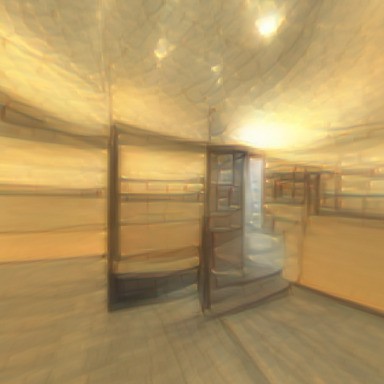}
    \end{minipage}
    \begin{minipage}[c]{\comparisonwidth\linewidth}
        \centering
    \includegraphics[width=\linewidth, height=\linewidth]{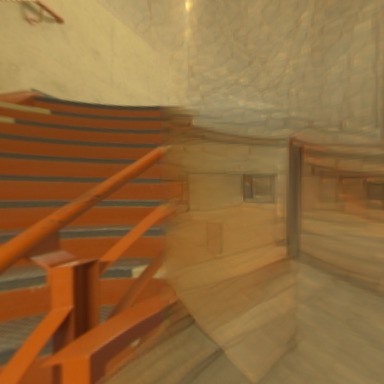}
    \end{minipage}
    \begin{minipage}[c]{\doublecomparisonwidth\linewidth}
        \centering
    \includegraphics[width=\linewidth, height=0.5\linewidth]{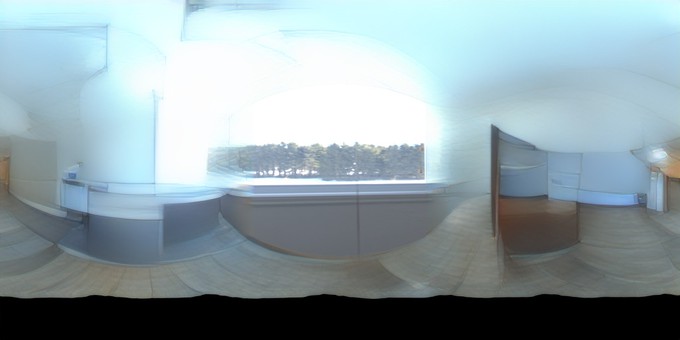}
    \end{minipage}
    \begin{minipage}[c]{\comparisonwidth\linewidth}
        \centering
    \includegraphics[width=\linewidth, height=\linewidth]{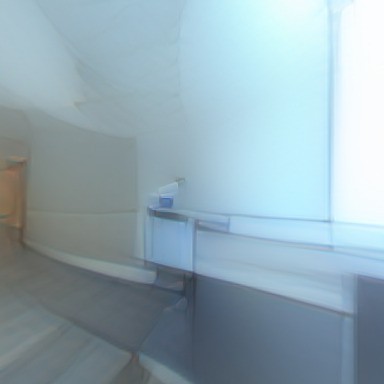}
    \end{minipage}
    \begin{minipage}[c]{\comparisonwidth\linewidth}
        \centering
    \includegraphics[width=\linewidth, height=\linewidth]{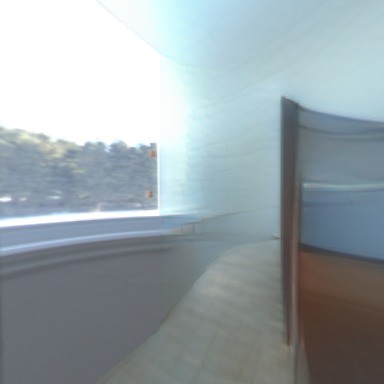}
    \end{minipage}
    \begin{minipage}[c][0.05\linewidth]{0.015\linewidth}
        \centering
        \tiny
        \rotatebox{90}{Image-only}
    \end{minipage}
    \begin{minipage}[c][0.1\linewidth]{0.03\linewidth}
        \centering
        \tiny
        \rotatebox{90}{PanoDiffusion}
    \end{minipage}
    \begin{minipage}[c]{\doublecomparisonwidth\linewidth}
        \centering
    \includegraphics[width=\linewidth, height=0.5\linewidth]{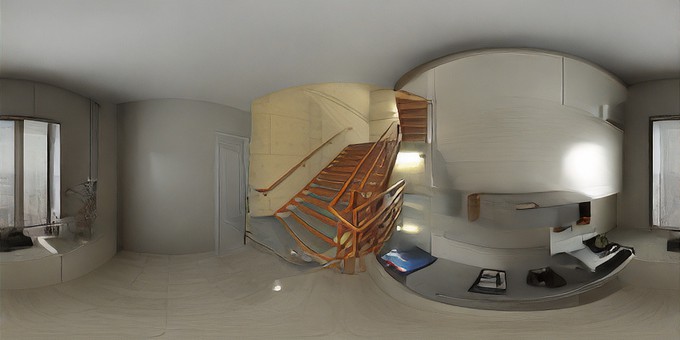}
    \end{minipage}
    \begin{minipage}[c]{\comparisonwidth\linewidth}
        \centering
    \includegraphics[width=\linewidth, height=\linewidth]{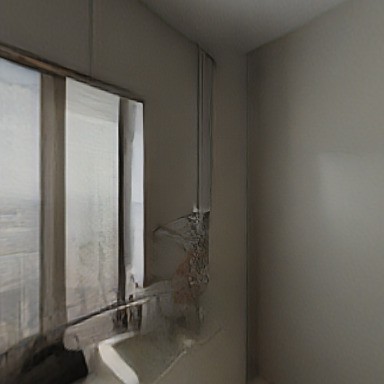}
    \end{minipage}
    \begin{minipage}[c]{\comparisonwidth\linewidth}
        \centering
    \includegraphics[width=\linewidth, height=\linewidth]{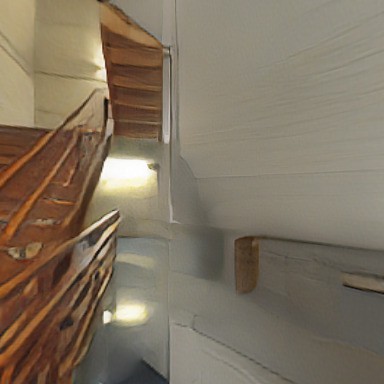}
    \end{minipage}
    \vspace{1mm}
    \begin{minipage}[c]{\doublecomparisonwidth\linewidth}
        \centering
    \includegraphics[width=\linewidth, height=0.5\linewidth]{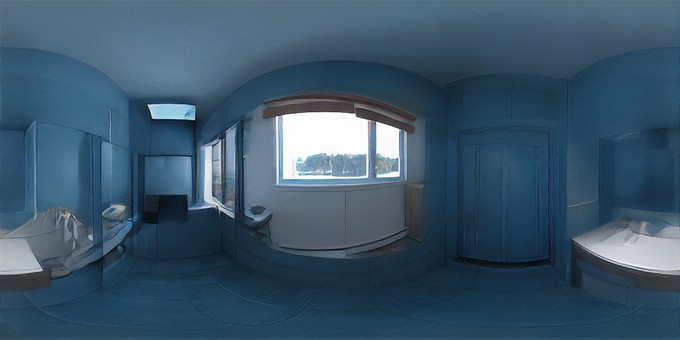}
    \end{minipage}
    \begin{minipage}[c]{\comparisonwidth\linewidth}
        \centering
    \includegraphics[width=\linewidth, height=\linewidth]{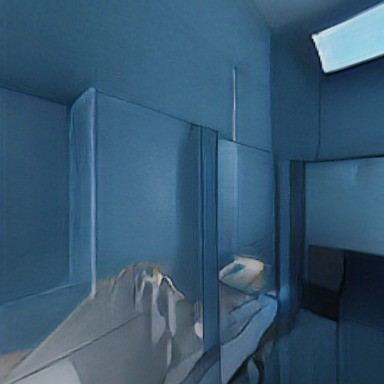}
    \end{minipage}
    \begin{minipage}[c]{\comparisonwidth\linewidth}
        \centering
    \includegraphics[width=\linewidth, height=\linewidth]{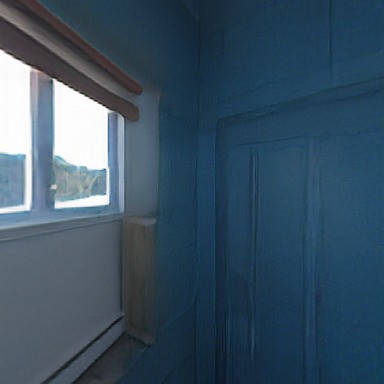}
    \end{minipage}
    \begin{minipage}[c][0.05\linewidth]{0.015\linewidth}
        \centering
        \tiny
        \phantom{\rotatebox{90}{Text2Light}}
    \end{minipage}
    \begin{minipage}[c][0.1\linewidth]{0.03\linewidth}
        \centering
        \tiny
        \rotatebox{90}{Ours\textsubscript{img}}
    \end{minipage}
    \begin{minipage}[c]{\doublecomparisonwidth\linewidth}
        \centering
    \includegraphics[width=\linewidth, height=0.5\linewidth]{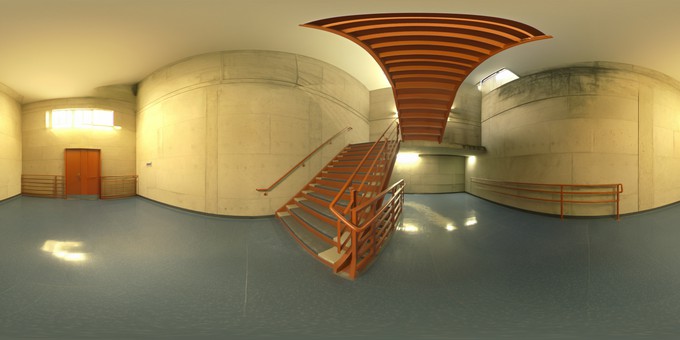}
    \end{minipage}
    \begin{minipage}[c]{\comparisonwidth\linewidth}
        \centering
    \includegraphics[width=\linewidth, height=\linewidth]{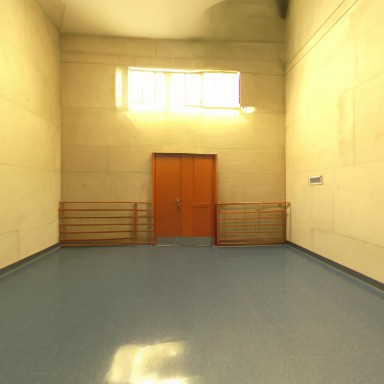}
    \end{minipage}
    \begin{minipage}[c]{\comparisonwidth\linewidth}
        \centering
    \includegraphics[width=\linewidth, height=\linewidth]{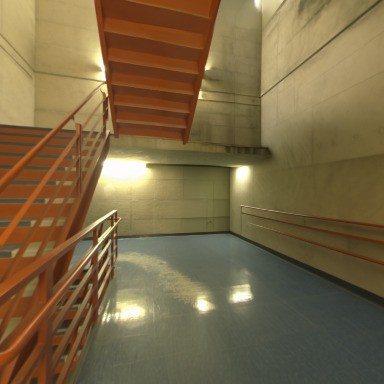}
    \end{minipage}
    \vspace{1mm}
    \begin{minipage}[c]{\doublecomparisonwidth\linewidth}
        \centering
    \includegraphics[width=\linewidth, height=0.5\linewidth]{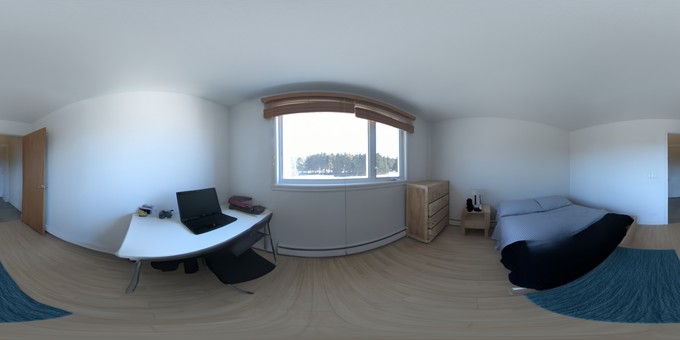}
    \end{minipage}
    \begin{minipage}[c]{\comparisonwidth\linewidth}
        \centering
    \includegraphics[width=\linewidth, height=\linewidth]{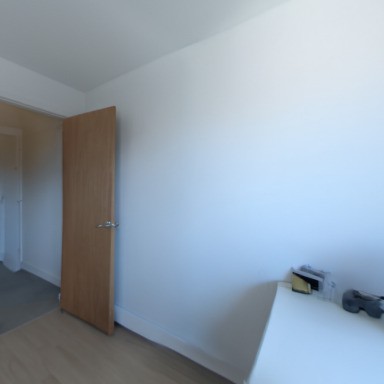}
    \end{minipage}
    \begin{minipage}[c]{\comparisonwidth\linewidth}
        \centering
    \includegraphics[width=\linewidth, height=\linewidth]{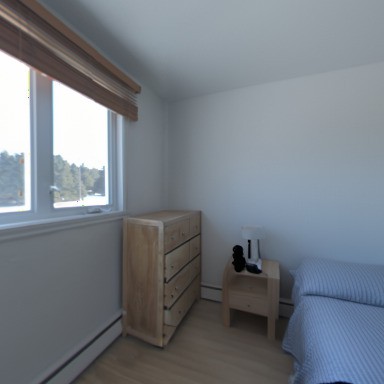}
    \end{minipage}
    \rule[0.5ex]{\linewidth}{0.5pt}     
    \vspace{3pt}
    \begin{minipage}[c][0.05\linewidth]{0.015\linewidth}
        \centering
        \tiny
        \phantom{\rotatebox{90}{Text2Light}}
    \end{minipage}
    \begin{minipage}[c][0.1\linewidth]{0.03\linewidth}
        \centering
        \tiny
        \rotatebox{90}{MVDiffusion}
    \end{minipage}
    \begin{minipage}[c]{\doublecomparisonwidth\linewidth}
        \centering
    \includegraphics[width=\linewidth, height=0.5\linewidth]{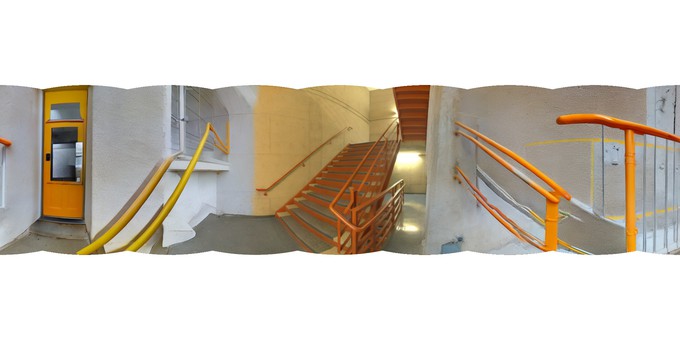}
    \end{minipage}
    \begin{minipage}[c]{\comparisonwidth\linewidth}
        \centering
    \includegraphics[width=\linewidth, height=\linewidth]{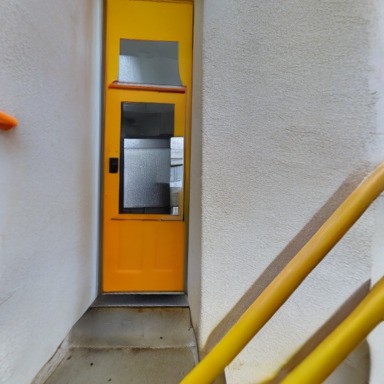}
    \end{minipage}
    \begin{minipage}[c]{\comparisonwidth\linewidth}
        \centering
    \includegraphics[width=\linewidth, height=\linewidth]{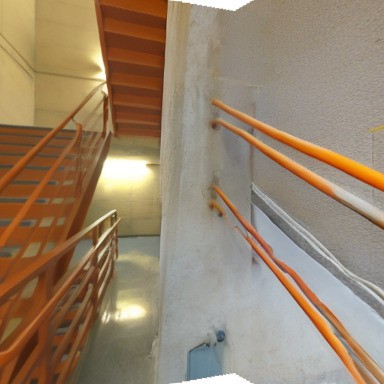}
    \end{minipage}
    \begin{minipage}[c]{\doublecomparisonwidth\linewidth}
        \centering
    \includegraphics[width=\linewidth, height=0.5\linewidth]{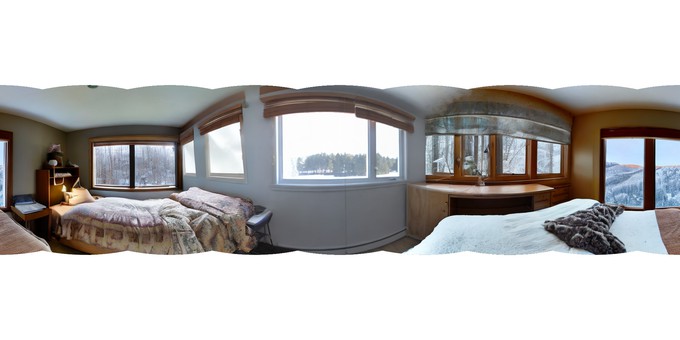}
    \end{minipage}
    \begin{minipage}[c]{\comparisonwidth\linewidth}
        \centering
    \includegraphics[width=\linewidth, height=\linewidth]{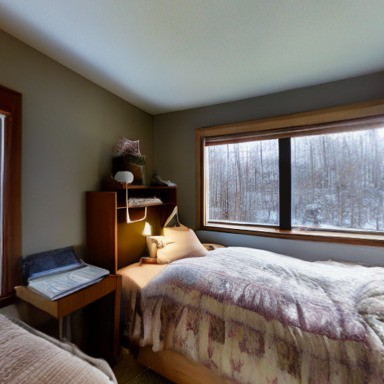}
    \end{minipage}
    \begin{minipage}[c]{\comparisonwidth\linewidth}
        \centering
    \includegraphics[width=\linewidth, height=\linewidth]{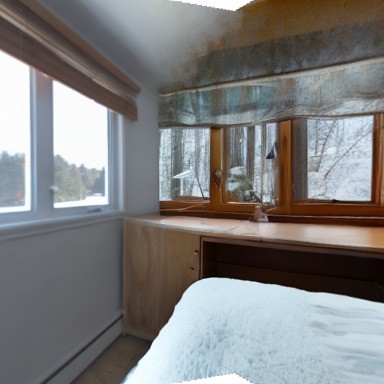}
    \end{minipage}
    \begin{minipage}[c][0.05\linewidth]{0.015\linewidth}
        \centering
        \tiny
        \rotatebox{90}{Image and Text}
    \end{minipage}
    \begin{minipage}[c][0.1\linewidth]{0.03\linewidth}
        \centering
        \tiny
        \rotatebox{90}{Diffusion360}
    \end{minipage}
    \begin{minipage}[c]{\doublecomparisonwidth\linewidth}
        \centering
    \includegraphics[width=\linewidth, height=0.5\linewidth]{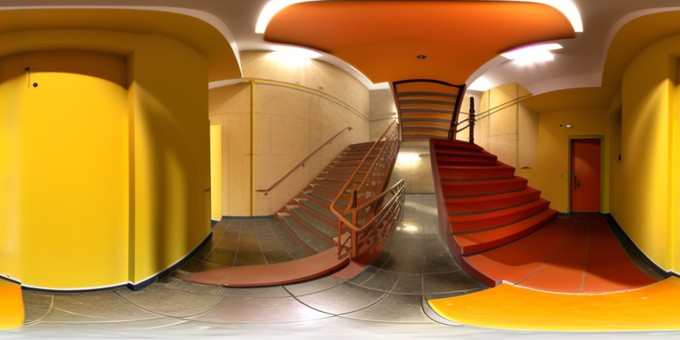}
    \end{minipage}
    \begin{minipage}[c]{\comparisonwidth\linewidth}
        \centering
    \includegraphics[width=\linewidth, height=\linewidth]{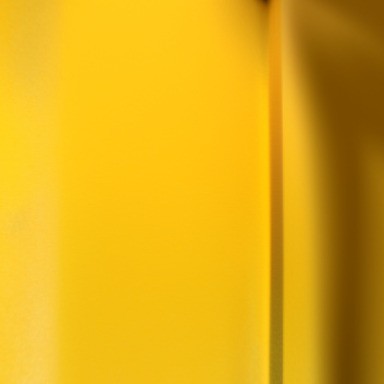}
    \end{minipage}
    \begin{minipage}[c]{\comparisonwidth\linewidth}
        \centering
    \includegraphics[width=\linewidth, height=\linewidth]{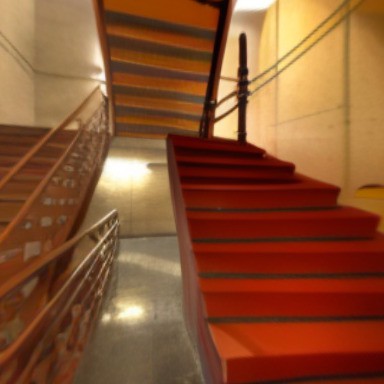}
    \end{minipage}
    \vspace{1mm}
    \begin{minipage}[c]{\doublecomparisonwidth\linewidth}
        \centering
    \includegraphics[width=\linewidth, height=0.5\linewidth]{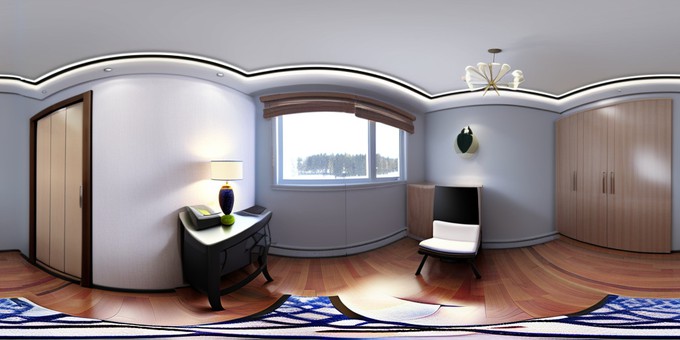}
    \end{minipage}
    \begin{minipage}[c]{\comparisonwidth\linewidth}
        \centering
    \includegraphics[width=\linewidth, height=\linewidth]{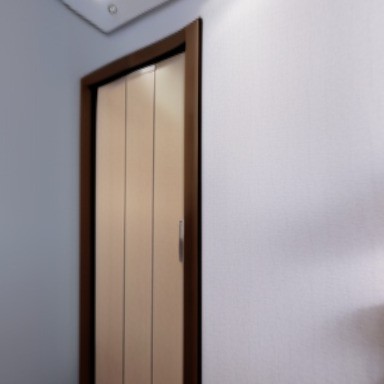}
    \end{minipage}
    \begin{minipage}[c]{\comparisonwidth\linewidth}
        \centering
    \includegraphics[width=\linewidth, height=\linewidth]{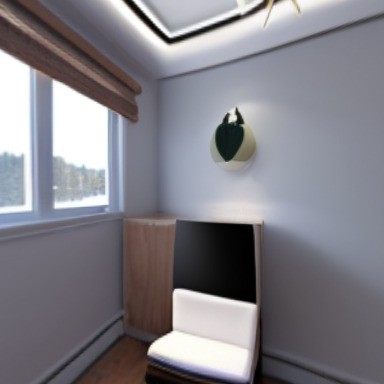}
    \end{minipage}

    \begin{minipage}[c][0.05\linewidth]{0.015\linewidth}
        \centering
        \tiny
        \phantom{\rotatebox{90}{Image and Text}}
    \end{minipage}
    \begin{minipage}[c][0.1\linewidth]{0.03\linewidth}
        \centering
        \tiny
        \rotatebox{90}{Ours\textsubscript{img+txt}}
    \end{minipage}
    \begin{minipage}[c]{\doublecomparisonwidth\linewidth}
        \centering
    \includegraphics[width=\linewidth, height=0.5\linewidth]{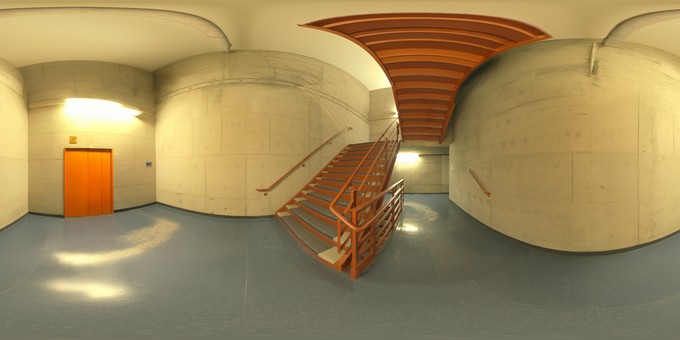}
    \end{minipage}
    \begin{minipage}[c]{\comparisonwidth\linewidth}
        \centering
    \includegraphics[width=\linewidth, height=\linewidth]{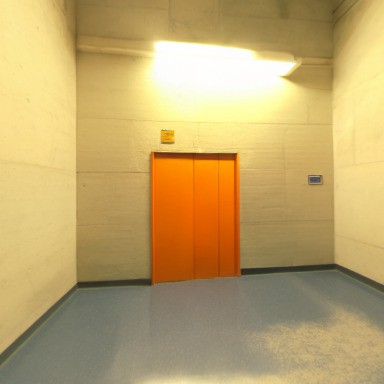}
    \end{minipage}
    \begin{minipage}[c]{\comparisonwidth\linewidth}
        \centering
    \includegraphics[width=\linewidth, height=\linewidth]{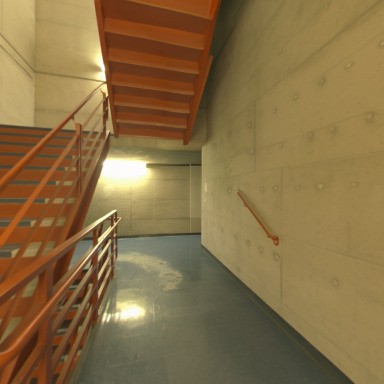}
    \end{minipage}
    \vspace{1mm}
    \begin{minipage}[c]{\doublecomparisonwidth\linewidth}
        \centering
    \includegraphics[width=\linewidth, height=0.5\linewidth]{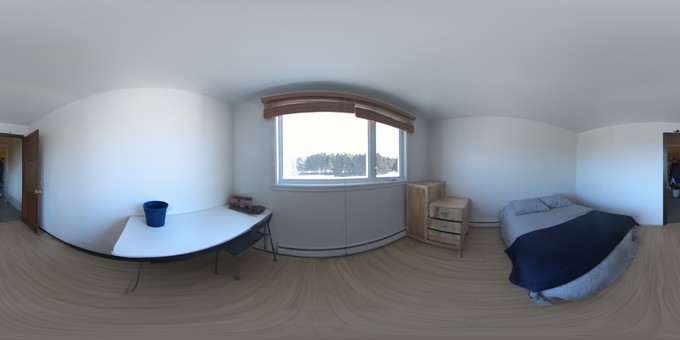}
    \end{minipage}
    \begin{minipage}[c]{\comparisonwidth\linewidth}
        \centering
    \includegraphics[width=\linewidth, height=\linewidth]{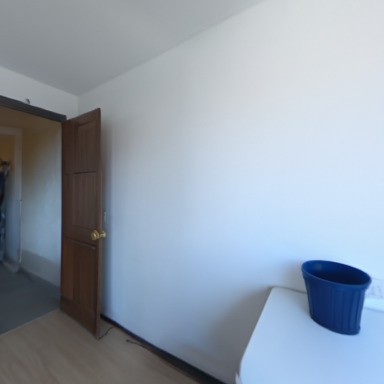}
    \end{minipage}
    \begin{minipage}[c]{\comparisonwidth\linewidth}
        \centering
    \includegraphics[width=\linewidth, height=\linewidth]{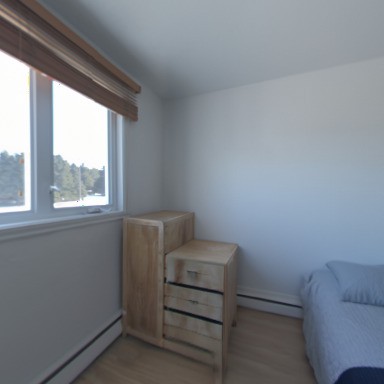}
    \end{minipage}
    \rule[0.5ex]{\linewidth}{0.5pt}
    \vspace{3pt}
    \begin{minipage}[c][0.05\linewidth]{0.015\linewidth}
        \centering
        \tiny
        \phantom{\rotatebox{90}{Image}}
    \end{minipage}
    \begin{minipage}[c][0.1\linewidth]{0.03\linewidth}
        \centering
        \tiny
        \rotatebox{90}{Ours\textsubscript{img+multitxt}}
    \end{minipage}
    \begin{minipage}[c]{\doublecomparisonwidth\linewidth}
        \centering
    \includegraphics[width=\linewidth, height=0.5\linewidth]{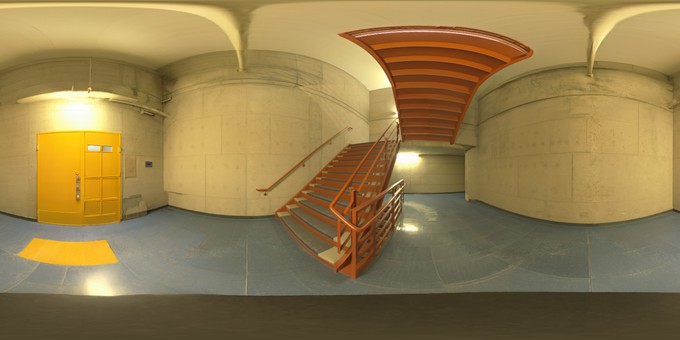}
    \end{minipage}
    \begin{minipage}[c]{\comparisonwidth\linewidth}
        \centering
    \includegraphics[width=\linewidth, height=\linewidth]{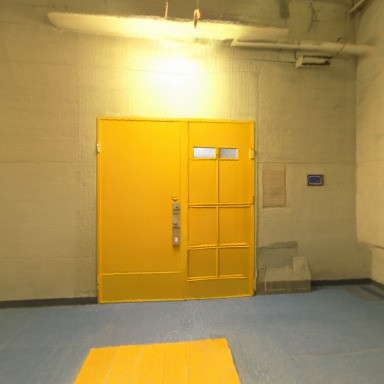}
    \end{minipage}
    \begin{minipage}[c]{\comparisonwidth\linewidth}
        \centering
    \includegraphics[width=\linewidth, height=\linewidth]{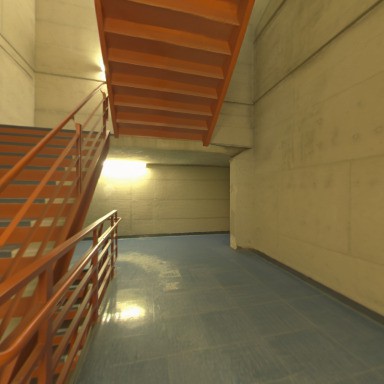}
    \end{minipage}
    \vspace{1mm}
    \begin{minipage}[c]{\doublecomparisonwidth\linewidth}
        \centering
    \includegraphics[width=\linewidth, height=0.5\linewidth]{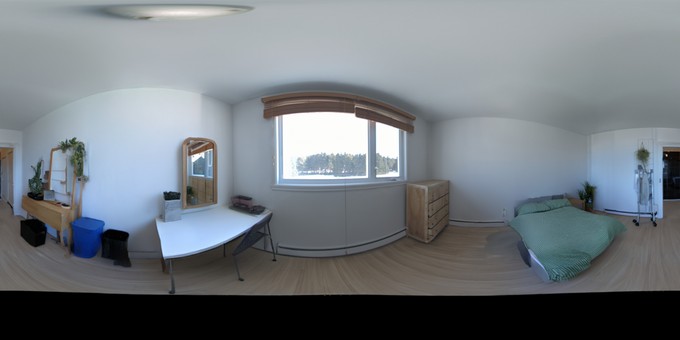}
    \end{minipage}
    \begin{minipage}[c]{\comparisonwidth\linewidth}
        \centering
    \includegraphics[width=\linewidth, height=\linewidth]{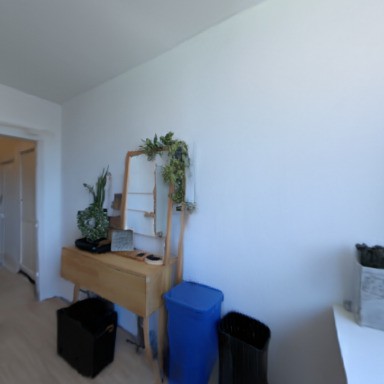}
    \end{minipage}
    \begin{minipage}[c]{\comparisonwidth\linewidth}
        \centering
    \includegraphics[width=\linewidth, height=\linewidth]{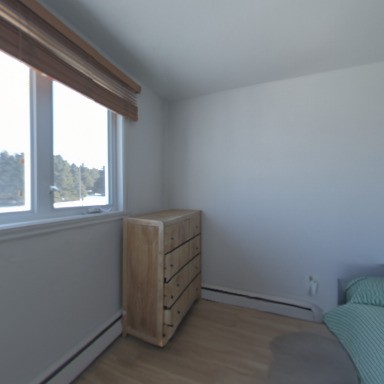}
    \end{minipage}
    \caption{\textbf{Qualitative comparison between \emph{CubeDiff} and baselines on the LAVAL Indoor Dataset.} Besides Text2Light, all panoramas are generated using the center face as input condition and additional text prompts if applicable. For each sample we show the panorama image as well as two projected images. Please zoom in to compare the different methods.}
    \label{fig:comparison}
\end{figure}

%% file: sections/conclusion.tex
\section{Conclusion}
This work introduces a novel approach to panorama generation leveraging pretrained text-to-image diffusion models applied to a cubemap representation. By enabling attention across the cube faces, our method achieves state-of-the-art results in terms of visual fidelity and coherence, while requiring minimal architectural changes. This approach not only inherits the strengths of existing diffusion models, including high-resolution synthesis and generalization capabilities, but also unlocks fine-grained text control over the generated panorama.  This opens up exciting new possibilities for creative applications and paves the way for future research in controllable panorama generation.

%% file: sections/appendix.tex
\newpage
\appendix
\section{Appendix}
\begin{figure}[t]
    \centering
        \vspace{1.5mm}
    \begin{minipage}[c]{\linewidth}
        \centering
        \includegraphics[width=\linewidth, height=0.5\linewidth]{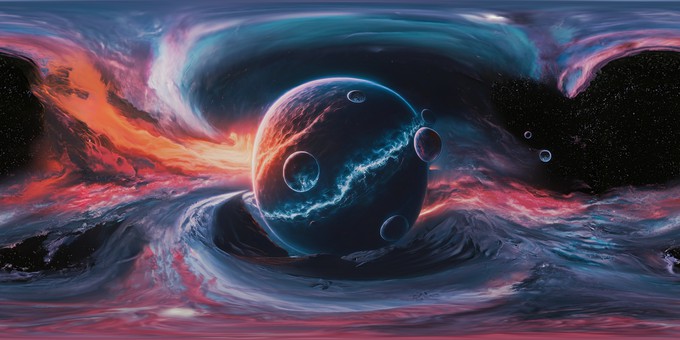}
        
    \end{minipage}
    \caption{\textbf{Panorama in a out of training-distribution setting.}}
    \label{fig:ood:first}
\end{figure}

\subsection{Out-Of-Distribution examples}
\begin{figure}[!ht]
    \centering
    \includegraphics[width=0.495\linewidth]{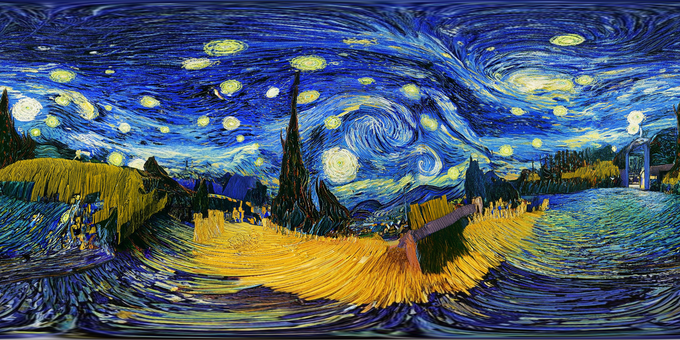}%
    \hfill
    \includegraphics[width=0.495\linewidth]{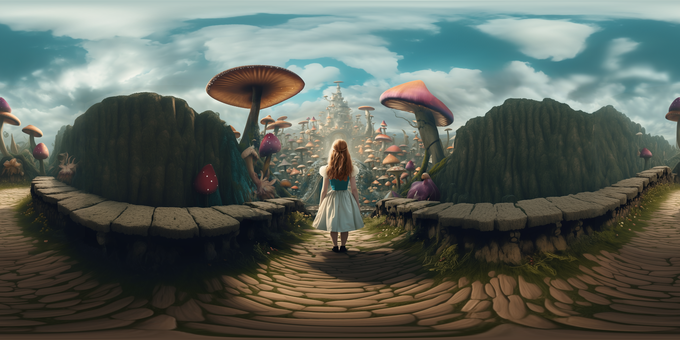}
    \includegraphics[width=0.495\linewidth]{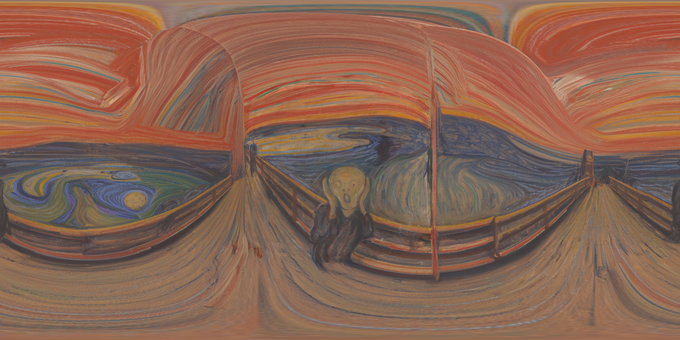}%
    \hfill
    \includegraphics[width=0.495\linewidth]{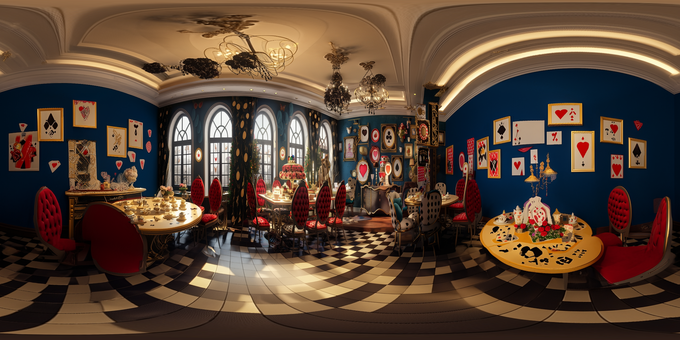}
    \includegraphics[width=0.495\linewidth]{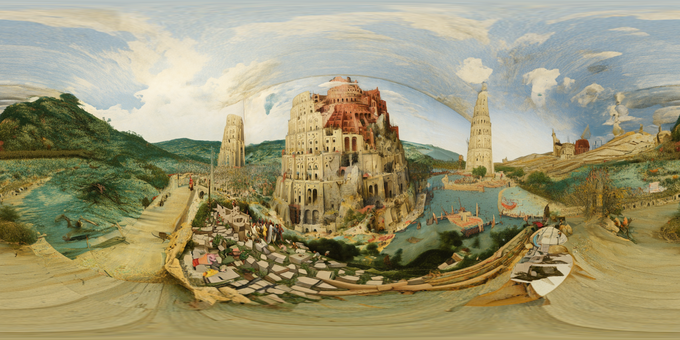}%
    \hfill
    \includegraphics[width=0.495\linewidth]{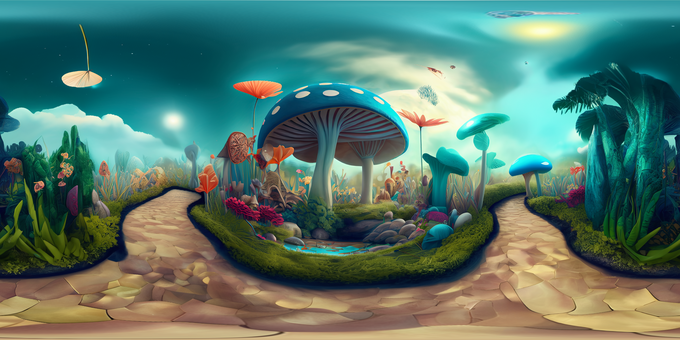}
    \caption{\textbf{Out-of-distribution examples generated by Cubediff.} On the right are artistic generations, on the left fantasy sceneries like Alice in Wonderland.}
    \label{fig:ood}
\end{figure}

\subsection{Text prompt examples}
The text prompts for the qualitative examples are the following:

\paragraph{City skyline}
\begin{itemize}
    \item A glittering cityscape at night, densely packed high-rise buildings illuminated against a dark sky.  The buildings are various heights and designs, reflecting light onto the calm water in the foreground. A hill or mountain range forms a dark backdrop behind the city.
    \item Dark, silhouetted mountains stretching across the entire view, with the faintest glow of city lights on the horizon. The sky is a deep, nighttime blue.
    \item The side profile of a mountainous region, dark and slightly textured, extending from the foreground to the distant horizon.  The city lights are visible in the distance to the right, providing a subtle contrast against the mountain's blackness.
    \item Similar to the right. A darker, less detailed section of the mountain range appears to the left.
    \item An expansive, dark night sky with subtle cloud formations. The upper portions of the city's tallest buildings are faintly visible as a horizontal line against the darkness.
    \item The dark, calm water of the bay reflecting the city lights, showing subtle ripples and disturbances. The reflection is most intense closest to the buildings and gradually fades into darkness.
\end{itemize}

\paragraph{Green house}
\begin{itemize}
    \item A dense, overgrown greenhouse teems with lush greenery, creating a vibrant, jungle-like scene. Large ferns, broad-leafed tropical plants, and hanging baskets overflowing with foliage dominate the view. Large rocks, blanketed in thick, *vibrant* green moss, suggest a path, nearly disappearing beneath the surrounding plants. The glass roof, tinged with green, casts a soft, diffused light.
    \item The right side of the greenhouse is a solid wall of plant life. Large, vibrant green leaves and hanging plants create a dense, tropical atmosphere. The rocks, almost entirely smothered in *bright* green moss and overflowing greenery, offer only the faintest hint of a path.  Wild, untamed growth dominates the scene.
    \item The greenhouse extends into the distance, an endless expanse of green fading into shadow. The rocks, now completely obscured by plants and a thick carpet of *emerald* moss, suggest a path swallowed by the jungle. The sheer volume of greenery creates a sense of depth and wildness. The glass roof is barely visible.
    \item To the left, a tall palm tree rises amidst the dense foliage. Large ferns and other leafy plants create a solid wall of green around it. The rocks, barely discernible beneath the thick layer of *brilliant* green moss and creeping vines, continue on this side, obscuring the path almost entirely.
    \item Looking up, the glass panes of the greenhouse roof, tinted green with clinging vines and moss, are mostly obscured by the dense canopy. Hanging plants, heavy with ferns and leafy vines, cascade downwards, creating a lush, verdant ceiling. The metal framework is almost entirely hidden.
    \item Looking down, the large rocks are almost completely hidden beneath a thick carpet of *luminescent*, almost *glowing* green moss, fallen leaves, vines, and other plant debris. Only the edges of the rocks peek through the dense greenery, making the path nearly invisible. The texture of the moss appears incredibly soft and velvety, a seamless blanket of vibrant green.
\end{itemize}

\paragraph{Living room}
\begin{itemize}
    \item A cozy, modern living room with large windows allowing natural light to flood in. The room is furnished with a soft, gray sofa facing the windows and a wooden coffee table in the center. A bookshelf filled with plants and books stands against the far wall. A fireplace in the corner crackles softly, casting a warm glow across the room.
    \item A side view of the living room from the right, showing the side of the gray sofa facing toward the large windows. The coffee table is positioned in front of the sofa. To the left, the wall-mounted television is visible above the fireplace, and the wooden floor stretches across the room. A small side table with a lamp sits next to the sofa.
    \item A view of the back wall of the room, where the bookshelf is the primary focus. The large windows let in a soft light, but the sofa and coffee table are out of sight from this angle. The fireplace is visible on the right side of the room, softly illuminating the space, and the television mounted above it is partially visible.
    \item A side view from the left side of the living room, showing the bookshelf along the far wall, and the curtains gently swaying in front of the large windows. The coffee table sits on a stylish rug, but the sofa itself is mostly out of view, hidden from this angle. The soft glow of the fireplace on the far side of the room adds warmth to the scene.
    \item Looking up at the ceiling, the room features modern recessed lighting, casting a soft, even glow across the space. Wooden beams accent the edges of the ceiling, adding a rustic touch. The tops of the windows and the moving curtains are visible from this angle, as the diffuse light from outside fills the room.
    \item Looking down at the floor, you see polished wooden floorboards and a stylish area rug that lies under the coffee table. The legs of the sofa are visible at the edge of the rug, and a few books and a small potted plant sit on the coffee table. The contrast between the rug and the wooden floor gives the room a balanced, warm feel.
\end{itemize}

\paragraph{Snow landscape}
\begin{itemize}
    \item A view of a frozen river winding through a snowy forest. The river is partially frozen, with snow-covered banks on either side and dark, still water visible in the center. Bare, leafless trees line the edges of the river, their branches covered in snow. The overcast sky casts a soft, cold light over the scene, creating a peaceful and serene winter atmosphere.
    \item A side view of the snowy forest, showing the frozen river cutting through the landscape. Bare trees with snow-covered branches stretch across the scene, and the forest extends into the distance, with the winding river creating a natural divide. The snowy ground and trees give the area a quiet, isolated feel, and the soft light from the sky casts long shadows across the snow.
    \item A view from behind the river, where the water flows into the distance, disappearing into the snow-covered forest. The winding shape of the river is prominent, with the snow forming smooth, white edges along the banks. The bare trees on either side create a tunnel-like effect as they stretch over the river. The air feels crisp and cold, and the entire landscape is blanketed in a thick layer of snow.
    \item A left side view of the river, with snow-covered branches hanging over the water. The frozen river winds through the snowy forest, with the bare trees standing tall on either side. The snow is thick and untouched, creating a pristine winter scene. The soft light of the overcast sky adds a calm, cold atmosphere to the landscape.
    \item An aerial view of the frozen river, showing its winding path through the snow-covered forest. The dark water contrasts sharply with the white snow, and the leafless trees form a web of branches stretching across the landscape. The river curves gently through the scene, and the blanket of snow gives the forest a peaceful, quiet feel.
    \item Looking down at the frozen river from above, the snow forms a thick blanket along the banks, with patches of dark water visible in the center. The ground is covered in snow, with bare tree branches reaching over the river. The scene feels still and cold, with no signs of movement, and the snow seems to absorb all sound, creating a peaceful, quiet atmosphere.
\end{itemize}

For completeness, we provide the text prompts used for the qualitative comparison on the Laval Indoor dataset. On the left, we depict the input image and on the right be provide the text prompt.
\begin{table}[ht]
    \centering
    \begin{tabular}{ll}
        \raisebox{-1.7cm}{\includegraphics[width=0.25\linewidth]{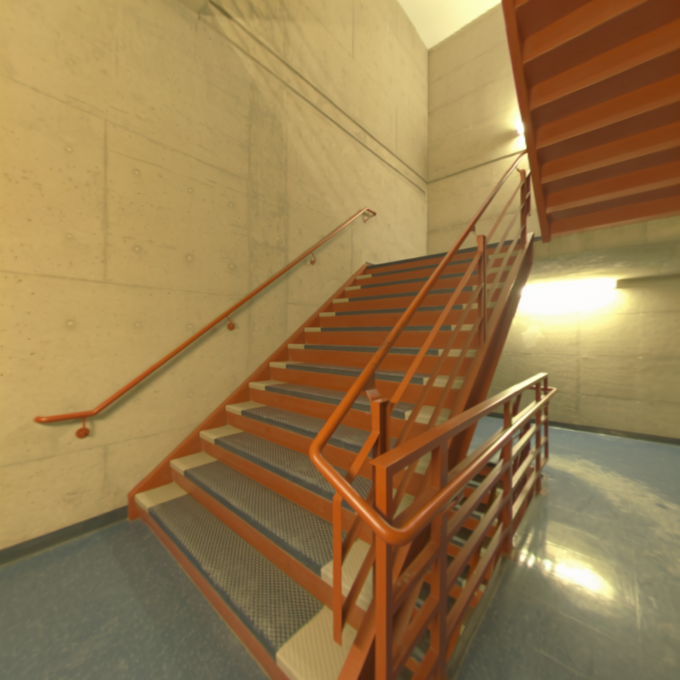}}  & \parbox{7cm}{ A concrete stairwell with orange railings leads up to a yellow door with a number 2 on it. } \\
         \raisebox{-1.7cm}{\includegraphics[width=0.25\linewidth]{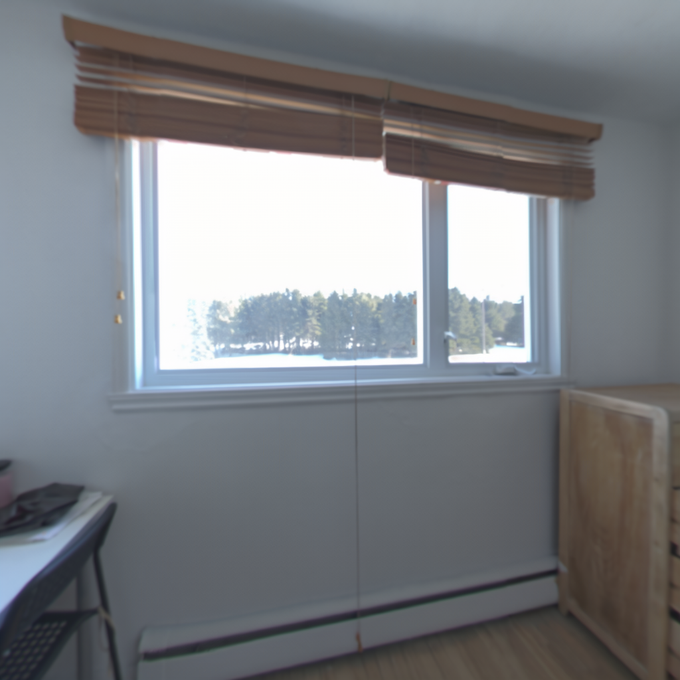}}  &
           \parbox{7cm}{A bedroom with a window overlooking a snowy forest, a bed, a desk, and a dresser.} \\
    \end{tabular}
\end{table}

\subsection{Perspective images for evaluation}
Please note that all perceptual/text alignment metrics require another network to be computed. However, as these networks are trained with perspective images alone, the metrics would not give meaningful results when computed on panoramas. To circumvent this problem, we instead we render 10 random perspective images with a FOV of 90$\degree$ for each panorama and use those for metric computation. Notice that we do not sample with an elevation of less than -45$\degree$ or more then 45$\degree$ as other works such as MVDiffusion do not generate full 360$\degree$ panoramas.

\subsection{User study}
\begin{figure}[ht]
    \centering
    \includegraphics[width=0.5\linewidth]{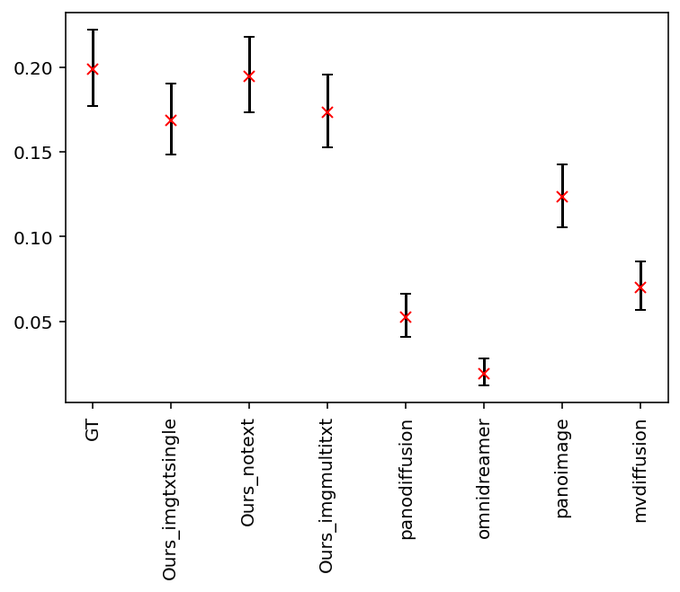}
    \caption{\textbf{Results of User study.} In this figure, we show the percentage of wins against all draws including the confidence interval}
    \label{fig:user-study}
\end{figure}

As described in the main paper we perform a two-alternative forced choice (2AFC) considering all competitors and variants of our method. In \Figref{fig:user-study}, we show the percentage of wins against all 1258 draws and corresponding confidence intervals. This indicates ours methods performs significantly better in the user study.

\subsection{More Qualitative Results}
\begin{figure}
    \centering
    \begin{minipage}[c]{0.997\linewidth}
        \centering
        \FramedBox{0.1\linewidth}{0.162\linewidth}{A pink leather couch sits in a brightly lit living room with a large window and a glass door.}
        \hspace{-0.25cm}
        \FramedBox{0.1\linewidth}{0.162\linewidth}{A living room with a window, a TV, and some bags on the floor.}
        \hspace{-0.25cm}
        \FramedBox{0.1\linewidth}{0.162\linewidth}{A white futon sits in a corner of a sparsely decorated living room with a TV and a mirror on the wall.}
        \hspace{-0.25cm}
        \FramedBox{0.1\linewidth}{0.162\linewidth}{A messy living room with a white futon, a white leather chair, a brown leather couch, and a pair of skis leaning against the wall.}
        \hspace{-0.25cm}
        \FramedBox{0.1\linewidth}{0.162\linewidth}{ white ceiling with a light fixture in the center.}
        \hspace{-0.25cm}
        \FramedBox{0.1\linewidth}{0.162\linewidth}{A black circle is in the center of a wooden floor surrounded by furniture and bags.}
    \end{minipage}
    \begin{minipage}[c]{0.997\linewidth}
        \centering
        \fboxrule=1pt
        \fcolorbox{green}{green}{\includegraphics[width=0.162\linewidth, height=0.162\linewidth]{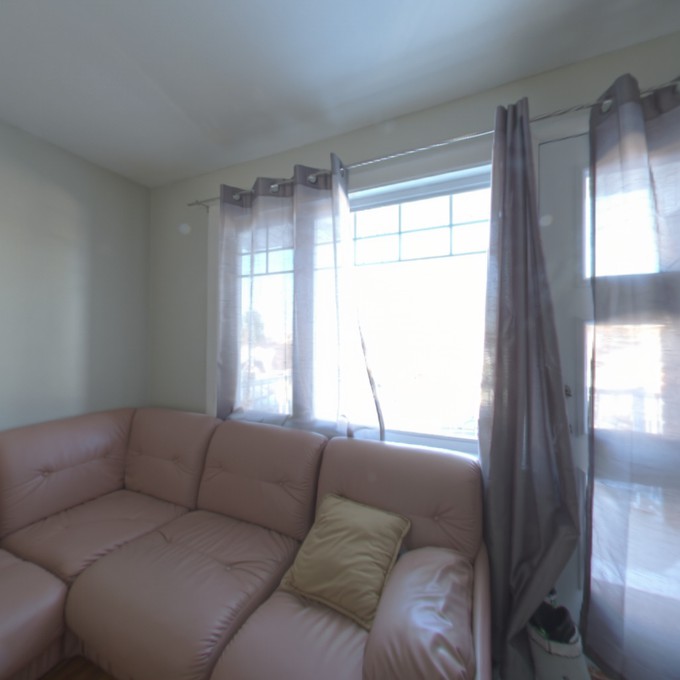}}%
        \hfill
        \includegraphics[width=0.162\linewidth, height=0.162\linewidth]{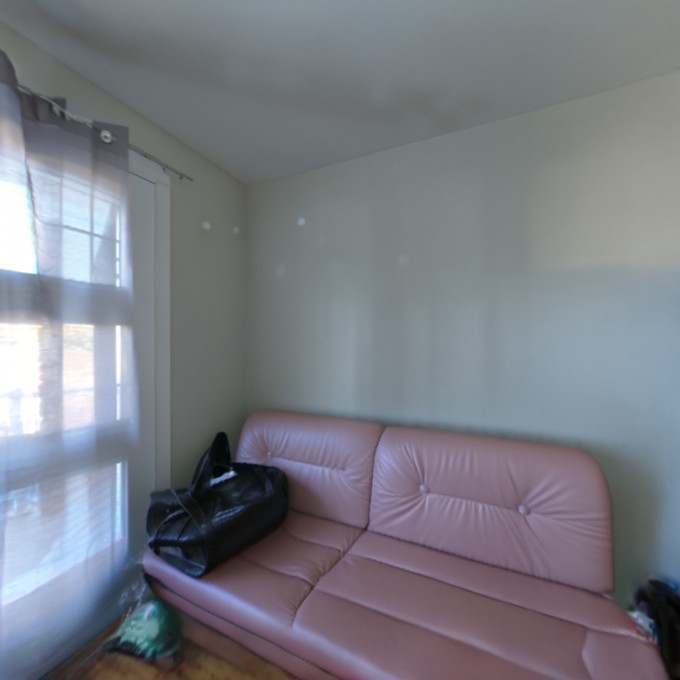}%
        \hfill
        \includegraphics[width=0.162\linewidth, height=0.162\linewidth]{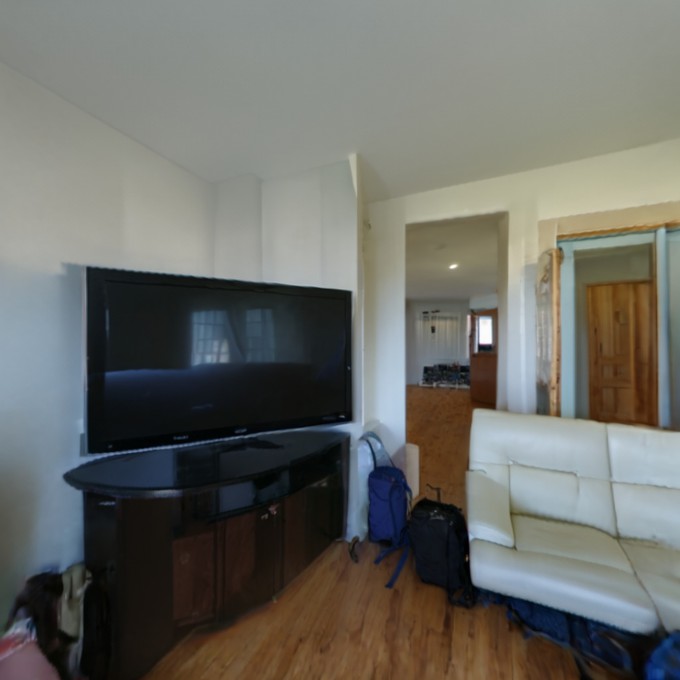}%
        \hfill
        \includegraphics[width=0.162\linewidth, height=0.162\linewidth]{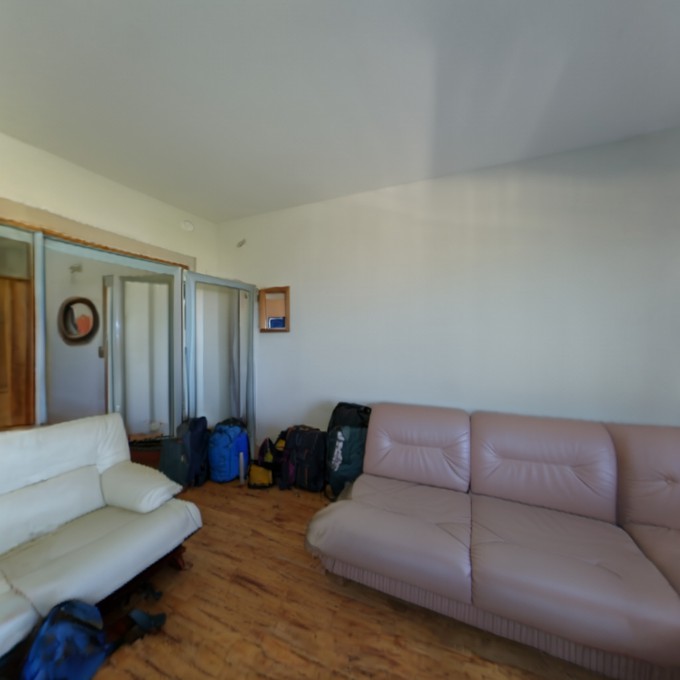}%
        \hfill
        \includegraphics[width=0.162\linewidth, height=0.162\linewidth]{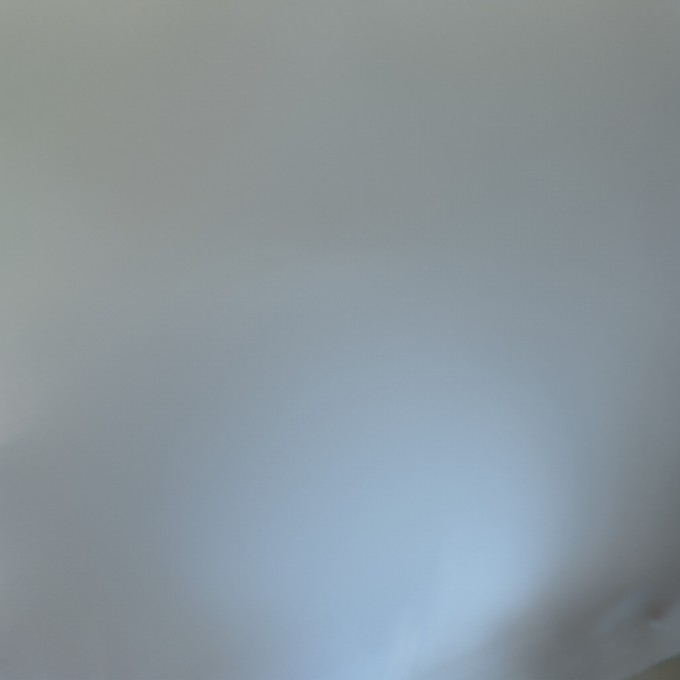}%
        \hfill
        \includegraphics[width=0.162\linewidth, height=0.162\linewidth]{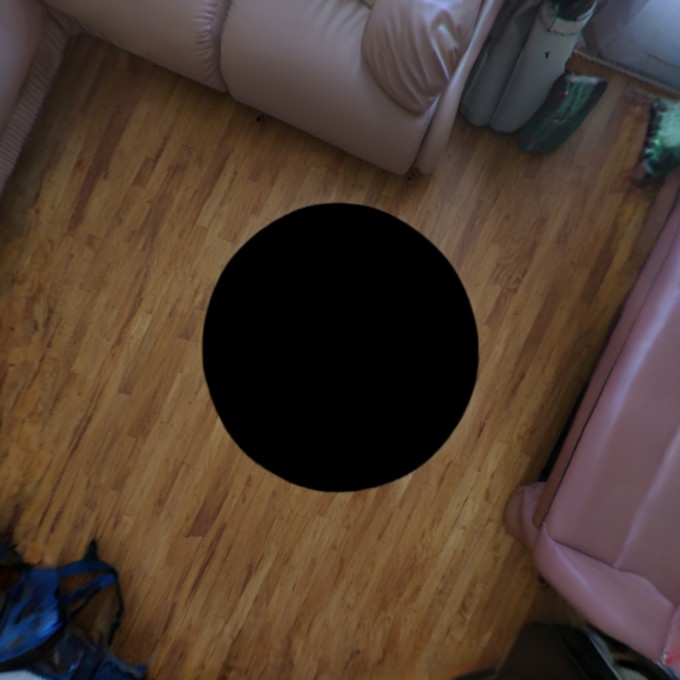}%
    \end{minipage}

    \begin{minipage}[c]{0.997\linewidth}
        \centering
        \includegraphics[width=\linewidth, height=0.5\linewidth]{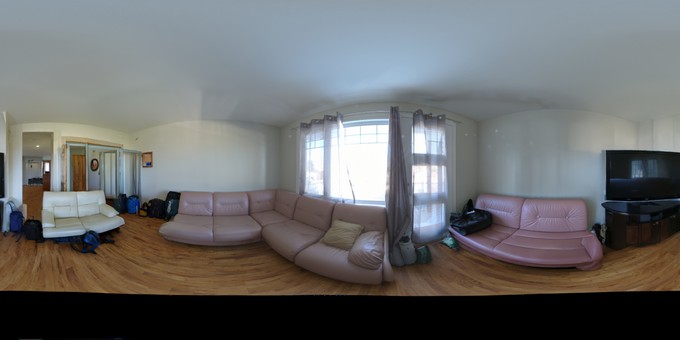}
    \end{minipage}

    \begin{minipage}[c]{0.997\linewidth}
        \centering
        \FramedBox{0.12\linewidth}{0.162\linewidth}{A white chest freezer sits in front of a window with partially open blinds, letting in bright sunlight, next to a white trash can in a room with light green walls.}
        \hspace{-0.25cm}
        \FramedBox{0.12\linewidth}{0.162\linewidth}{Sunlight streams through a window and illuminates two black dressers, one tall and one short, with various items on top of them.}
        \hspace{-0.25cm}
        \FramedBox{0.12\linewidth}{0.162\linewidth}{A messy bedroom with an open closet door revealing clothes, a desk with a printer and jackets on the chair, and a bulletin board on the wall.}
        \hspace{-0.25cm}
        \FramedBox{0.12\linewidth}{0.162\linewidth}{A messy bed with a laptop on it sits in the corner of a bedroom with a closet door open.}
        \hspace{-0.25cm}
        \FramedBox{0.12\linewidth}{0.162\linewidth}{A single, round, recessed light fixture illuminates a smooth, beige ceiling.}
        \hspace{-0.25cm}
        \FramedBox{0.12\linewidth}{0.162\linewidth}{A circular hole in the wooden floor of a bedroom is seen from above.}
    \end{minipage}
    \begin{minipage}[c]{0.997\linewidth}
        \centering
        \fboxrule=1pt
        \fcolorbox{green}{green}{\includegraphics[width=0.162\linewidth, height=0.162\linewidth]{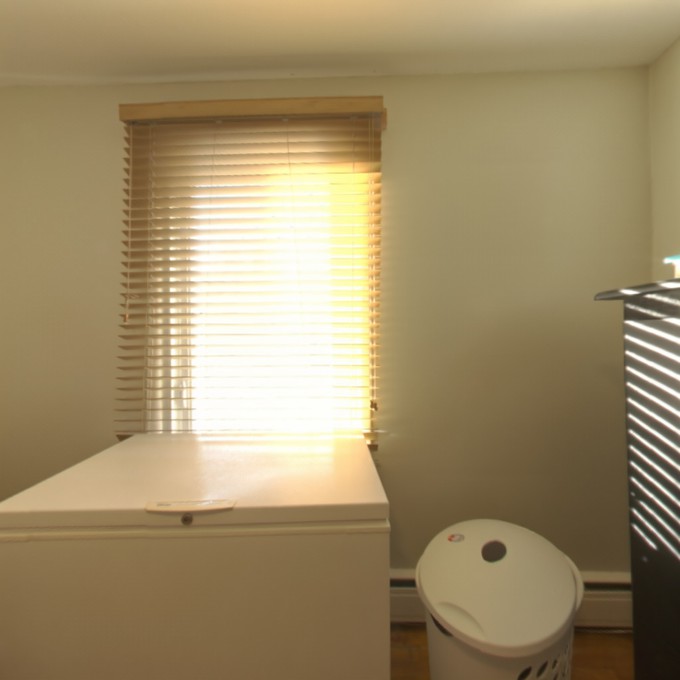}}%
        \hfill
        \includegraphics[width=0.162\linewidth, height=0.162\linewidth]{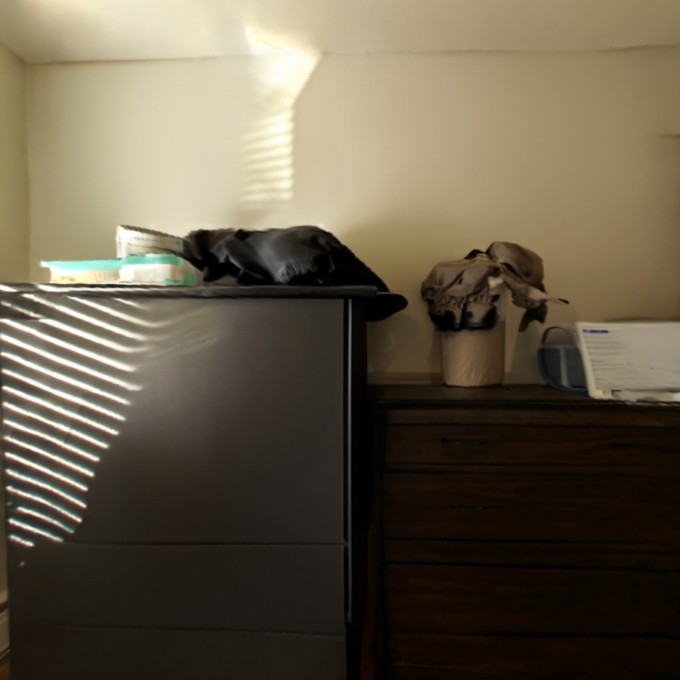}%
        \hfill
        \includegraphics[width=0.162\linewidth, height=0.162\linewidth]{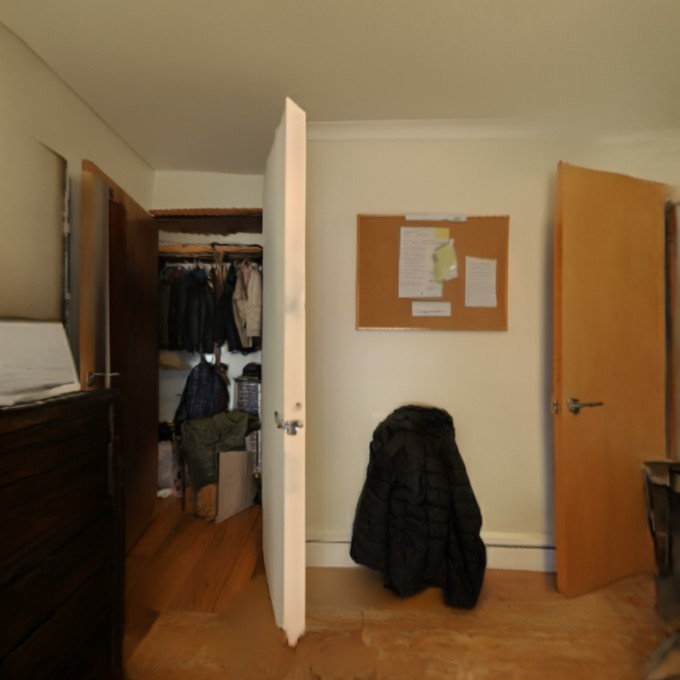}%
        \hfill
        \includegraphics[width=0.162\linewidth, height=0.162\linewidth]{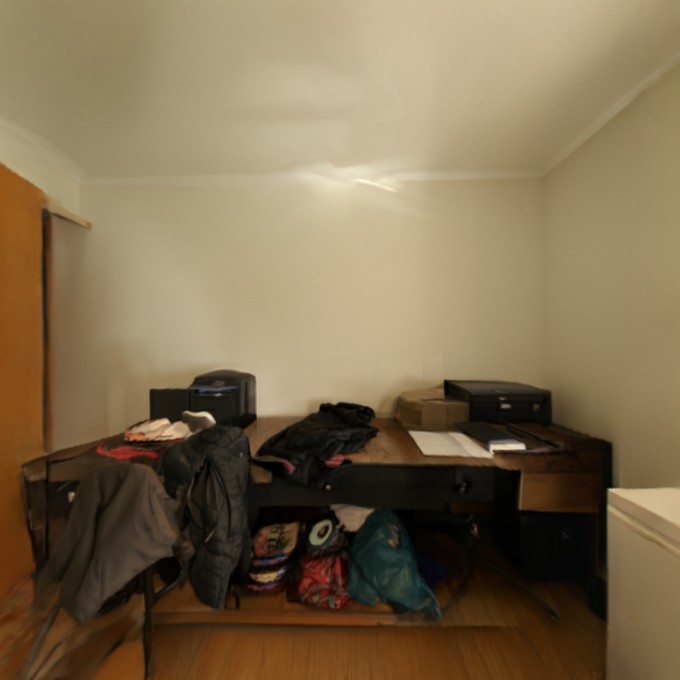}%
        \hfill
        \includegraphics[width=0.162\linewidth, height=0.162\linewidth]{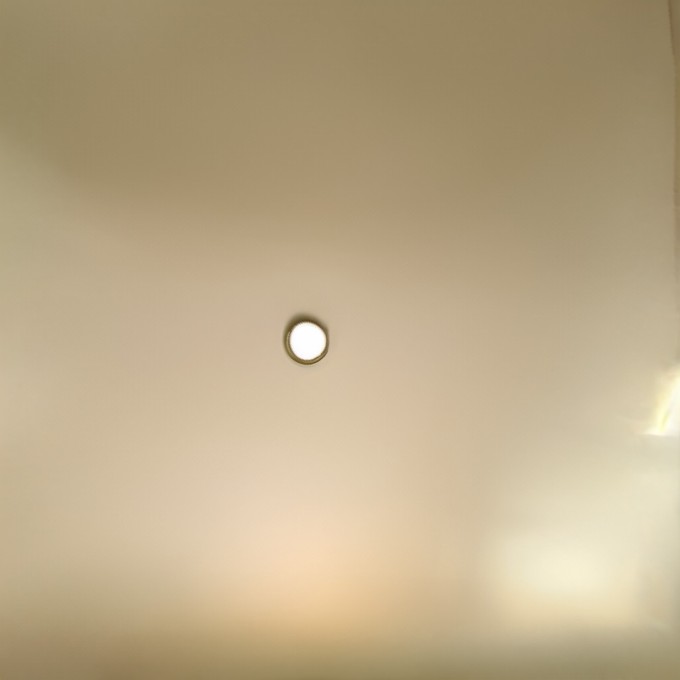}%
        \hfill
        \includegraphics[width=0.162\linewidth, height=0.162\linewidth]{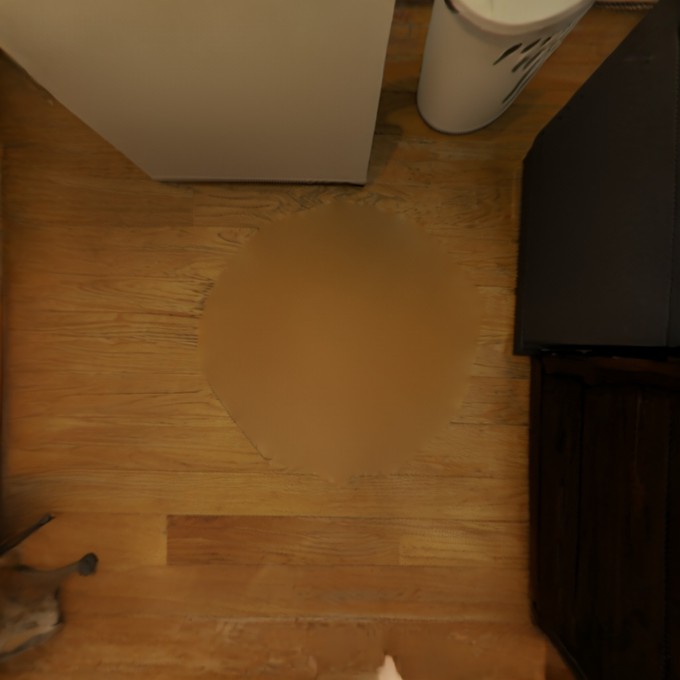}%
    \end{minipage}

    \begin{minipage}[c]{0.997\linewidth}
        \centering
        \includegraphics[width=\linewidth, height=0.5\linewidth]{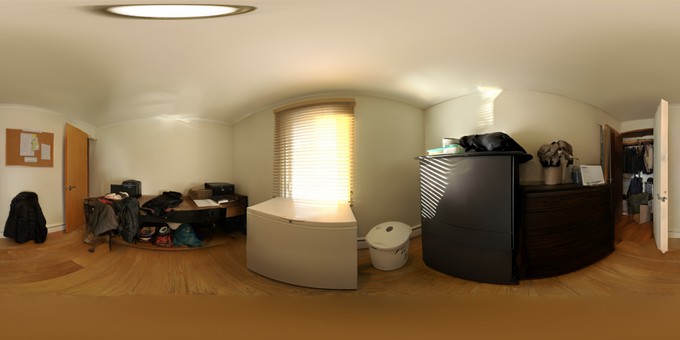}
    \end{minipage}
    \hfill
    \caption{\textbf{Generated panoramas with multiple text prompts and image condition}}
    \label{fig:mulittext}
\end{figure}

\begin{figure}
    \centering
    \begin{minipage}[c]{0.997\linewidth}
        \centering
        \FramedBox{0.1\linewidth}{0.162\linewidth}{A hallway with a brick wall, a women's bathroom door, and three recycling bins.}
        \hspace{-0.25cm}
        \FramedBox{0.1\linewidth}{0.162\linewidth}{A brightly lit cafeteria with tables and chairs is mostly empty, with sunlight coming in from the right side.}
        \hspace{-0.25cm}
        \FramedBox{0.1\linewidth}{0.162\linewidth}{ hallway with a brick wall and a brown door with a wheelchair symbol on it leads to a cafeteria with red and white chairs and tables.}
        \hspace{-0.25cm}
        \FramedBox{0.1\linewidth}{0.162\linewidth}{A hallway with a tiled floor and brick walls leads to a gymnasium with a basketball court visible through the open double doors.}
        \hspace{-0.25cm}
        \FramedBox{0.1\linewidth}{0.162\linewidth}{ A bright fluorescent light fixture shines brightly against a beige ceiling with square tiles.}
        \hspace{-0.25cm}
        \FramedBox{0.1\linewidth}{0.162\linewidth}{A black circular hole is in the center of a tiled floor.}
    \end{minipage}
    \begin{minipage}[c]{0.997\linewidth}
        \centering
        \fboxrule=1pt
        \fcolorbox{green}{green}{\includegraphics[width=0.162\linewidth, height=0.162\linewidth]{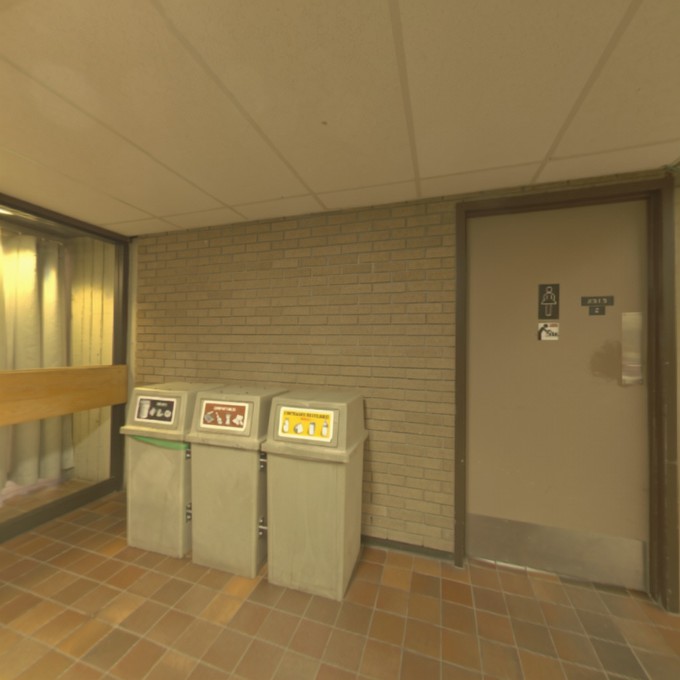}}%
        \hfill
        \includegraphics[width=0.162\linewidth, height=0.162\linewidth]{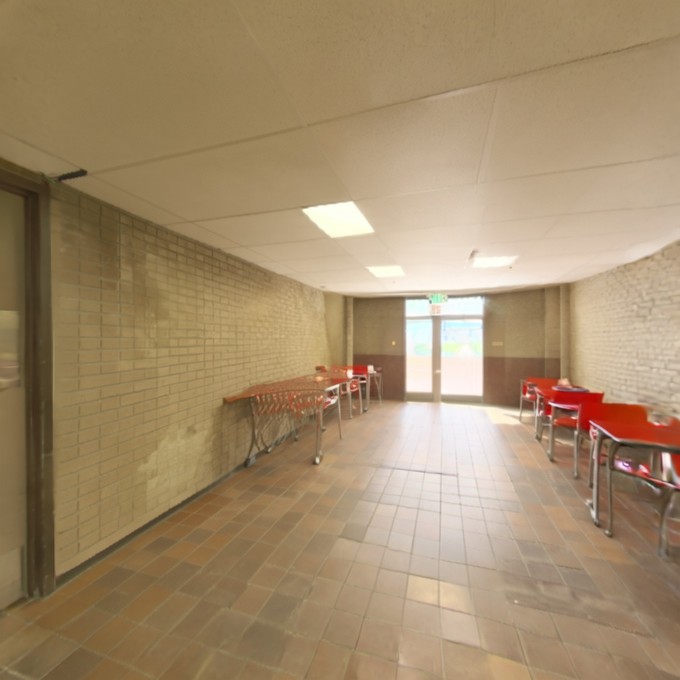}%
        \hfill
        \includegraphics[width=0.162\linewidth, height=0.162\linewidth]{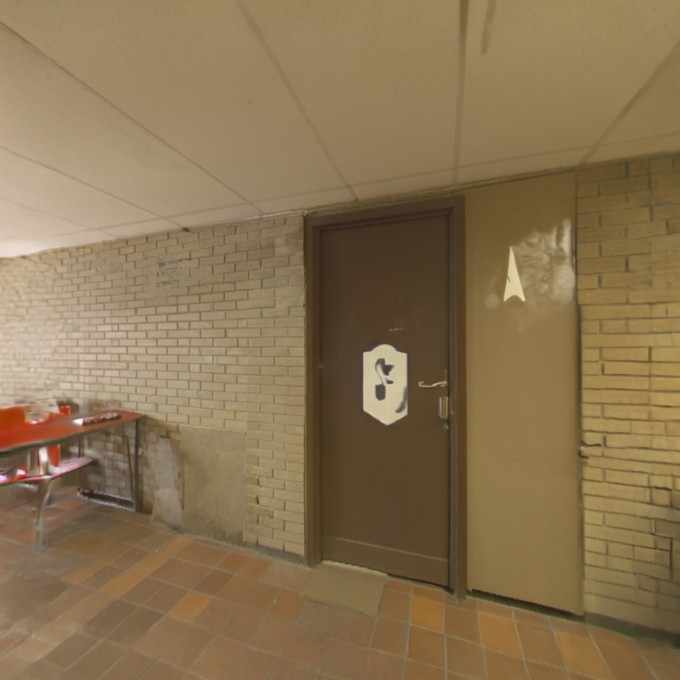}%
        \hfill
        \includegraphics[width=0.162\linewidth, height=0.162\linewidth]{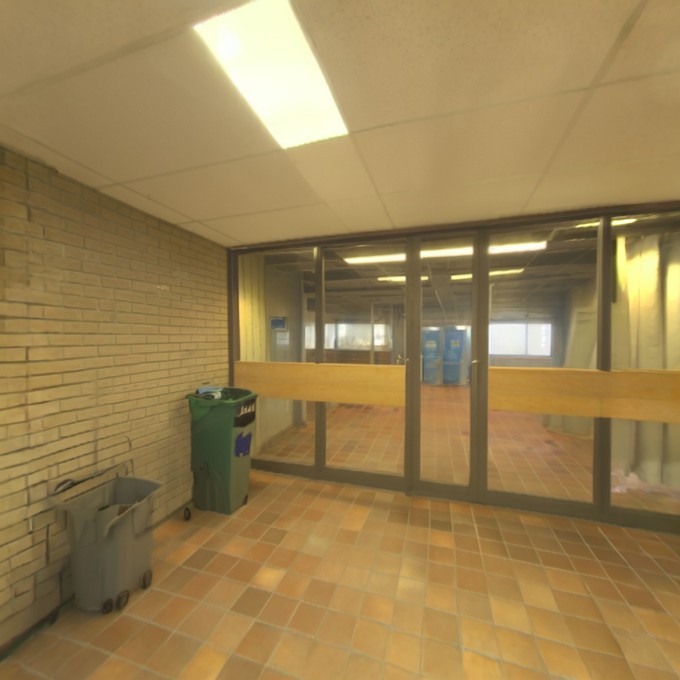}%
        \hfill
        \includegraphics[width=0.162\linewidth, height=0.162\linewidth]{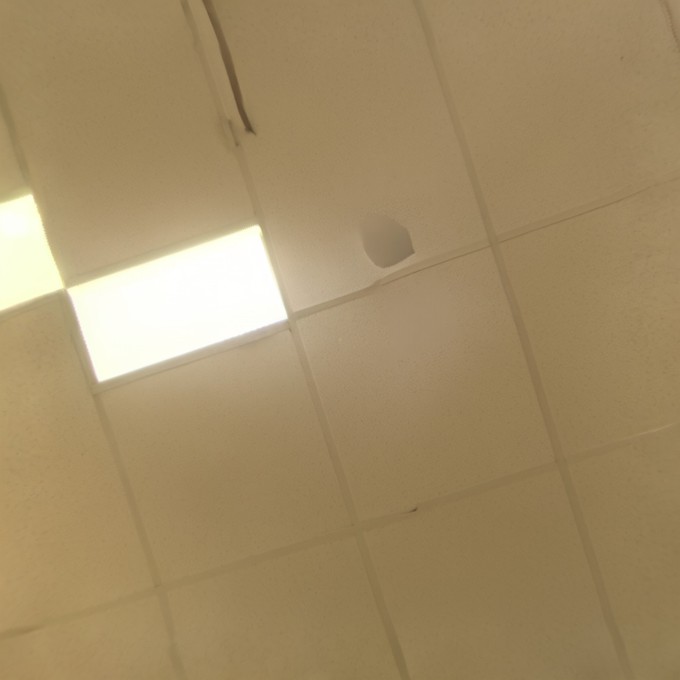}%
        \hfill
        \includegraphics[width=0.162\linewidth, height=0.162\linewidth]{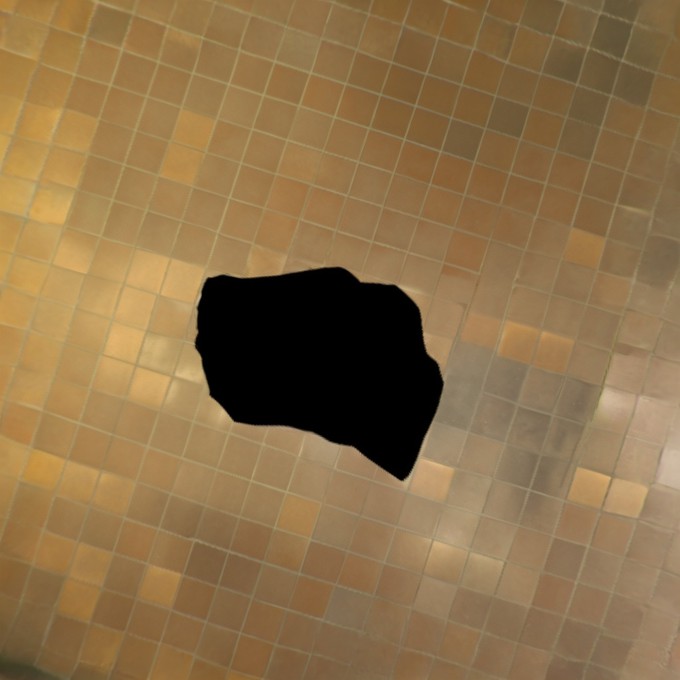}%
    \end{minipage}

    \begin{minipage}[c]{0.997\linewidth}
        \centering
        \includegraphics[width=\linewidth, height=0.5\linewidth]{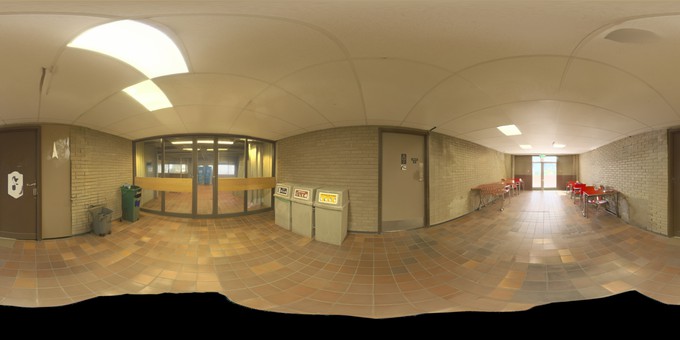}
    \end{minipage}

    \begin{minipage}[c]{0.997\linewidth}
        \centering
        \FramedBox{0.12\linewidth}{0.162\linewidth}{A museum exhibit with a large black and white photo of people working on a dock on the left and a smaller photo of a storefront on the right.}
        \hspace{-0.25cm}
        \FramedBox{0.12\linewidth}{0.162\linewidth}{A museum hallway with exhibits on the left and right, including a large painting of a canal, a sign that says "Erie Canal," and a bench.}
        \hspace{-0.25cm}
        \FramedBox{0.12\linewidth}{0.162\linewidth}{A museum exhibit features a painting of a historical scene with wooden buildings and a pipe, along with other artifacts and informational displays.}
        \hspace{-0.25cm}
        \FramedBox{0.12\linewidth}{0.162\linewidth}{A museum exhibit shows a black and white photo of a factory on the left and a diorama of a factory floor with a treadmill on the right.}
        \hspace{-0.25cm}
        \FramedBox{0.12\linewidth}{0.162\linewidth}{A low-angle view of a dark ceiling with exposed pipes, ductwork, and electrical wiring, illuminated by spotlights and a fluorescent light strip.}
        \hspace{-0.25cm}
        \FramedBox{0.12\linewidth}{0.162\linewidth}{A black hole is silhouetted against a bright accretion disk of swirling gas and dust.}
    \end{minipage}
    \begin{minipage}[c]{0.997\linewidth}
        \centering
        \fboxrule=1pt
        \fcolorbox{green}{green}{\includegraphics[width=0.162\linewidth, height=0.162\linewidth]{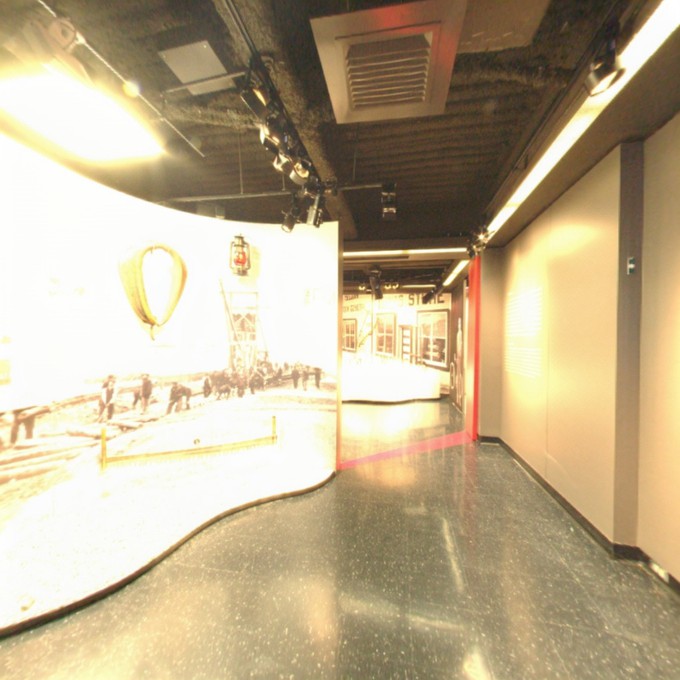}}%
        \hfill
        \includegraphics[width=0.162\linewidth, height=0.162\linewidth]{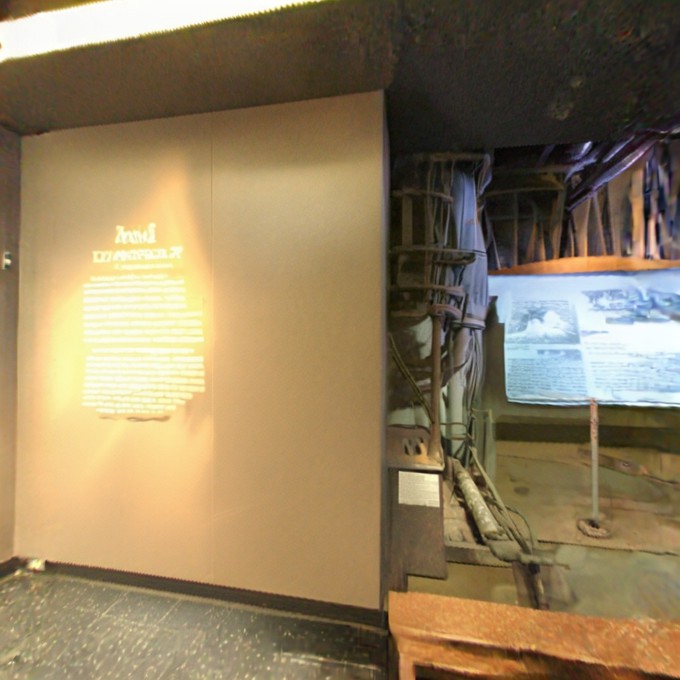}%
        \hfill
        \includegraphics[width=0.162\linewidth, height=0.162\linewidth]{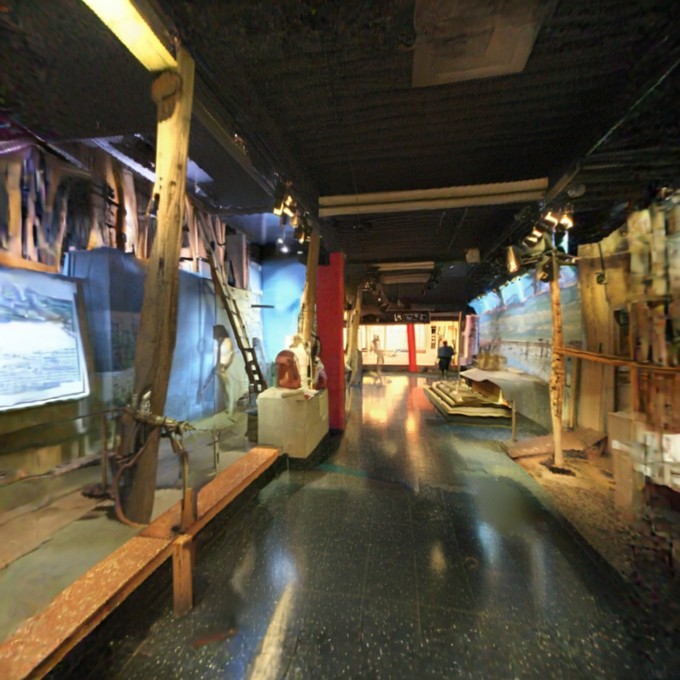}%
        \hfill
        \includegraphics[width=0.162\linewidth, height=0.162\linewidth]{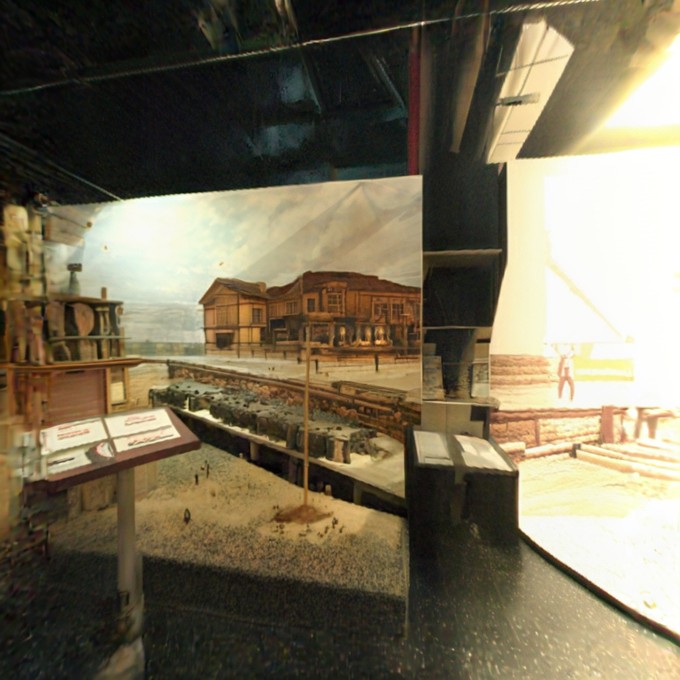}%
        \hfill
        \includegraphics[width=0.162\linewidth, height=0.162\linewidth]{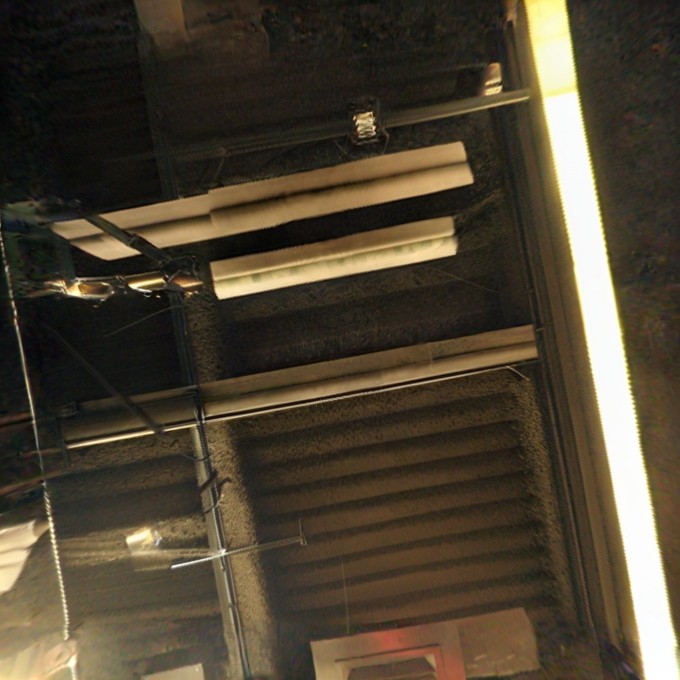}%
        \hfill
        \includegraphics[width=0.162\linewidth, height=0.162\linewidth]{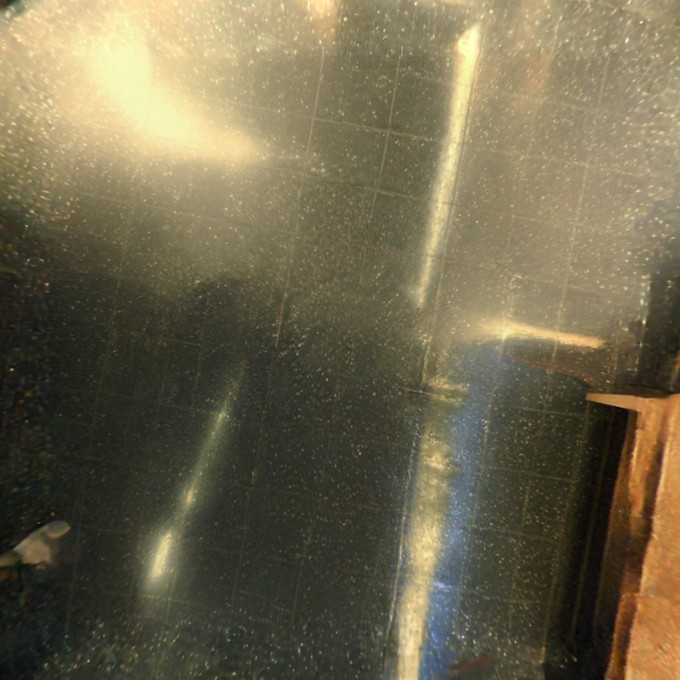}%
    \end{minipage}

    \begin{minipage}[c]{0.997\linewidth}
        \centering
        \includegraphics[width=\linewidth, height=0.5\linewidth]{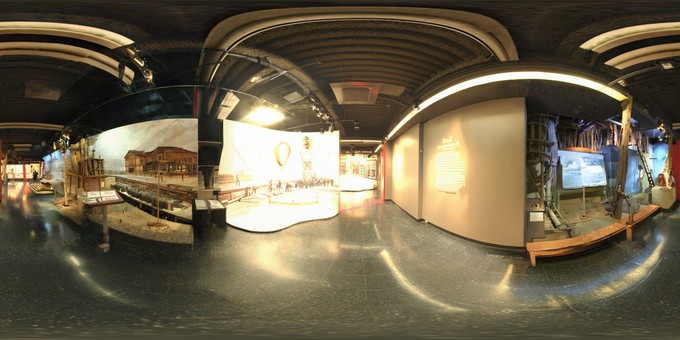}
    \end{minipage}
    \hfill
    \caption{\textbf{Generated panoramas with multiple text prompts and image condition}}
    \label{fig:mulittexttwo}
\end{figure}

We provide more qualtitative results. In \Cref{fig:mulittext,fig:mulittexttwo}, we show the input modalities in green and the generated faces and panorama. We see that our model aligns well to the given text prompts.

\newpage
\subsection{More results of unsynchronized GroupNorm}

\begin{figure}[!ht]
    \centering
    \includegraphics[width=0.49\linewidth]{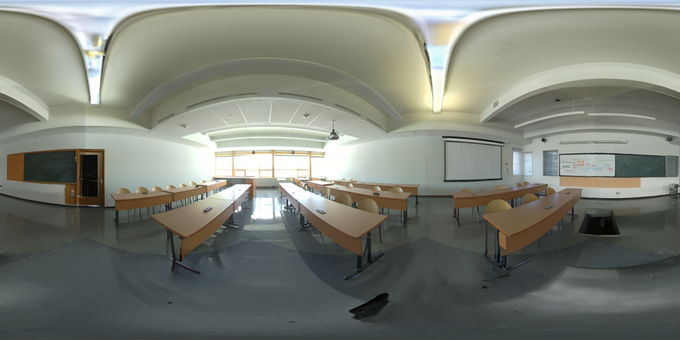}
    \includegraphics[width=0.49\linewidth]{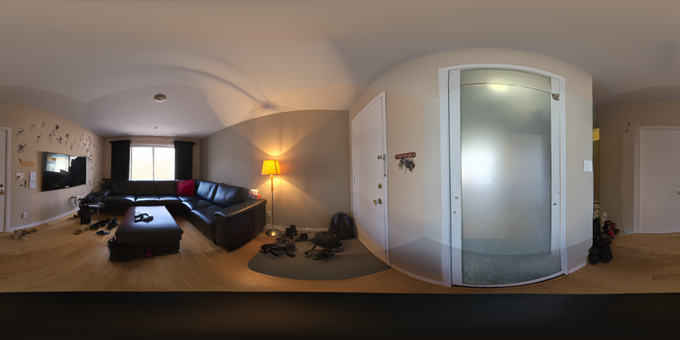}
    \includegraphics[width=0.49\linewidth]{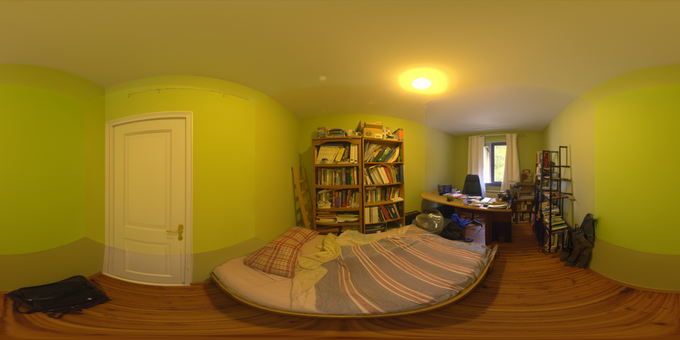}
    \includegraphics[width=0.49\linewidth]{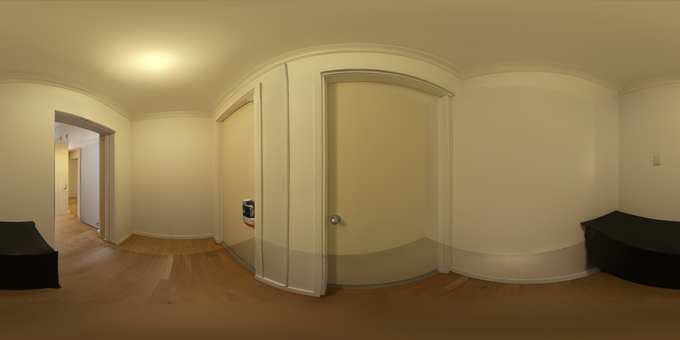}
    \includegraphics[width=0.49\linewidth]{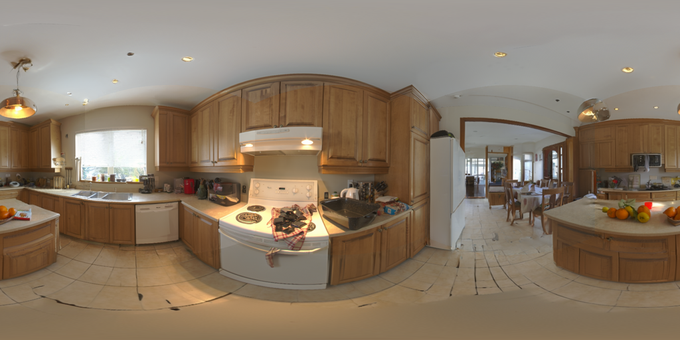}
    \includegraphics[width=0.49\linewidth]{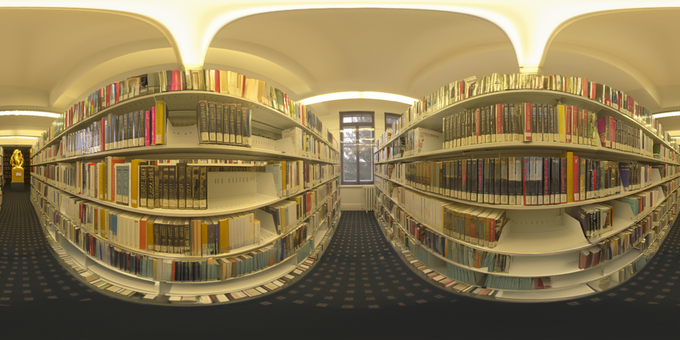}
    \includegraphics[width=0.49\linewidth]{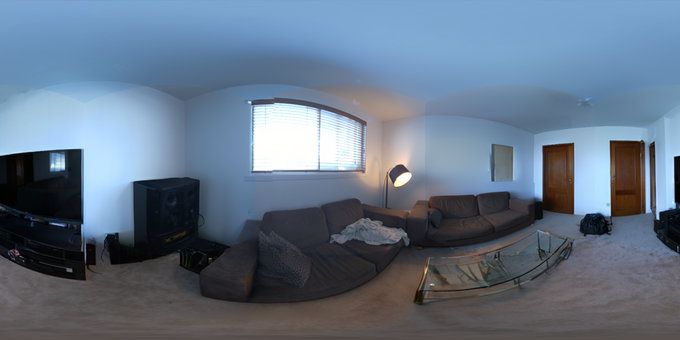}
    \includegraphics[width=0.49\linewidth]{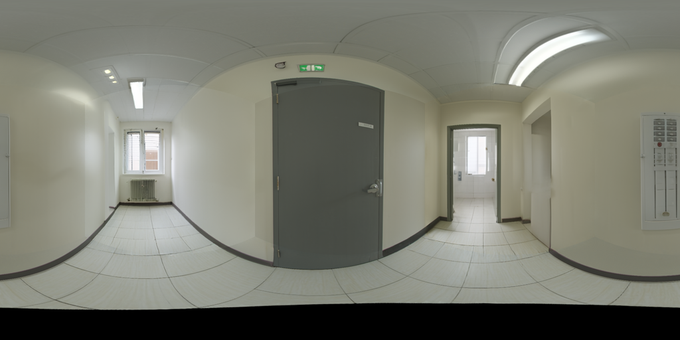}
    \includegraphics[width=0.49\linewidth]{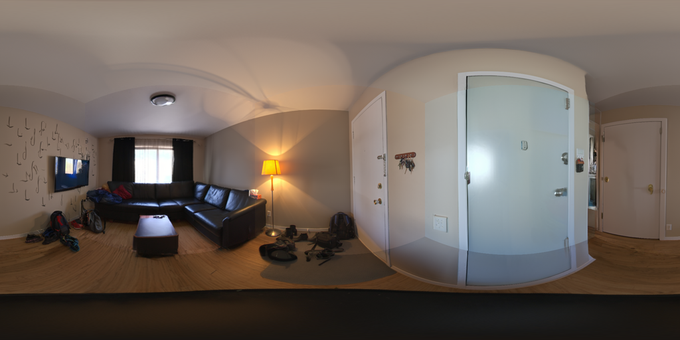}
    \includegraphics[width=0.49\linewidth]{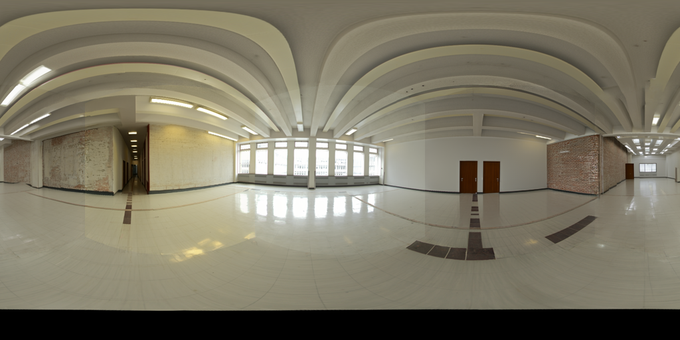}
    \includegraphics[width=0.49\linewidth]{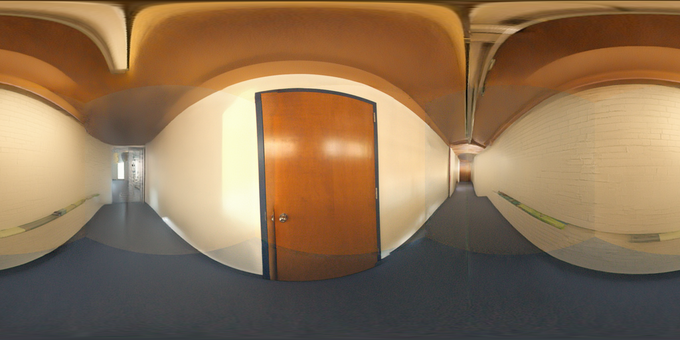}
    \includegraphics[width=0.49\linewidth]{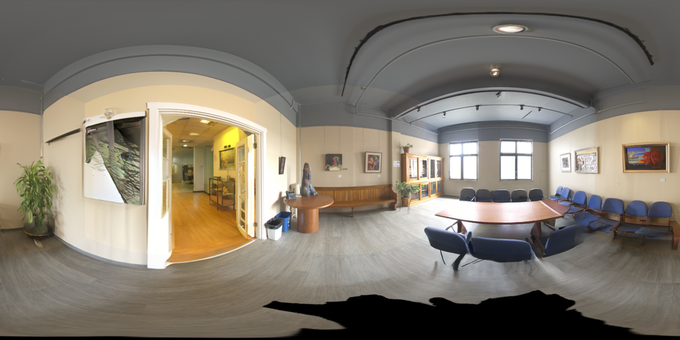}
    \caption{\textbf{Additional results of predictions with unsynchronized Group Norm.}}
    \label{fig:appendix:unsyncedgn}
\end{figure}

\newpage
\subsection{More results of non-overlapping predictions}

\begin{figure}[!ht]
    \centering
    \includegraphics[width=0.49\linewidth]{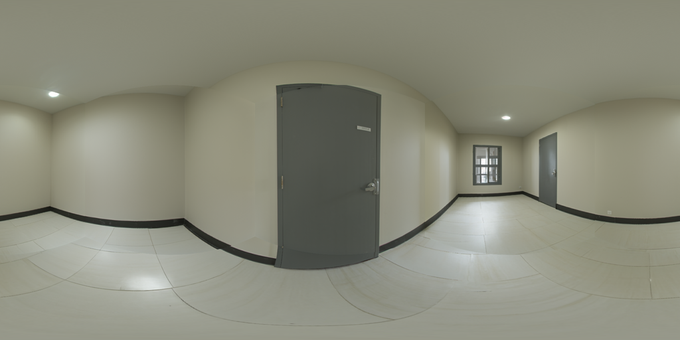}
    \includegraphics[width=0.49\linewidth]{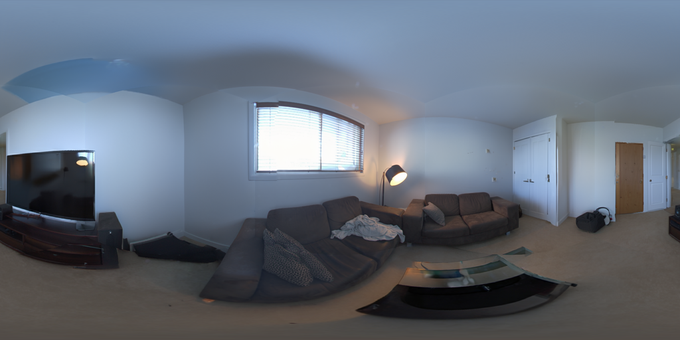}
    \includegraphics[width=0.49\linewidth]{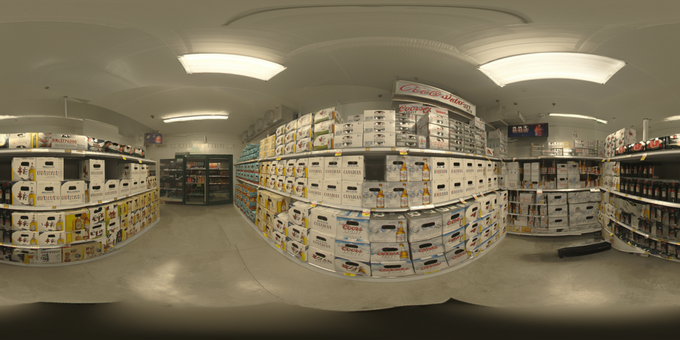}
    \includegraphics[width=0.49\linewidth]{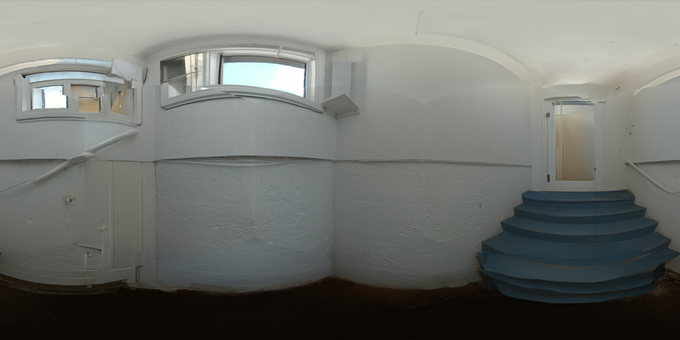}
    \includegraphics[width=0.49\linewidth]{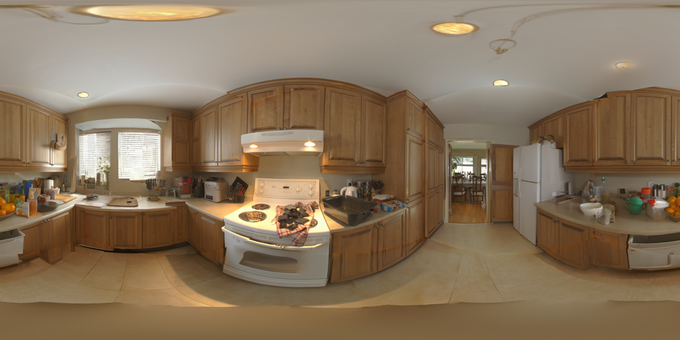}
    \includegraphics[width=0.49\linewidth]{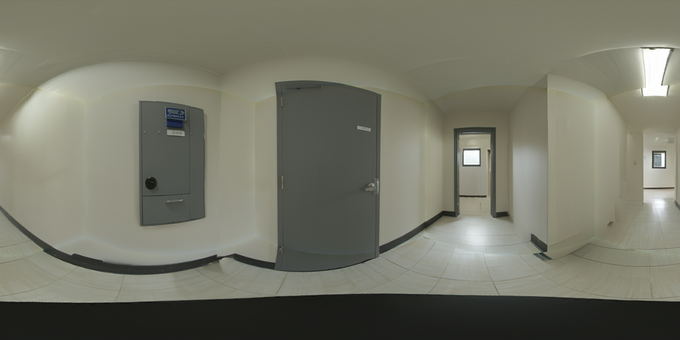}
    \includegraphics[width=0.49\linewidth]{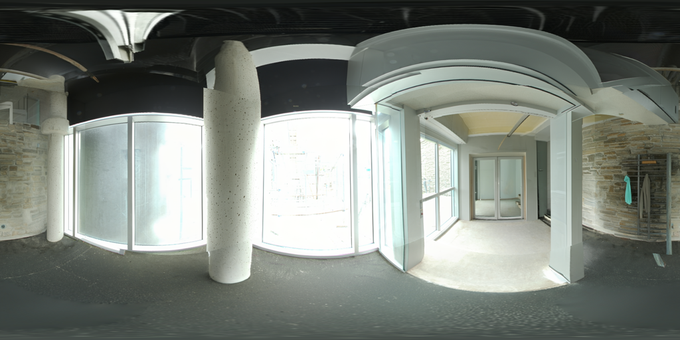}
    \includegraphics[width=0.49\linewidth]{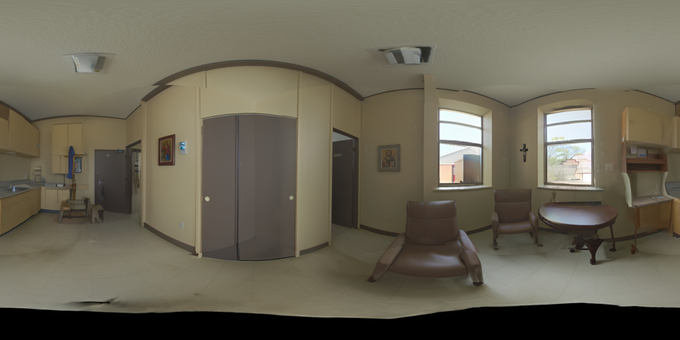}
    \includegraphics[width=0.49\linewidth]{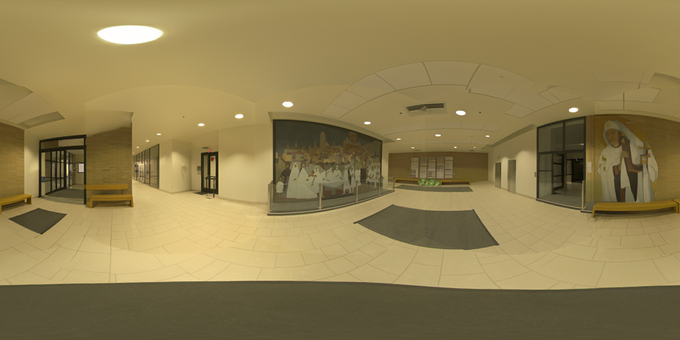}
    \includegraphics[width=0.49\linewidth]{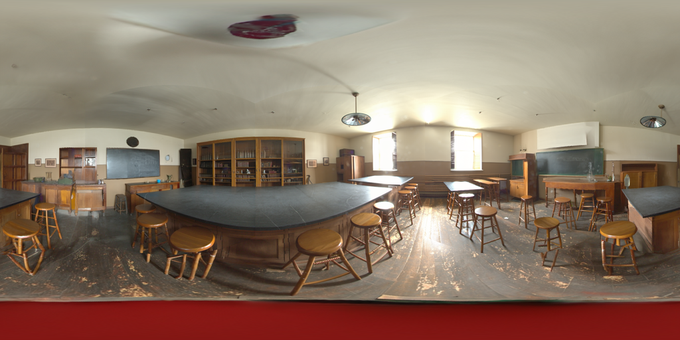}
    \includegraphics[width=0.49\linewidth]{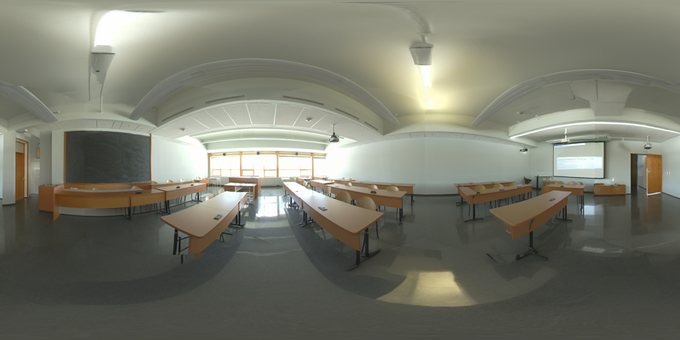}
    \includegraphics[width=0.49\linewidth]{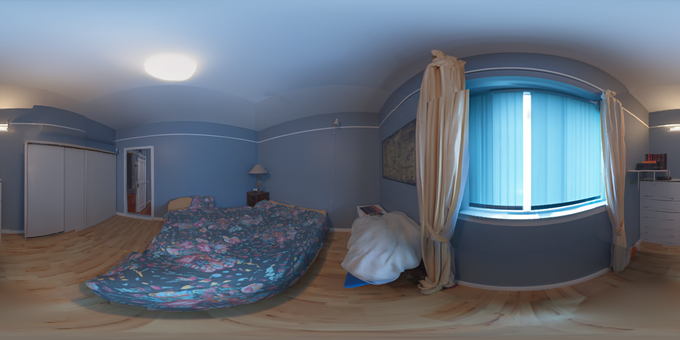}
    \caption{\textbf{Additional results of non-overlapping predictions.}}
    \label{fig:appendix:overlap}
\end{figure}

\newpage

\subsection{Individual face overlaps from qualitative comparison}
Here, we depict the individual faces generated by our three CubeDiff methods used in the equirectangular panoramas in Figure 4 in the main paper. We show both uncropped and cropped faces as requested by reviewers. Additionally, we show the ground truth panoramas, the individual textual face descriptions. The corresponding input conditioning image is always the first (front) image of the individual faces (and thus equal for all models). 

\paragraph{Ours\textsubscript{img}} \phantom{a}
\begin{figure}[!ht]
    \centering
    \begin{overpic}[width=0.162\textwidth]{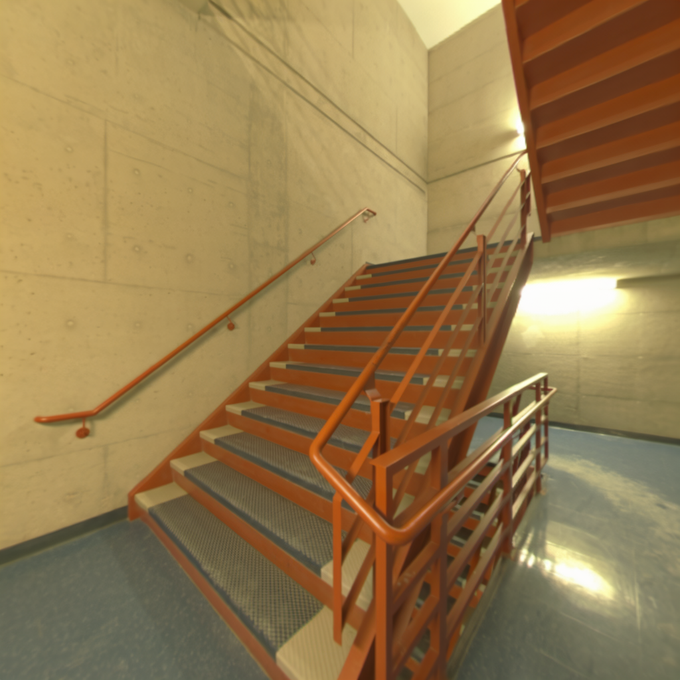}
    \put(6.2,6.2){\color{green}\framebox(87.6, 87.6){}}
    \end{overpic}\!\!
    \hfill
    \begin{overpic}[width=0.162\textwidth]{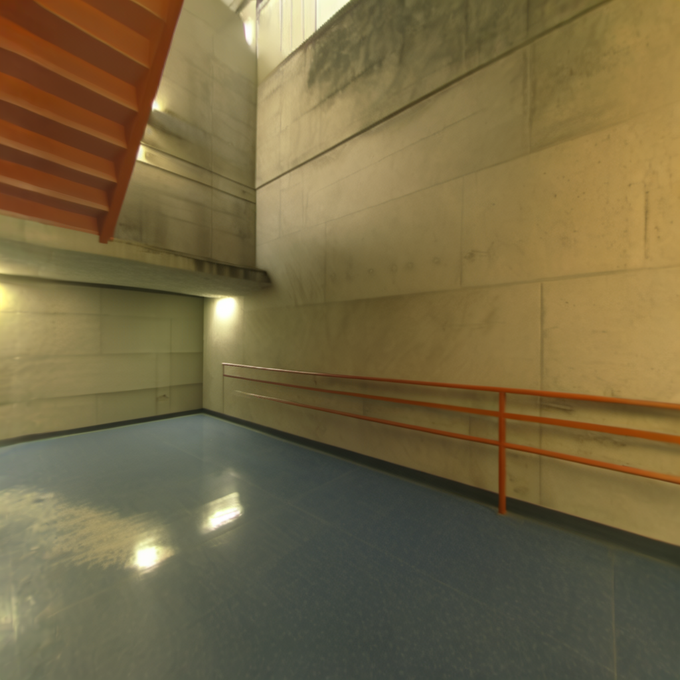}
    \put(6.2,6.2){\color{green}\framebox(87.6, 87.6){}}
    \end{overpic}\!\!
    \hfill
    \begin{overpic}[width=0.162\textwidth]{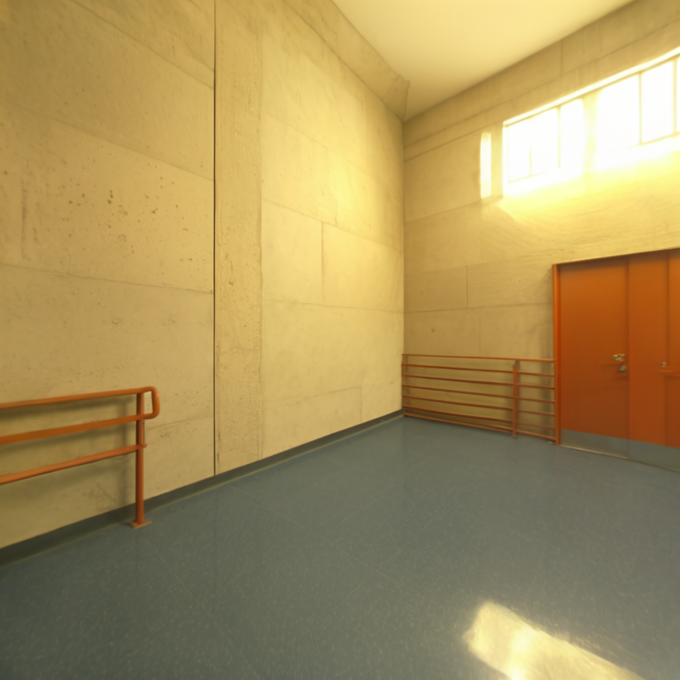}
    \put(6.2,6.2){\color{green}\framebox(87.6, 87.6){}}
    \end{overpic}\!\!
    \hfill
    \begin{overpic}[width=0.162\textwidth]{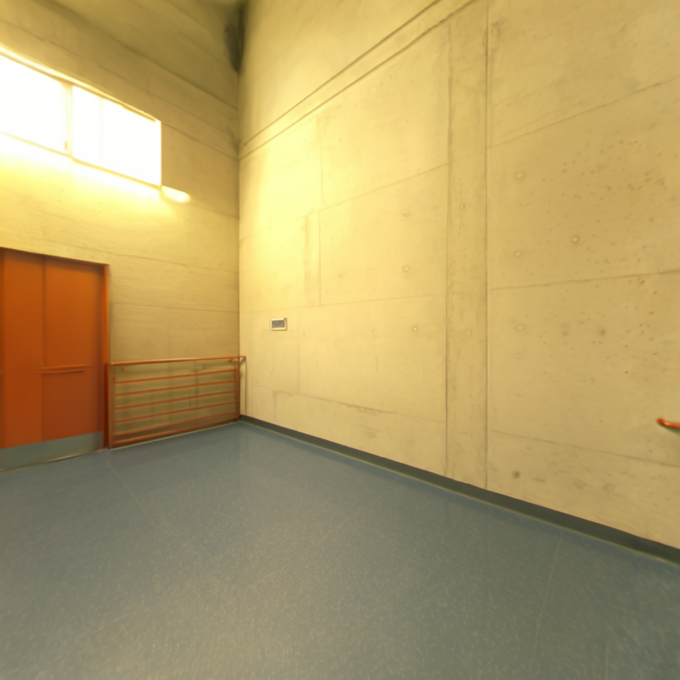}
    \put(6.2,6.2){\color{green}\framebox(87.6, 87.6){}}
    \end{overpic}\!\!
    \hfill
    \begin{overpic}[width=0.162\textwidth]{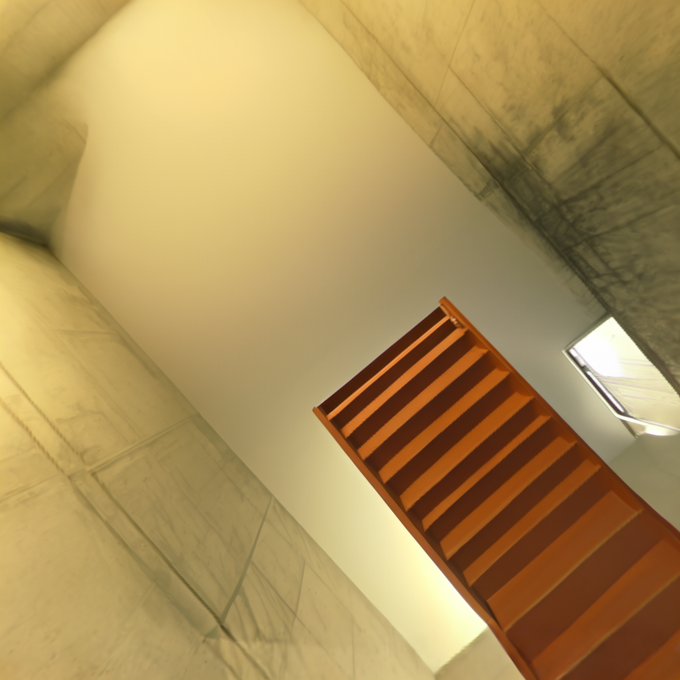}
    \put(6.2,6.2){\color{green}\framebox(87.6, 87.6){}}
    \end{overpic}\!\!
    \hfill
    \begin{overpic}[width=0.162\textwidth]{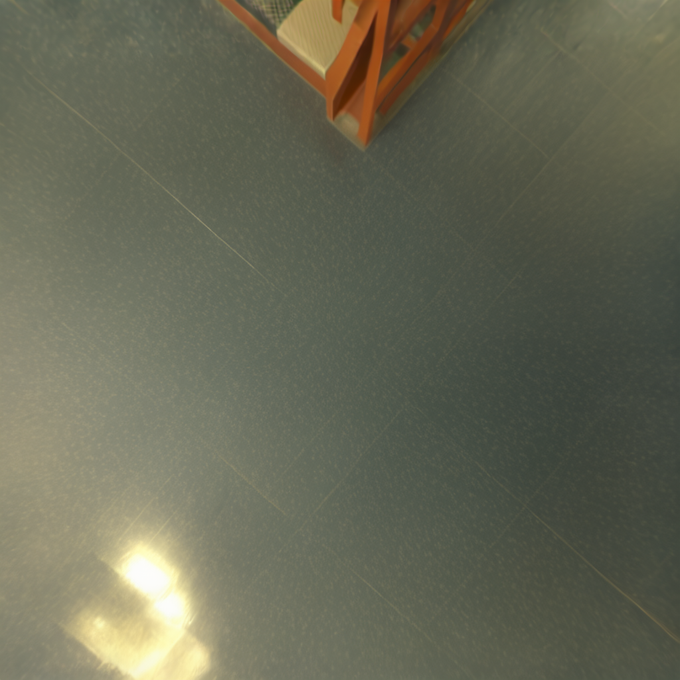}
    \put(6.2,6.2){\color{green}\framebox(87.6, 87.6){}}
    \end{overpic}
    \includegraphics[width=0.162\linewidth,clip,trim=43px 43px 43px 43px]{figures/qualitativecomp/faces/notext/uncropped/9C4A0076-46c51ec2c2/0_0.png}%
    \hfill
    \includegraphics[width=0.162\linewidth,clip,trim=43px 43px 43px 43px]{figures/qualitativecomp/faces/notext/uncropped/9C4A0076-46c51ec2c2/0_1.png}%
    \hfill
    \includegraphics[width=0.162\linewidth,clip,trim=43px 43px 43px 43px]{figures/qualitativecomp/faces/notext/uncropped/9C4A0076-46c51ec2c2/0_2.png}%
    \hfill
    \includegraphics[width=0.162\linewidth,clip,trim=43px 43px 43px 43px]{figures/qualitativecomp/faces/notext/uncropped/9C4A0076-46c51ec2c2/0_3.png}%
    \hfill
    \includegraphics[width=0.162\linewidth,clip,trim=43px 43px 43px 43px]{figures/qualitativecomp/faces/notext/uncropped/9C4A0076-46c51ec2c2/0_4.png}%
    \hfill
    \includegraphics[width=0.162\linewidth,clip,trim=43px 43px 43px 43px]{figures/qualitativecomp/faces/notext/uncropped/9C4A0076-46c51ec2c2/0_5.png}
    \caption{Our generated faces with input conditioning image only. Top row shows the uncropped faces, bottom row shows the cropped faces.}
\end{figure}

\begin{figure}[!ht]
    \centering
    \begin{overpic}[width=0.162\textwidth]{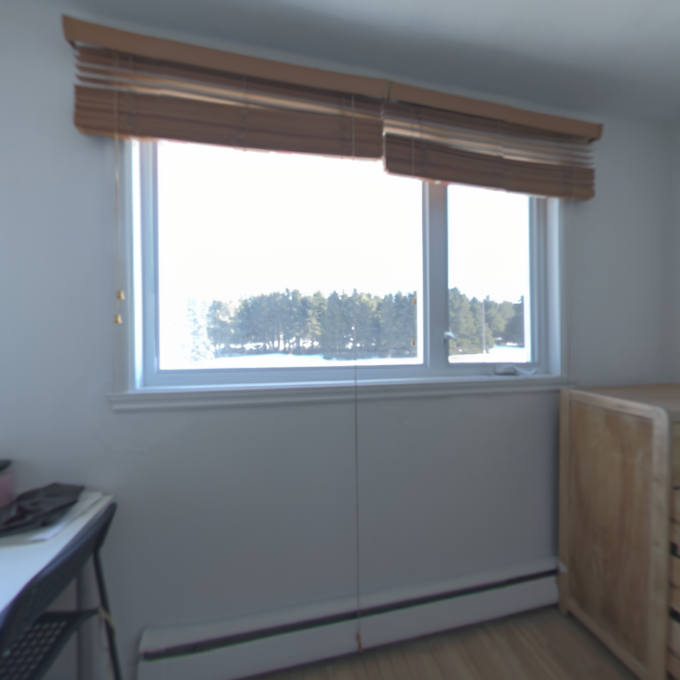}
    \put(6.2,6.2){\color{green}\framebox(87.6, 87.6){}}
    \end{overpic}\!\!
    \hfill
    \begin{overpic}[width=0.162\textwidth]{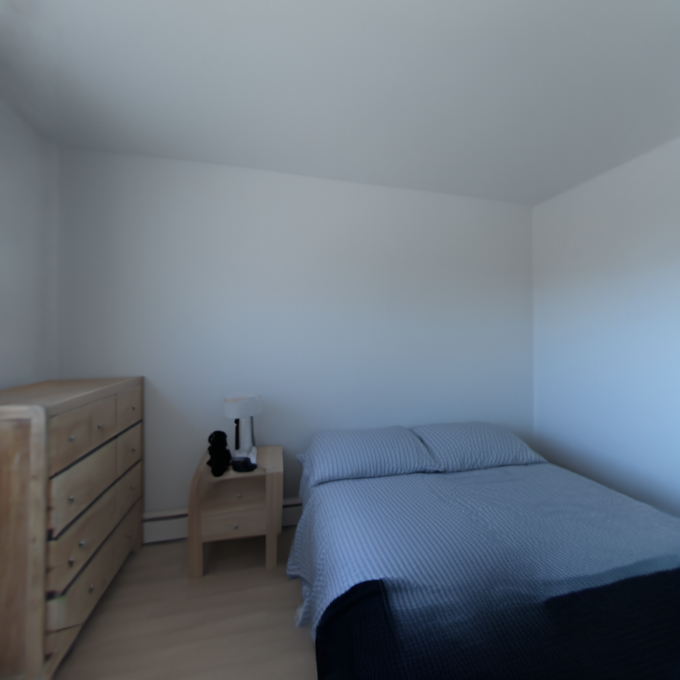}
    \put(6.2,6.2){\color{green}\framebox(87.6, 87.6){}}
    \end{overpic}\!\!
    \hfill
    \begin{overpic}[width=0.162\textwidth]{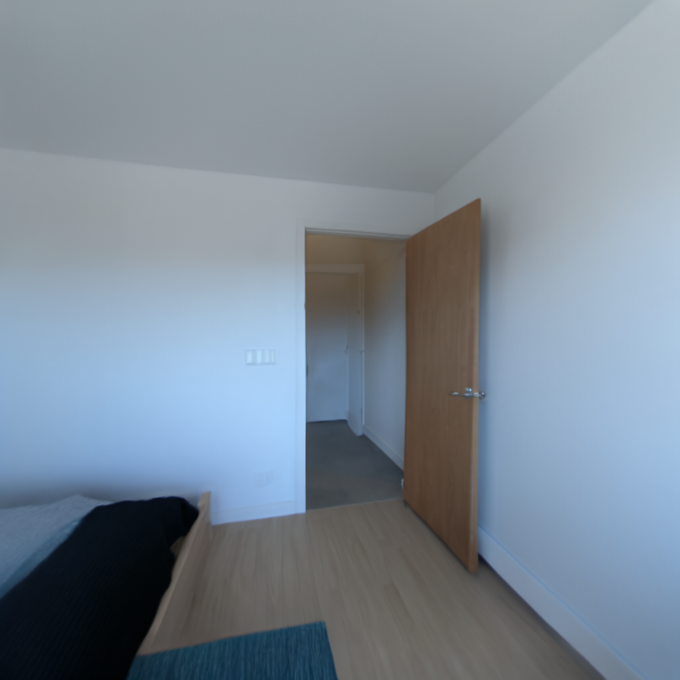}
    \put(6.2,6.2){\color{green}\framebox(87.6, 87.6){}}
    \end{overpic}\!\!
    \hfill
    \begin{overpic}[width=0.162\textwidth]{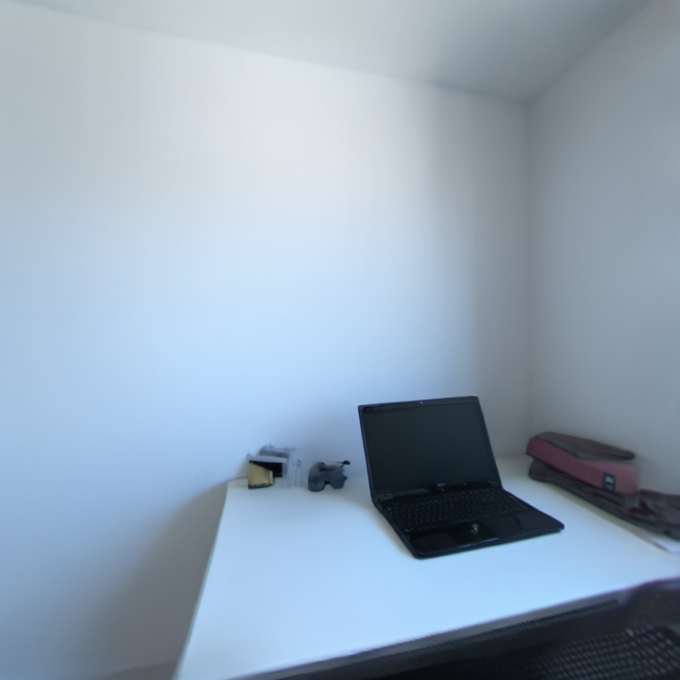}
    \put(6.2,6.2){\color{green}\framebox(87.6, 87.6){}}
    \end{overpic}\!\!
    \hfill
    \begin{overpic}[width=0.162\textwidth]{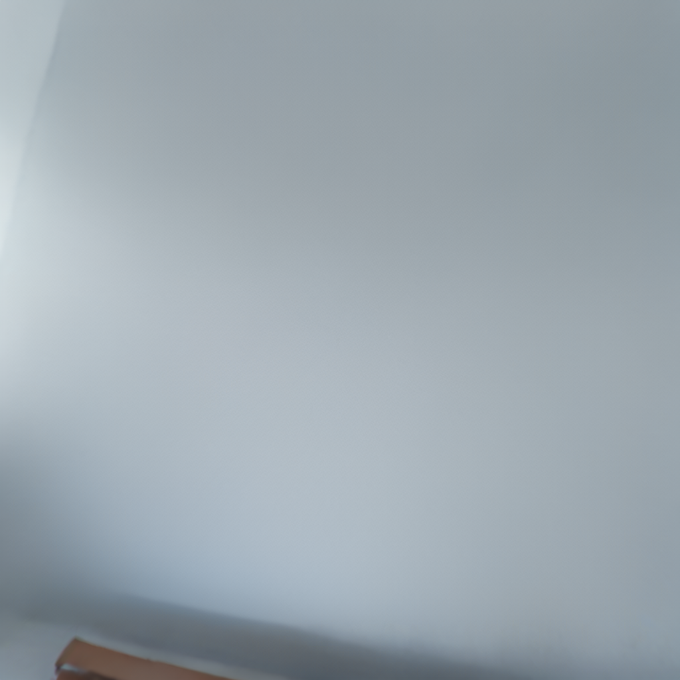}
    \put(6.2,6.2){\color{green}\framebox(87.6, 87.6){}}
    \end{overpic}\!\!
    \hfill
    \begin{overpic}[width=0.162\textwidth]{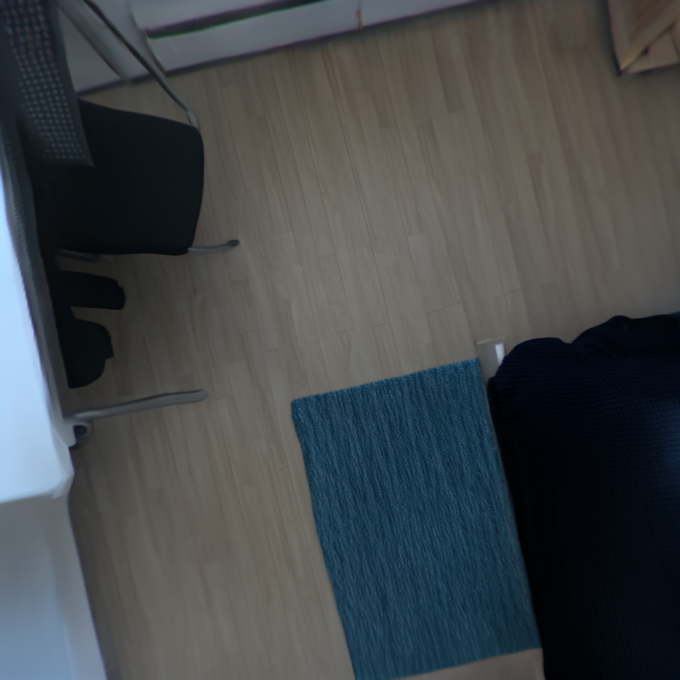}
    \put(6.2,6.2){\color{green}\framebox(87.6, 87.6){}}
    \end{overpic}
    \includegraphics[width=0.162\linewidth,clip,trim=43px 43px 43px 43px]{figures/qualitativecomp/faces/notext/uncropped/9C4A0330-55d4beffc9/0_0.png}%
    \hfill
    \includegraphics[width=0.162\linewidth,clip,trim=43px 43px 43px 43px]{figures/qualitativecomp/faces/notext/uncropped/9C4A0330-55d4beffc9/0_1.png}%
    \hfill
    \includegraphics[width=0.162\linewidth,clip,trim=43px 43px 43px 43px]{figures/qualitativecomp/faces/notext/uncropped/9C4A0330-55d4beffc9/0_2.png}%
    \hfill
    \includegraphics[width=0.162\linewidth,clip,trim=43px 43px 43px 43px]{figures/qualitativecomp/faces/notext/uncropped/9C4A0330-55d4beffc9/0_3.png}%
    \hfill
    \includegraphics[width=0.162\linewidth,clip,trim=43px 43px 43px 43px]{figures/qualitativecomp/faces/notext/uncropped/9C4A0330-55d4beffc9/0_4.png}%
    \hfill
    \includegraphics[width=0.162\linewidth,clip,trim=43px 43px 43px 43px]{figures/qualitativecomp/faces/notext/uncropped/9C4A0330-55d4beffc9/0_5.png}
    \caption{Our generated faces with input conditioning image. Top row shows the uncropped faces, bottom row shows the cropped faces.}
    
\end{figure}
\newpage
\paragraph{Ours\textsubscript{img+txt}}
\begin{itemize}
    \item A concrete stairwell with orange railings leads up to a yellow door with a number 2 on it.
\end{itemize}
\begin{figure}[!ht]
    \centering
    \begin{overpic}[width=0.162\textwidth]{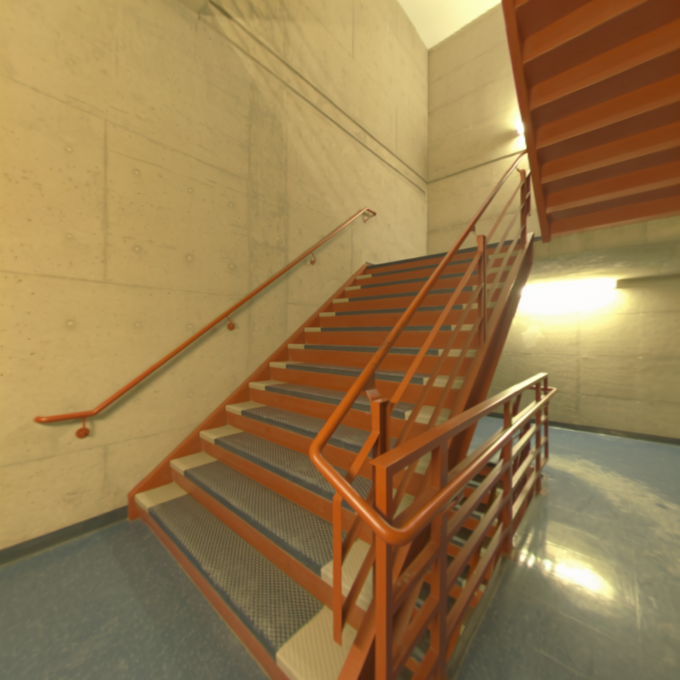}
    \put(6.2,6.2){\color{green}\framebox(87.6, 87.6){}}
    \end{overpic}\!\!
    \hfill
    \begin{overpic}[width=0.162\textwidth]{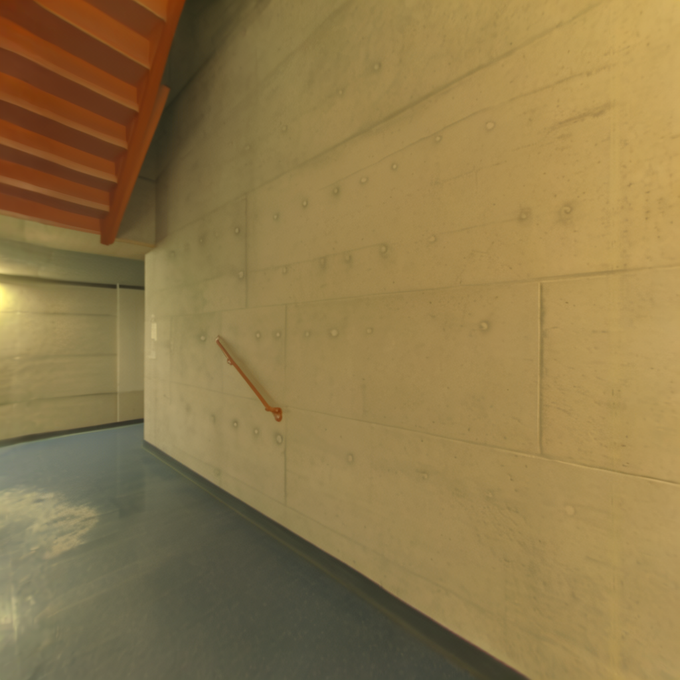}
    \put(6.2,6.2){\color{green}\framebox(87.6, 87.6){}}
    \end{overpic}\!\!
    \hfill
    \begin{overpic}[width=0.162\textwidth]{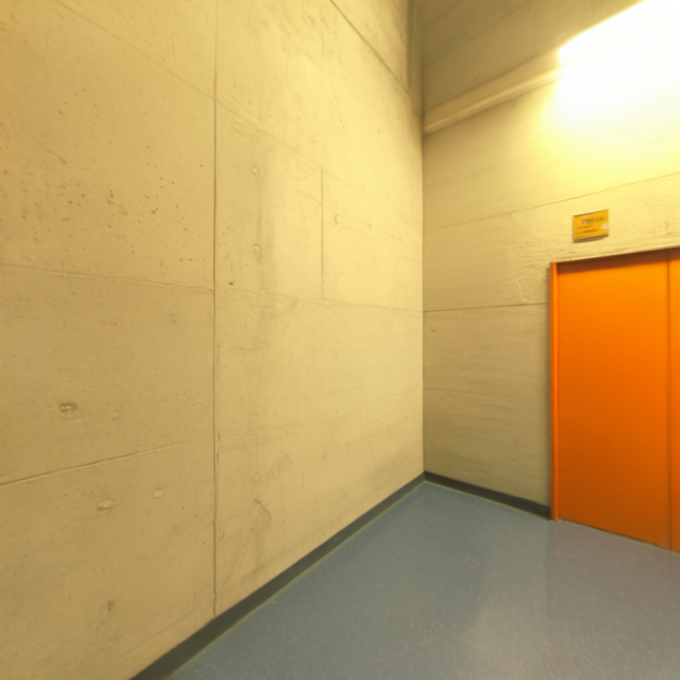}
    \put(6.2,6.2){\color{green}\framebox(87.6, 87.6){}}
    \end{overpic}\!\!
    \hfill
    \begin{overpic}[width=0.162\textwidth]{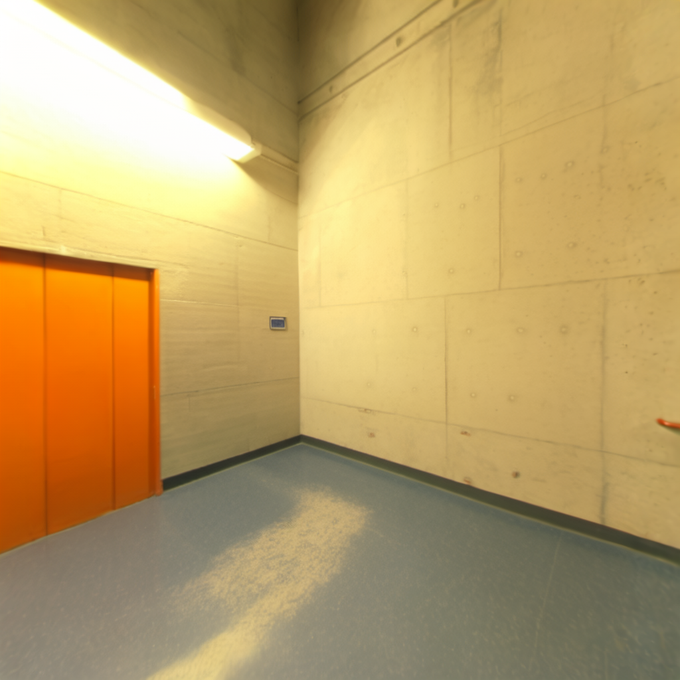}
    \put(6.2,6.2){\color{green}\framebox(87.6, 87.6){}}
    \end{overpic}\!\!
    \hfill
    \begin{overpic}[width=0.162\textwidth]{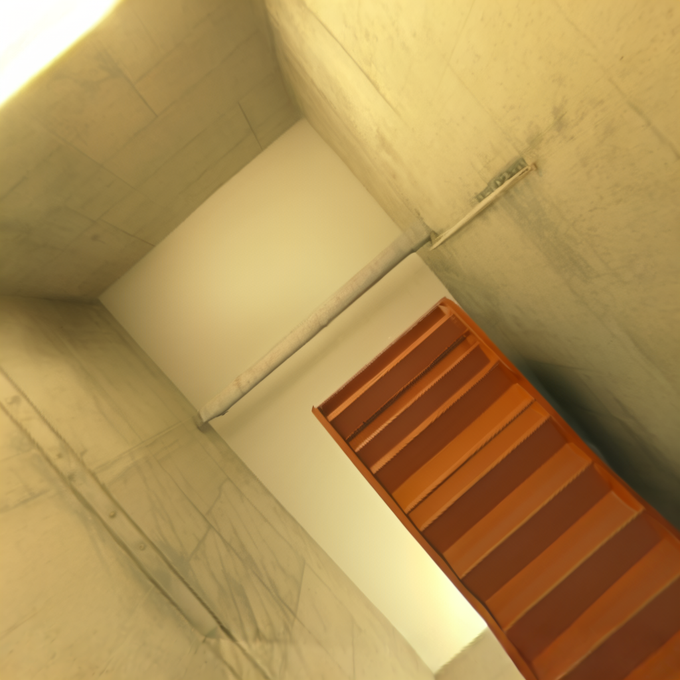}
    \put(6.2,6.2){\color{green}\framebox(87.6, 87.6){}}
    \end{overpic}\!\!
    \hfill
    \begin{overpic}[width=0.162\textwidth]{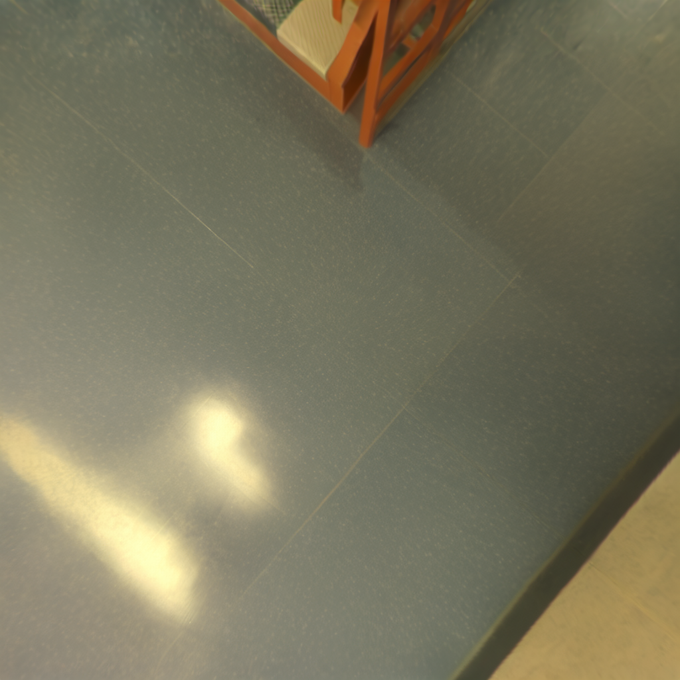}
    \put(6.2,6.2){\color{green}\framebox(87.6, 87.6){}}
    \end{overpic}
    \includegraphics[width=0.162\linewidth,clip,trim=43px 43px 43px 43px]{figures/qualitativecomp/faces/single/uncropped/9C4A0076-46c51ec2c2/0_0.png}%
    \hfill
    \includegraphics[width=0.162\linewidth,clip,trim=43px 43px 43px 43px]{figures/qualitativecomp/faces/single/uncropped/9C4A0076-46c51ec2c2/0_1.png}%
    \hfill
    \includegraphics[width=0.162\linewidth,clip,trim=43px 43px 43px 43px]{figures/qualitativecomp/faces/single/uncropped/9C4A0076-46c51ec2c2/0_2.png}%
    \hfill
    \includegraphics[width=0.162\linewidth,clip,trim=43px 43px 43px 43px]{figures/qualitativecomp/faces/single/uncropped/9C4A0076-46c51ec2c2/0_3.png}%
    \hfill
    \includegraphics[width=0.162\linewidth,clip,trim=43px 43px 43px 43px]{figures/qualitativecomp/faces/single/uncropped/9C4A0076-46c51ec2c2/0_4.png}%
    \hfill
    \includegraphics[width=0.162\linewidth,clip,trim=43px 43px 43px 43px]{figures/qualitativecomp/faces/single/uncropped/9C4A0076-46c51ec2c2/0_5.png}
    \caption{Our generated faces with single caption input. Top row shows the uncropped faces, bottom row shows the cropped faces.}
\end{figure}

\begin{itemize}
    \item A bedroom with a window overlooking a snowy forest, a bed, a desk, and a dresser.
\end{itemize}

\begin{figure}[!ht]
    \centering
    \begin{overpic}[width=0.162\textwidth]{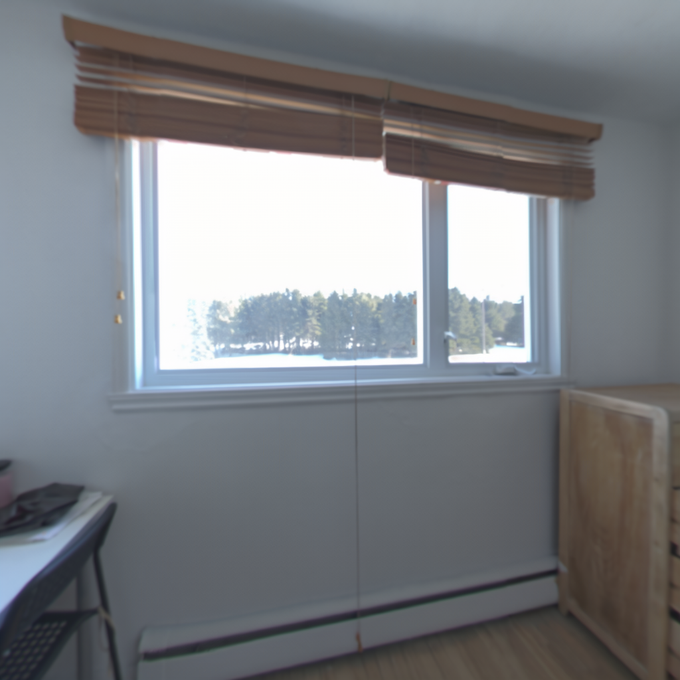}
    \put(6.2,6.2){\color{green}\framebox(87.6, 87.6){}}
    \end{overpic}\!\!
    \hfill
    \begin{overpic}[width=0.162\textwidth]{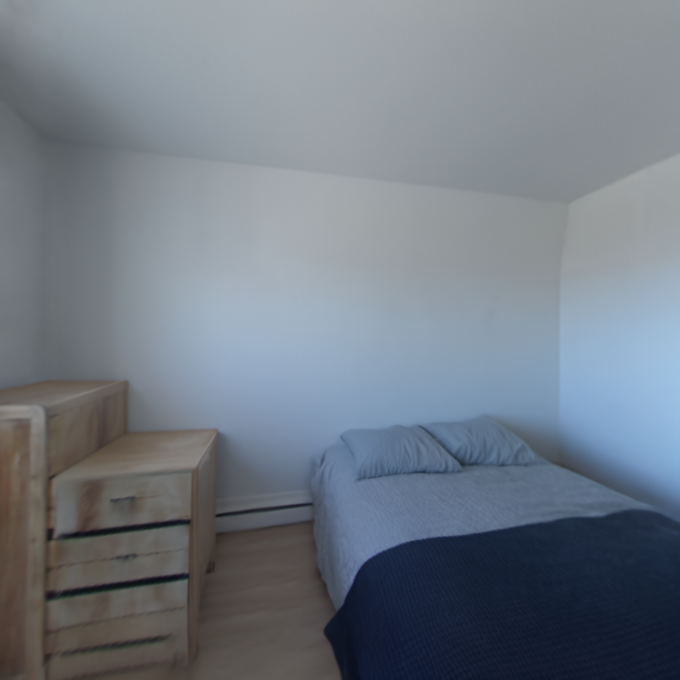}
    \put(6.2,6.2){\color{green}\framebox(87.6, 87.6){}}
    \end{overpic}\!\!
    \hfill
    \begin{overpic}[width=0.162\textwidth]{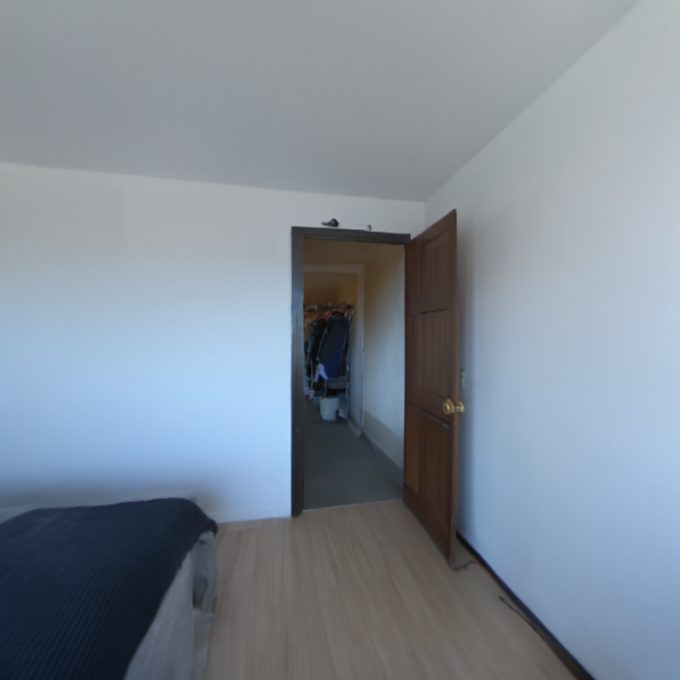}
    \put(6.2,6.2){\color{green}\framebox(87.6, 87.6){}}
    \end{overpic}\!\!
    \hfill
    \begin{overpic}[width=0.162\textwidth]{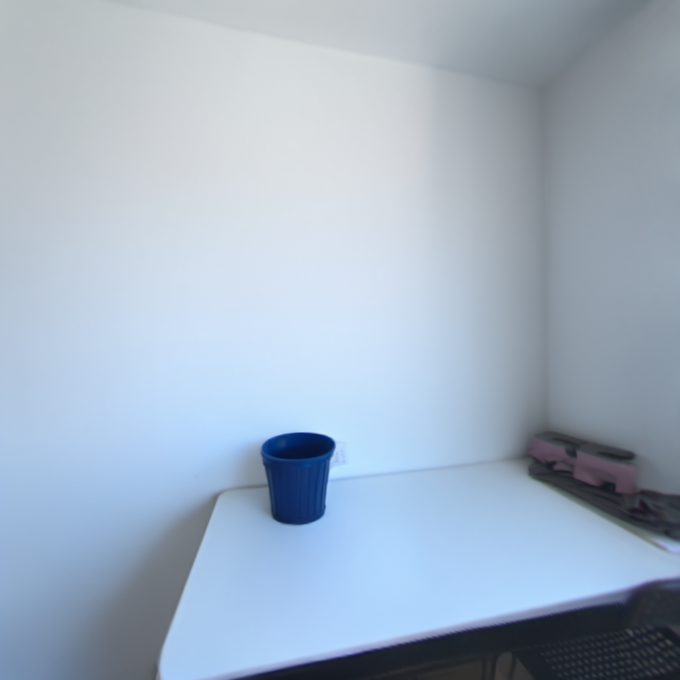}
    \put(6.2,6.2){\color{green}\framebox(87.6, 87.6){}}
    \end{overpic}\!\!
    \hfill
    \begin{overpic}[width=0.162\textwidth]{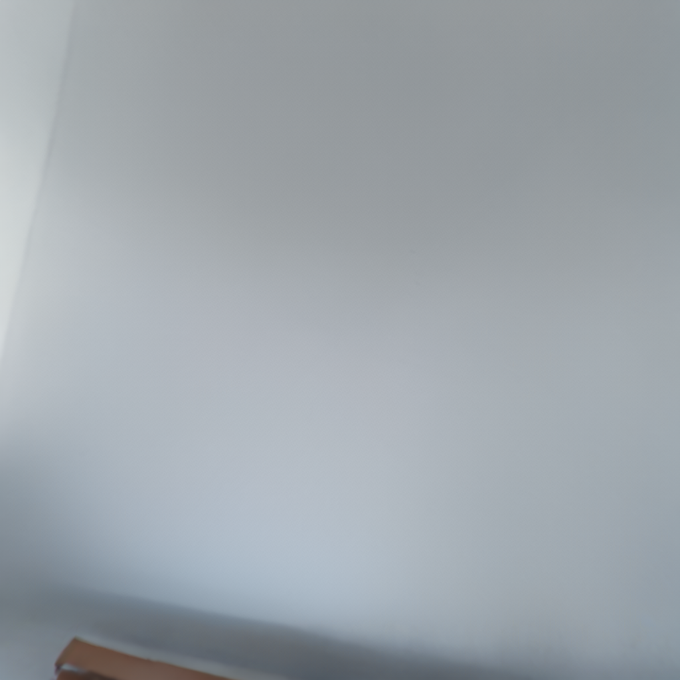}
    \put(6.2,6.2){\color{green}\framebox(87.6, 87.6){}}
    \end{overpic}\!\!
    \hfill
    \begin{overpic}[width=0.162\textwidth]{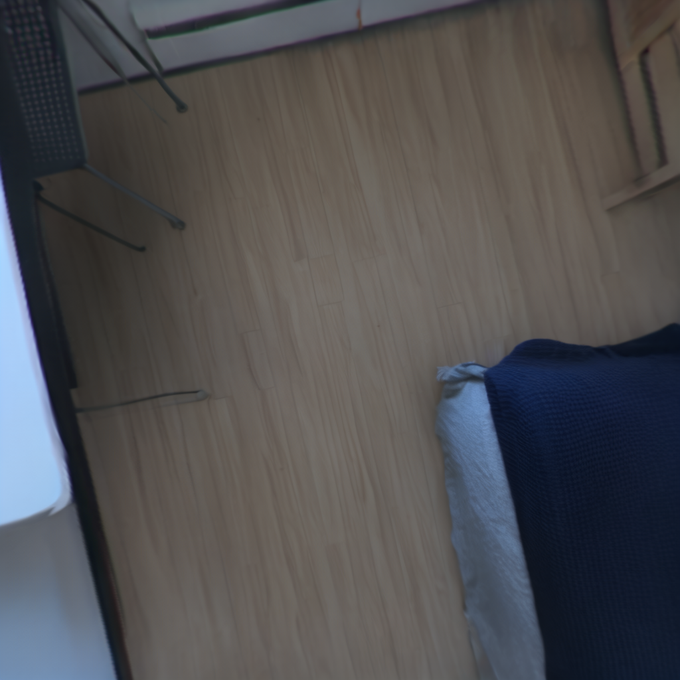}
    \put(6.2,6.2){\color{green}\framebox(87.6, 87.6){}}
    \end{overpic}
    \includegraphics[width=0.162\linewidth,clip,trim=43px 43px 43px 43px]{figures/qualitativecomp/faces/single/uncropped/9C4A0330-55d4beffc9/0_0.png}%
    \hfill
    \includegraphics[width=0.162\linewidth,clip,trim=43px 43px 43px 43px]{figures/qualitativecomp/faces/single/uncropped/9C4A0330-55d4beffc9/0_1.png}%
    \hfill
    \includegraphics[width=0.162\linewidth,clip,trim=43px 43px 43px 43px]{figures/qualitativecomp/faces/single/uncropped/9C4A0330-55d4beffc9/0_2.png}%
    \hfill
    \includegraphics[width=0.162\linewidth,clip,trim=43px 43px 43px 43px]{figures/qualitativecomp/faces/single/uncropped/9C4A0330-55d4beffc9/0_3.png}%
    \hfill
    \includegraphics[width=0.162\linewidth,clip,trim=43px 43px 43px 43px]{figures/qualitativecomp/faces/single/uncropped/9C4A0330-55d4beffc9/0_4.png}%
    \hfill
    \includegraphics[width=0.162\linewidth,clip,trim=43px 43px 43px 43px]{figures/qualitativecomp/faces/single/uncropped/9C4A0330-55d4beffc9/0_5.png}
    \caption{Our generated faces with single caption input. Top row shows the uncropped faces, bottom row shows the cropped faces.}
    
\end{figure}

\newpage

\paragraph{Ours\textsubscript{img+multitxt}}
\phantom{a}
\begin{figure}[!ht]
    \centering
    \begin{minipage}[c]{0.997\linewidth}
        \centering
        \FramedBox{0.12\linewidth}{0.162\linewidth}{A concrete stairwell with orange metal handrails and steps leads up to a landing with a light on the wall.}
        \hspace{-0.25cm}
        \FramedBox{0.12\linewidth}{0.162\linewidth}{A concrete hallway with a blue floor leads to a corner with a staircase with orange railings.}
        \hspace{-0.25cm}
        \FramedBox{0.12\linewidth}{0.162\linewidth}{A yellow double door with a small window in each door is set in a concrete wall at the end of a hallway with a blue floor.}
        \hspace{-0.25cm}
        \FramedBox{0.12\linewidth}{0.162\linewidth}{A yellow hallway with a fire hydrant and pipes on the wall.}
        \hspace{-0.25cm}
        \FramedBox{0.12\linewidth}{0.162\linewidth}{A concrete room with a gold ceiling and an orange staircase leading up to a white doorway.}
        \hspace{-0.25cm}
        \FramedBox{0.12\linewidth}{0.162\linewidth}{A black circle is in the center of a tiled floor with a yellow square tile in the bottom right corner.}
    \end{minipage}
    \begin{overpic}[width=0.162\textwidth]{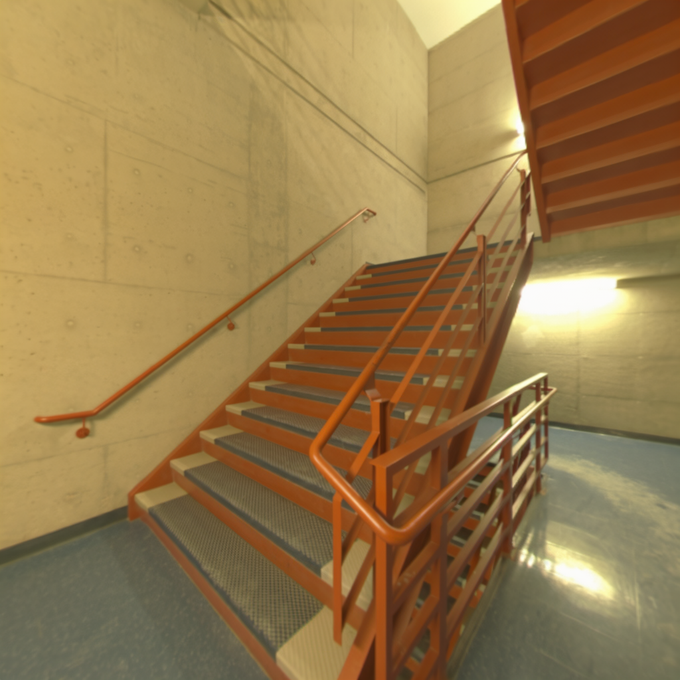}
    \put(6.2,6.2){\color{green}\framebox(87.6, 87.6){}}
    \end{overpic}\!\!
    \hfill
    \begin{overpic}[width=0.162\textwidth]{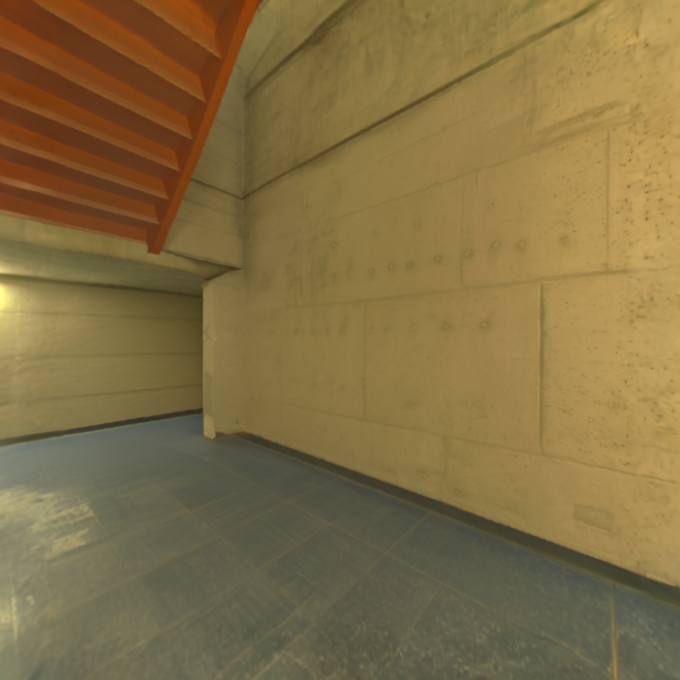}
    \put(6.2,6.2){\color{green}\framebox(87.6, 87.6){}}
    \end{overpic}\!\!
    \hfill
    \begin{overpic}[width=0.162\textwidth]{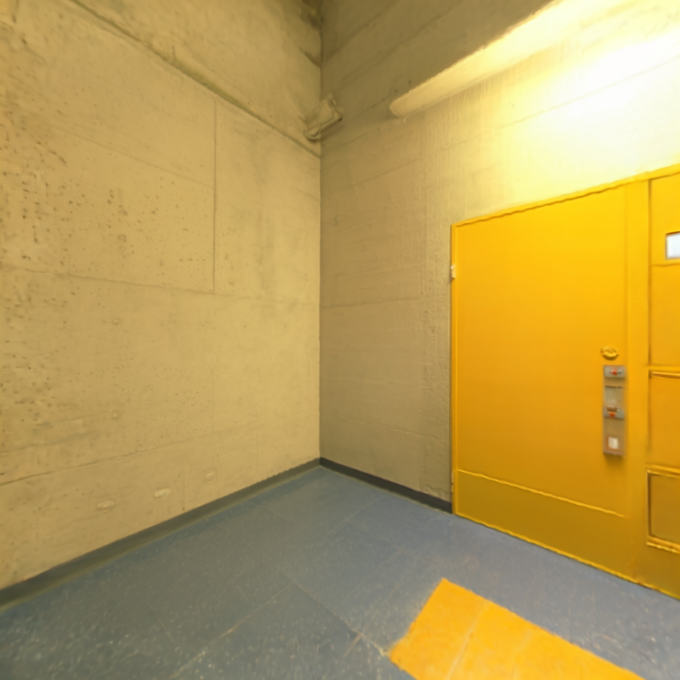}
    \put(6.2,6.2){\color{green}\framebox(87.6, 87.6){}}
    \end{overpic}\!\!
    \hfill
    \begin{overpic}[width=0.162\textwidth]{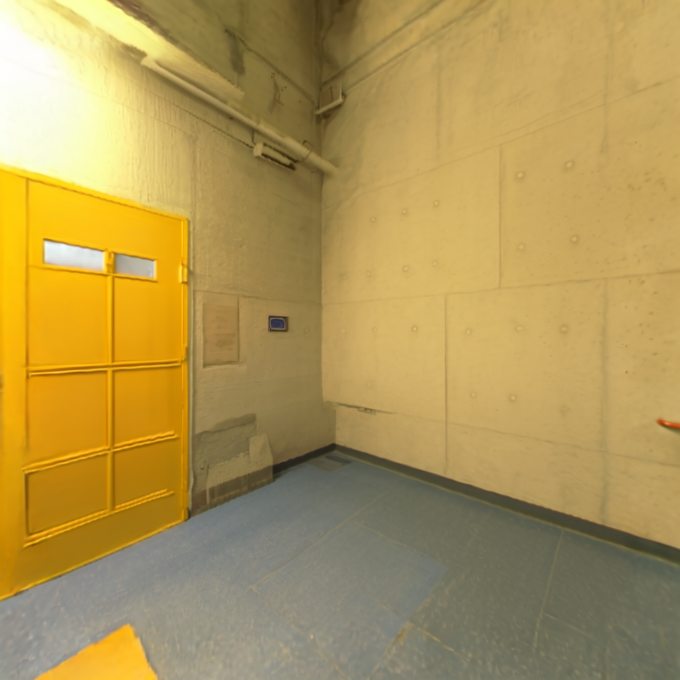}
    \put(6.2,6.2){\color{green}\framebox(87.6, 87.6){}}
    \end{overpic}\!\!
    \hfill
    \begin{overpic}[width=0.162\textwidth]{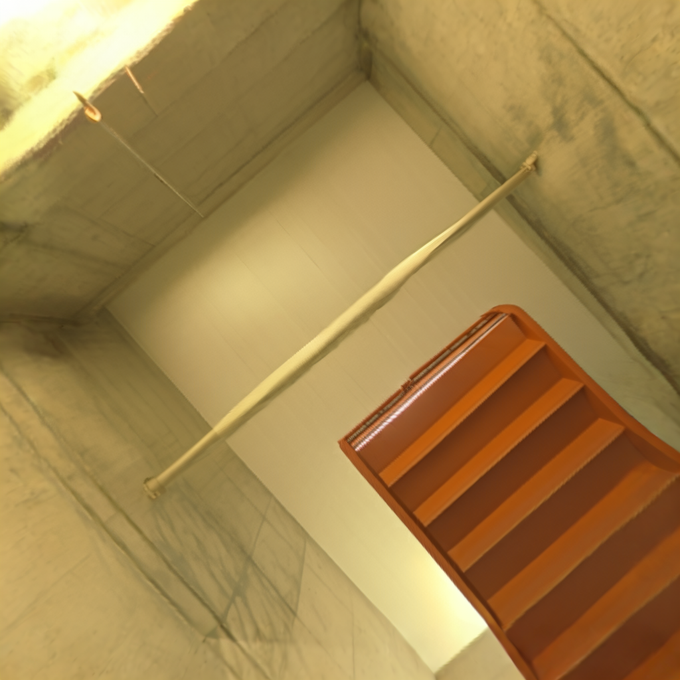}
    \put(6.2,6.2){\color{green}\framebox(87.6, 87.6){}}
    \end{overpic}\!\!
    \hfill
    \begin{overpic}[width=0.162\textwidth]{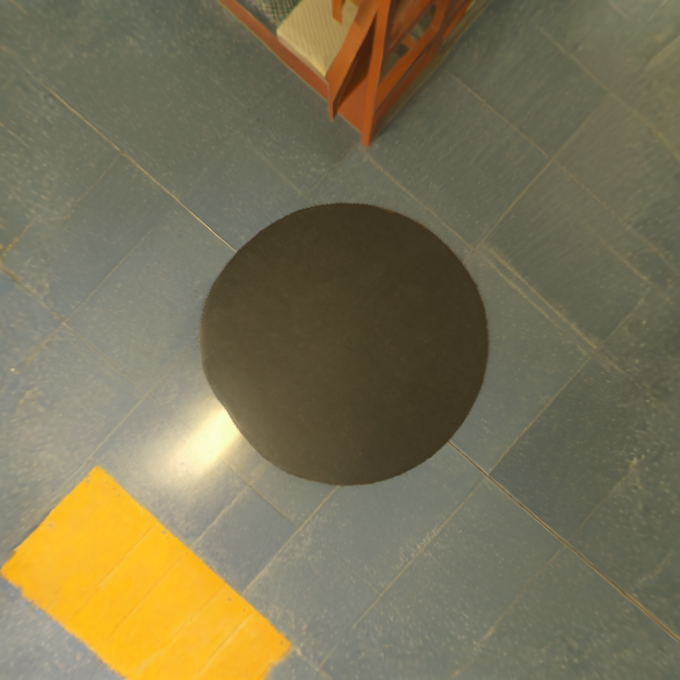}
    \put(6.2,6.2){\color{green}\framebox(87.6, 87.6){}}
    \end{overpic}
    \includegraphics[width=0.162\linewidth,clip,trim=43px 43px 43px 43px]{figures/qualitativecomp/faces/multi/uncropped/9C4A0076-46c51ec2c2/0_0.png}%
    \hfill
    \includegraphics[width=0.162\linewidth,clip,trim=43px 43px 43px 43px]{figures/qualitativecomp/faces/multi/uncropped/9C4A0076-46c51ec2c2/0_1.png}%
    \hfill
    \includegraphics[width=0.162\linewidth,clip,trim=43px 43px 43px 43px]{figures/qualitativecomp/faces/multi/uncropped/9C4A0076-46c51ec2c2/0_2.png}%
    \hfill
    \includegraphics[width=0.162\linewidth,clip,trim=43px 43px 43px 43px]{figures/qualitativecomp/faces/multi/uncropped/9C4A0076-46c51ec2c2/0_3.png}%
    \hfill
    \includegraphics[width=0.162\linewidth,clip,trim=43px 43px 43px 43px]{figures/qualitativecomp/faces/multi/uncropped/9C4A0076-46c51ec2c2/0_4.png}%
    \hfill
    \includegraphics[width=0.162\linewidth,clip,trim=43px 43px 43px 43px]{figures/qualitativecomp/faces/multi/uncropped/9C4A0076-46c51ec2c2/0_5.png}
    \caption{Our generated faces with multi caption input. Top row shows the uncropped faces, bottom row shows the cropped faces.}
\end{figure}

\begin{figure}[!ht]
    \centering
    \begin{minipage}[c]{0.997\linewidth}
        \centering
        \FramedBox{0.12\linewidth}{0.162\linewidth}{A window with a view of a snowy forest and lake is the focal point of a bright, minimalist room with a desk and wooden dresser.}
        \hspace{-0.25cm}
        \FramedBox{0.12\linewidth}{0.162\linewidth}{A bedroom with a green and white striped bedspread, a wooden dresser, and white walls.}
        \hspace{-0.25cm}
        \FramedBox{0.12\linewidth}{0.162\linewidth}{A messy bedroom with a bed, a nightstand, a clothes rack, and a drying rack, with a doorway leading to a black and white tiled bathroom.}
        \hspace{-0.25cm}
        \FramedBox{0.12\linewidth}{0.162\linewidth}{A cluttered white bedroom with a desk, mirror, and plants.}
        \hspace{-0.25cm}
        \FramedBox{0.12\linewidth}{0.162\linewidth}{A white ceiling with a single light bulb and a smoke detector.}
        \hspace{-0.25cm}
        \FramedBox{0.12\linewidth}{0.162\linewidth}{A black circle is in the center of a distorted image of a room with wooden floors, a bed, and a blue bin.}
    \end{minipage}
    \begin{overpic}[width=0.162\textwidth]{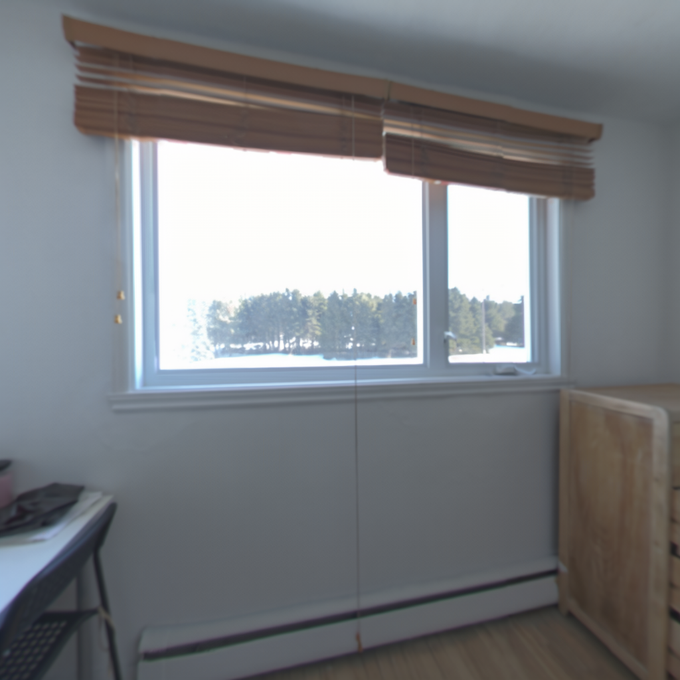}
    \put(6.2,6.2){\color{green}\framebox(87.6, 87.6){}}
    \end{overpic}\!\!
    \hfill
    \begin{overpic}[width=0.162\textwidth]{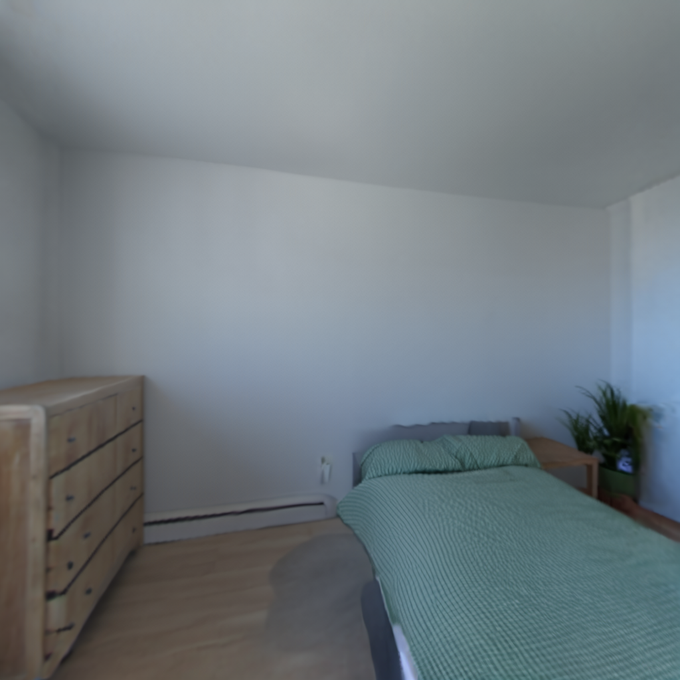}
    \put(6.2,6.2){\color{green}\framebox(87.6, 87.6){}}
    \end{overpic}\!\!
    \hfill
    \begin{overpic}[width=0.162\textwidth]{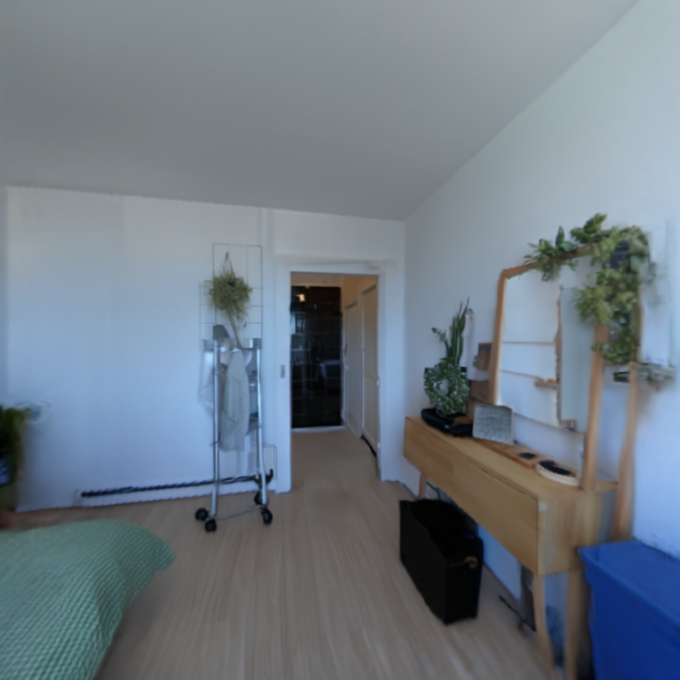}
    \put(6.2,6.2){\color{green}\framebox(87.6, 87.6){}}
    \end{overpic}\!\!
    \hfill
    \begin{overpic}[width=0.162\textwidth]{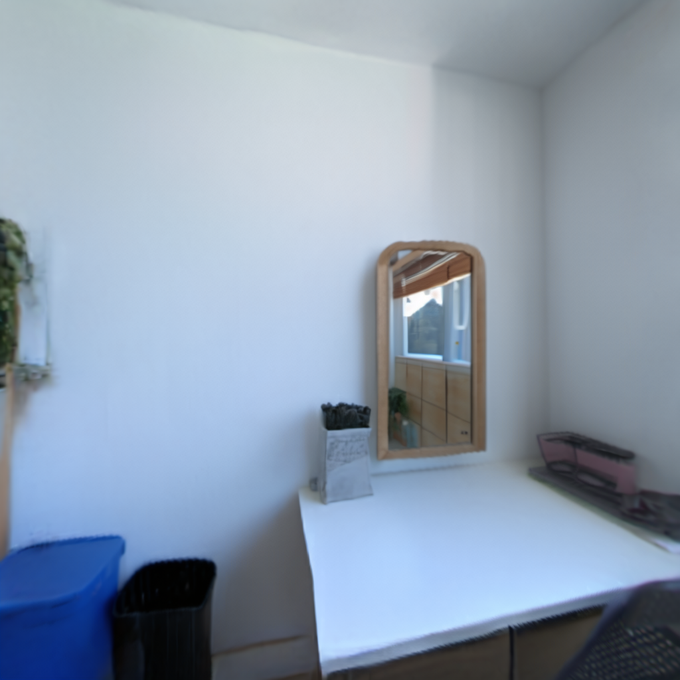}
    \put(6.2,6.2){\color{green}\framebox(87.6, 87.6){}}
    \end{overpic}\!\!
    \hfill
    \begin{overpic}[width=0.162\textwidth]{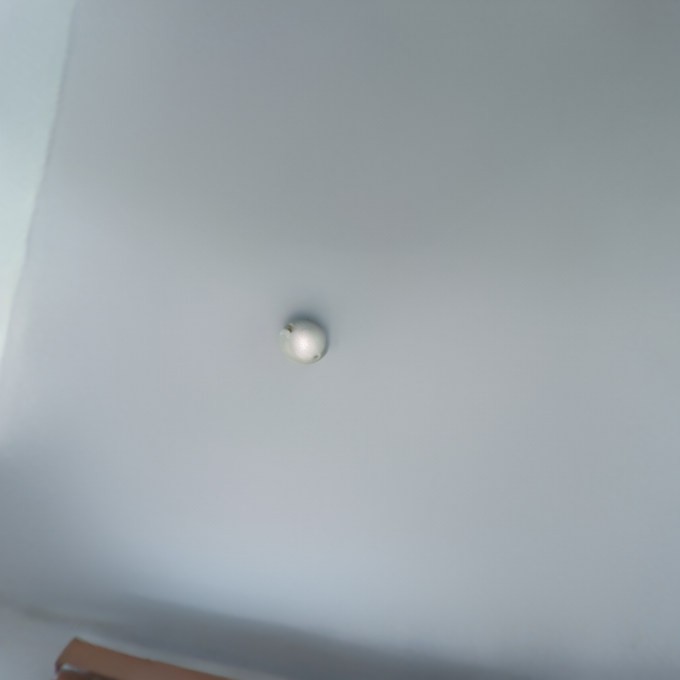}
    \put(6.2,6.2){\color{green}\framebox(87.6, 87.6){}}
    \end{overpic}\!\!
    \hfill
    \begin{overpic}[width=0.162\textwidth]{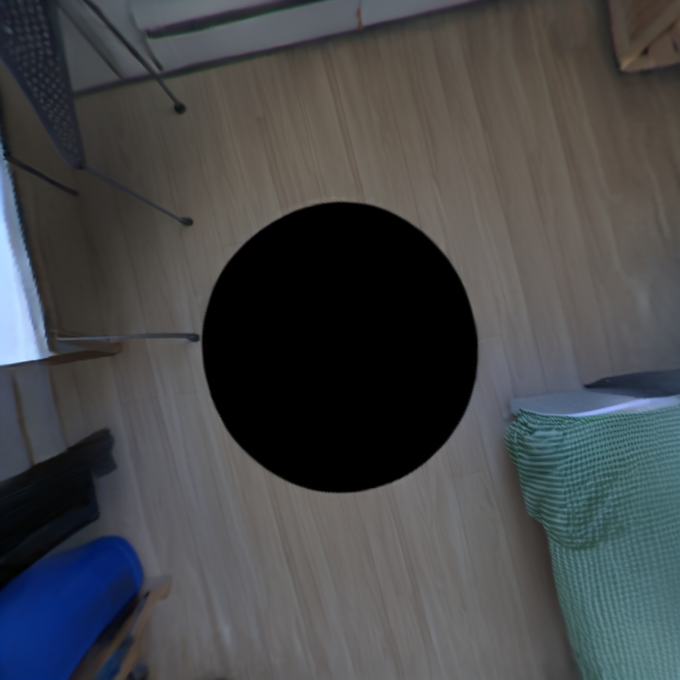}
    \put(6.2,6.2){\color{green}\framebox(87.6, 87.6){}}
    \end{overpic}
    \includegraphics[width=0.162\linewidth,clip,trim=43px 43px 43px 43px]{figures/qualitativecomp/faces/multi/uncropped/9C4A0330-55d4beffc9/0_0.png}%
    \hfill
    \includegraphics[width=0.162\linewidth,clip,trim=43px 43px 43px 43px]{figures/qualitativecomp/faces/multi/uncropped/9C4A0330-55d4beffc9/0_1.png}%
    \hfill
    \includegraphics[width=0.162\linewidth,clip,trim=43px 43px 43px 43px]{figures/qualitativecomp/faces/multi/uncropped/9C4A0330-55d4beffc9/0_2.png}%
    \hfill
    \includegraphics[width=0.162\linewidth,clip,trim=43px 43px 43px 43px]{figures/qualitativecomp/faces/multi/uncropped/9C4A0330-55d4beffc9/0_3.png}%
    \hfill
    \includegraphics[width=0.162\linewidth,clip,trim=43px 43px 43px 43px]{figures/qualitativecomp/faces/multi/uncropped/9C4A0330-55d4beffc9/0_4.png}%
    \hfill
    \includegraphics[width=0.162\linewidth,clip,trim=43px 43px 43px 43px]{figures/qualitativecomp/faces/multi/uncropped/9C4A0330-55d4beffc9/0_5.png}
    \caption{Our generated faces with multi caption input. Top row shows the uncropped faces, bottom row shows the cropped faces.}
\end{figure}

\paragraph{Ground truth}\phantom{a}
\begin{figure}[!ht]
    \centering
    \includegraphics[width=0.49\linewidth]{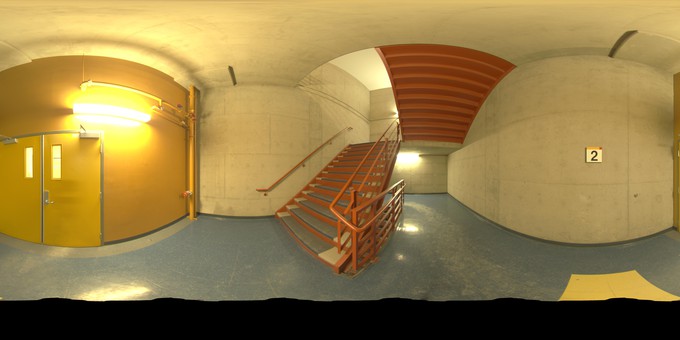}
    \includegraphics[width=0.49\linewidth]{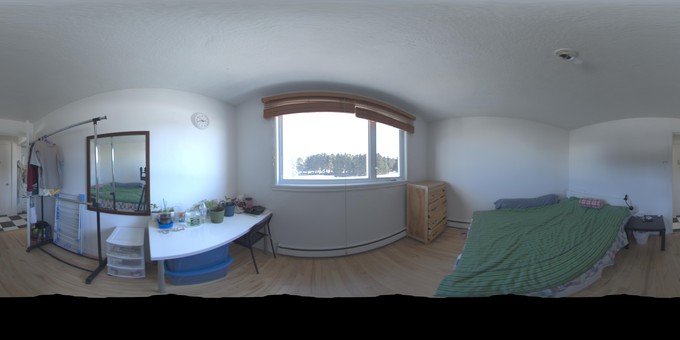}
    
    \caption{Ground truth panoramas from the Laval Indoor dataset.}
    \label{fig:enter-label}
\end{figure}

\newpage

\subsection{VAE reconstructions}
Below, we present pairs of images: the ground truth perspective images with a 95° field of view (FoV) and their corresponding reconstructed images. The reconstructed images are produced by passing the ground truth images through the encoder of our VAE and then decoding the resulting latent representations using the decoder of the same VAE.

\begin{figure}[!ht]
    \centering
    \begin{minipage}[c]{0.245\linewidth}
        \begin{center}
            Original
        \end{center}
    \end{minipage}
    \begin{minipage}[c]{0.245\linewidth}
        \begin{center}
            Reconstructed
        \end{center}
    \end{minipage}
    \begin{minipage}[c]{0.245\linewidth}
        \begin{center}
            Original
        \end{center}
    \end{minipage}
    \begin{minipage}[c]{0.245\linewidth}
        \begin{center}
            Reconstructed
        \end{center}
    \end{minipage}
    \includegraphics[width=0.245\linewidth]{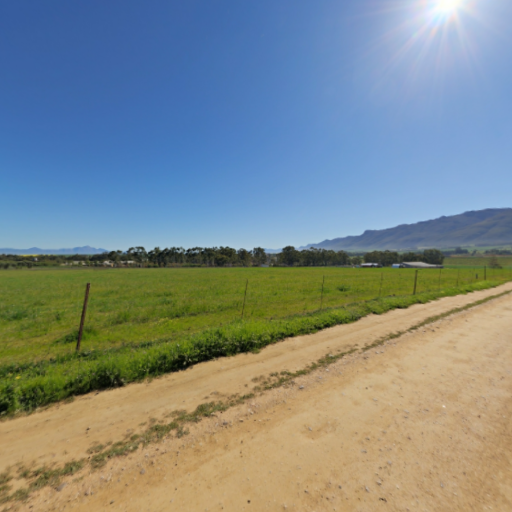}%
    \hfill
    \includegraphics[width=0.245\linewidth]{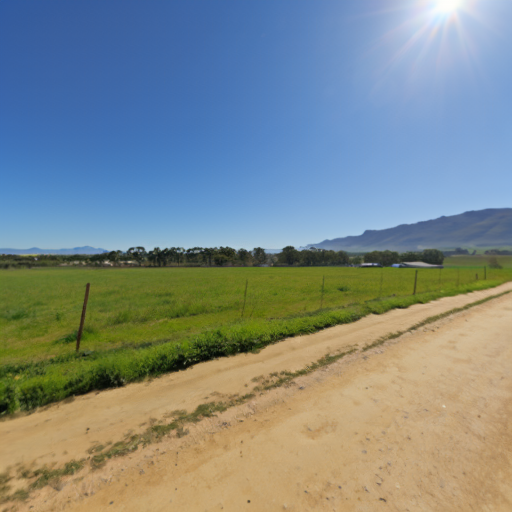}%
    \hfill
    \includegraphics[width=0.245\linewidth]{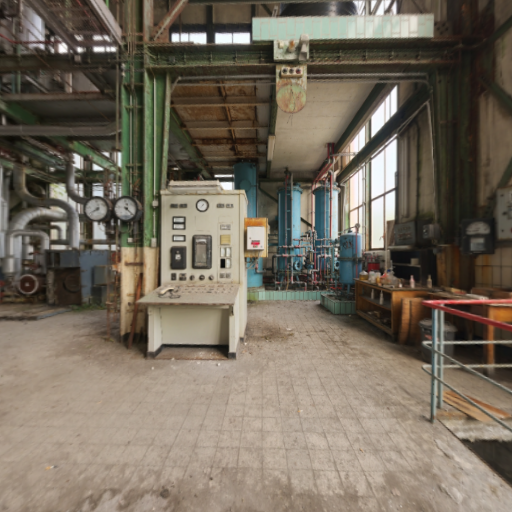}%
    \hfill
    \includegraphics[width=0.245\linewidth]{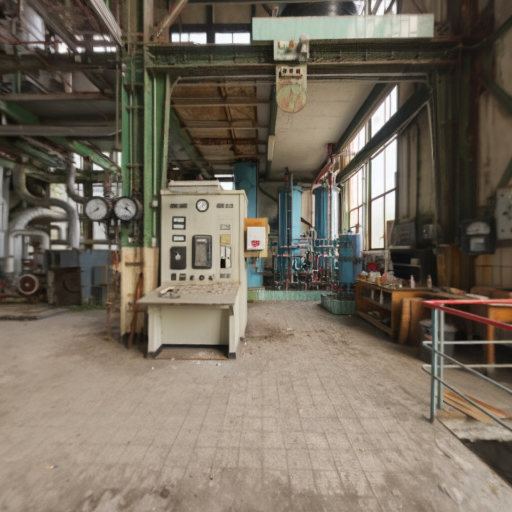}
    \includegraphics[width=0.245\linewidth]{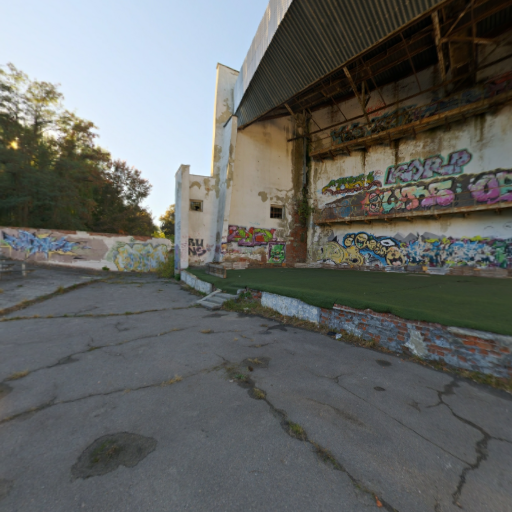}%
    \hfill
    \includegraphics[width=0.245\linewidth]{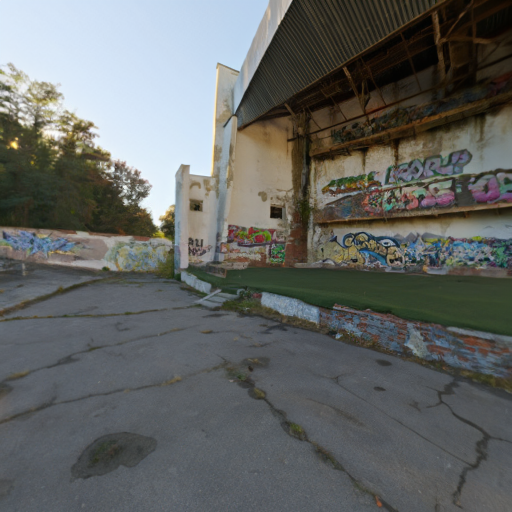}%
    \hfill
    \includegraphics[width=0.245\linewidth]{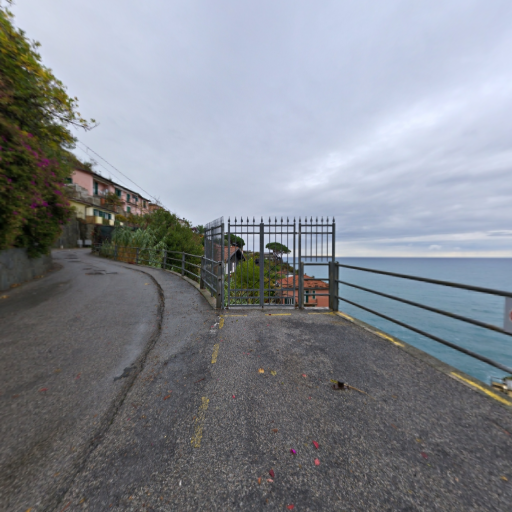}%
    \hfill
    \includegraphics[width=0.245\linewidth]{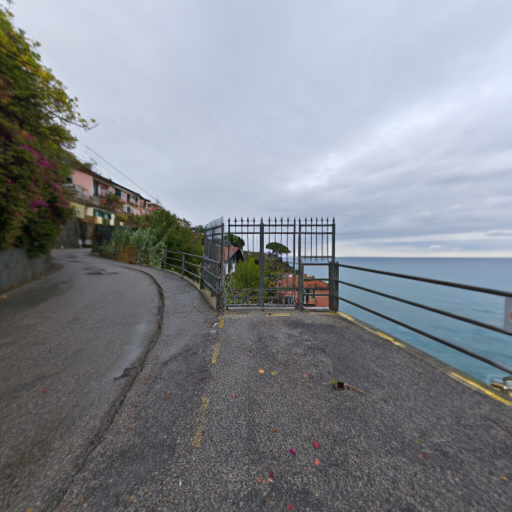}
    \includegraphics[width=0.245\linewidth]{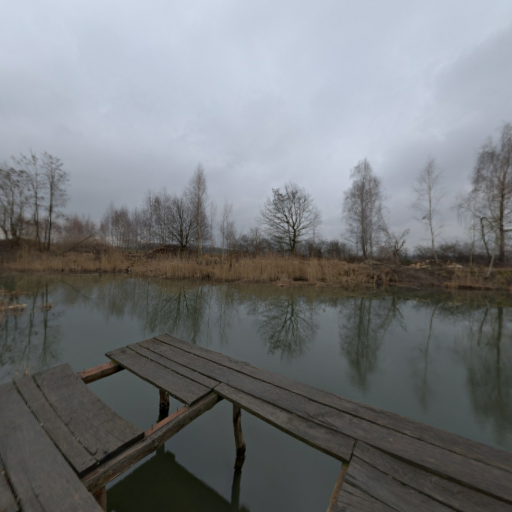}%
    \hfill
    \includegraphics[width=0.245\linewidth]{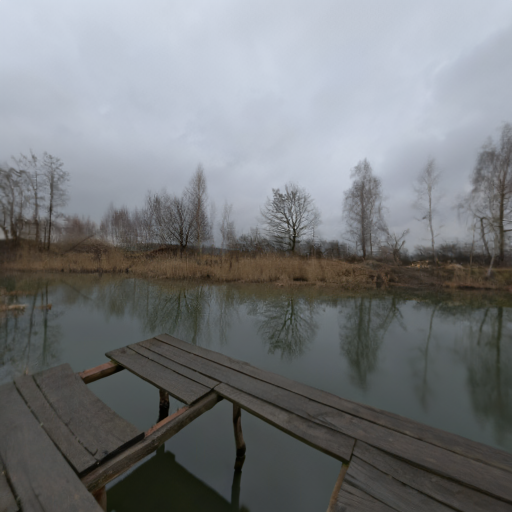}%
    \hfill
    \includegraphics[width=0.245\linewidth]{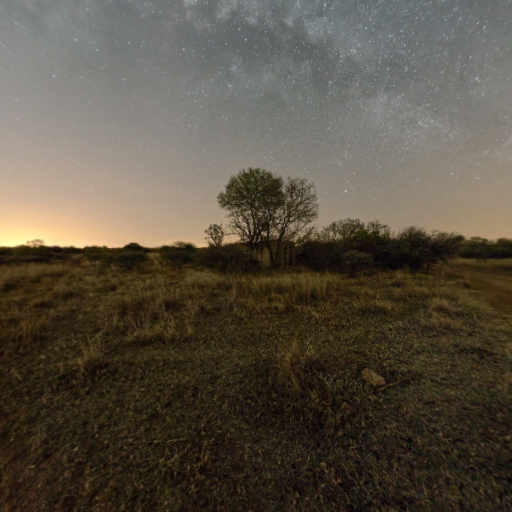}%
    \hfill
    \includegraphics[width=0.245\linewidth]{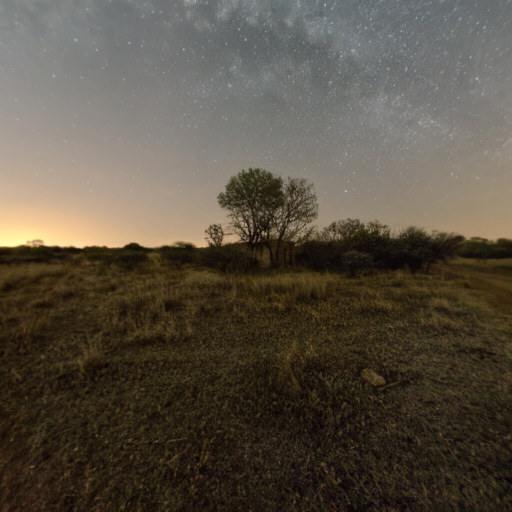}
    \includegraphics[width=0.245\linewidth]{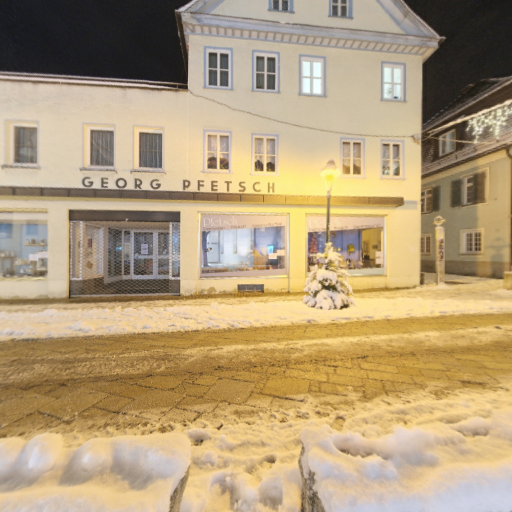}%
    \hfill
    \includegraphics[width=0.245\linewidth]{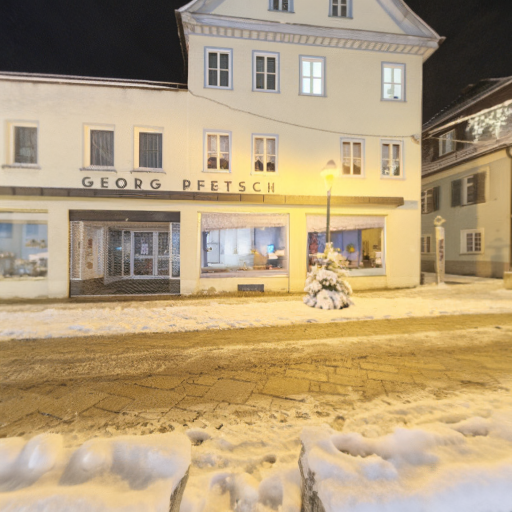}%
    \hfill
    \includegraphics[width=0.245\linewidth]{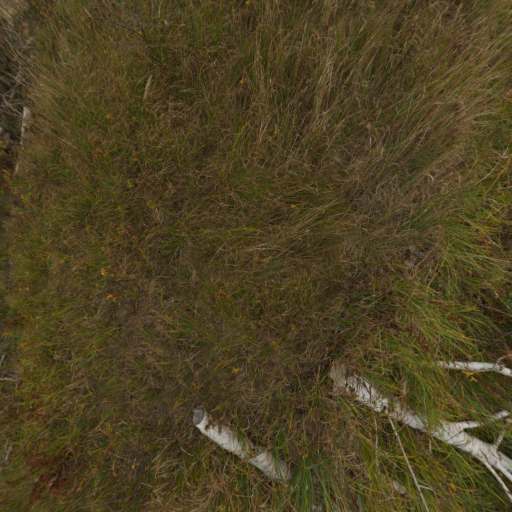}%
    \hfill
    \includegraphics[width=0.245\linewidth]{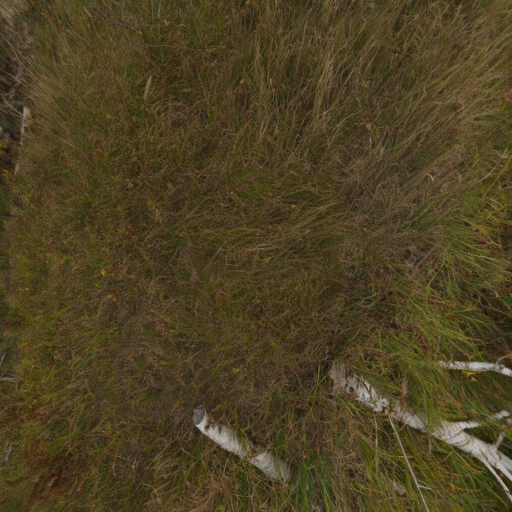}
    \caption{Examples of ground truth and encoded-decoded perspective images with a FoV of 95° using our VAE. The VAE is capable of reconstructing perspective images with out loss of quality.}
\end{figure}

\newpage

\subsection{Detailed architecture}
\begin{figure}[!ht]
    \centering
    \includegraphics[width=0.75\linewidth]{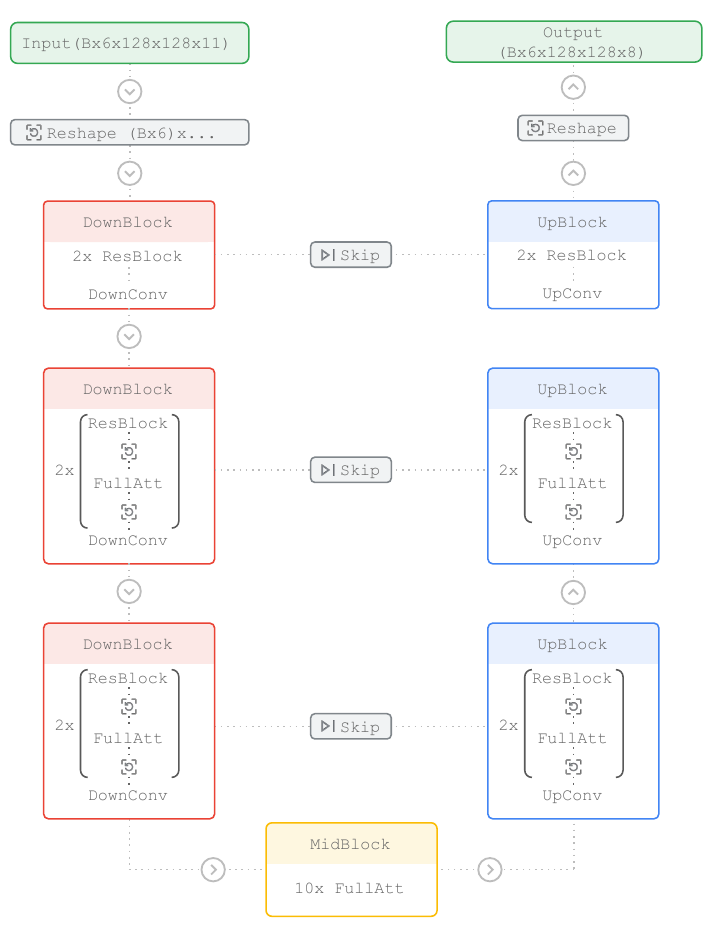}
    \caption{Illustration of our latent diffusion model.}\label{fig:detailedarch}
\end{figure}

We illustrate our latent diffusion model in \Cref{fig:detailedarch}. The model's input is a concatenation of encoded latents, positional encodings, and an input mask indicating the conditioning image. To generate the initial latents, the input image is encoded with a VAE, while Gaussian noise is sampled for the other five faces. The VAE architecture is identical to that of Stable Diffusion's VAE, with one modification: all GroupNorm layers are replaced with synchronized GroupNorms, where normalization is computed across both the spatial and frame dimensions.

The combined input is downsampled three times to a resolution of $B\times 6 \times 32 \times 32$. The first and last blocks of the model exclude attention layers and operate independently on each face. Once the final layer is computed, the output is processed through the synchronized decoder. Notably, except for the GroupNorm layers, all computations are performed per face, with no awareness of the overall panorama structure.

Despite this simplicity, our approach outperforms existing methods, as demonstrated by our results.

\newpage
\subsection{Ablations on panoramic data}

To evaluate the impact of dataset size on the performance of our method, we conducted an ablation study by training CubeDiff on three subsets of panoramic data: a tiny dataset containing approximately 700 panoramas from the Polyhaven dataset, a medium dataset of about 20,000 panoramas from the Structured3D dataset (the same dataset PanoDiffusion used and comparable in size to MVDiffusion), and a full dataset with over 40,000 panoramas. The results demonstrate that CubeDiff performs robustly across all settings. Even the tiny model, trained on only 700 panoramas, achieves competitive results, while the medium model closely matches the performance of the full model and significantly outperforms baseline methods in most metrics. Qualitative results further confirm the ability of the tiny and medium models to generate visually consistent and high-quality panoramas, demonstrating CubeDiff’s robustness even with constrained data. These findings indicate that the superior performance of CubeDiff stems not only from data volume but also from the strength of the cubemap representation and its compatibility with pretrained latent diffusion models.

\begin{table}[!ht]
\centering
\small
\setlength{\tabcolsep}{3pt} %
\renewcommand{\arraystretch}{1.1}

\resizebox{0.9\linewidth}{!}{
\begin{tabular}{lccccccccccc}
        \specialrule{0.75pt}{0pt}{1pt}
        & \multicolumn{3}{c}{\textbf{LAVAL Indoor}} &  & \multicolumn{3}{c}{\textbf{SUN360}} \\
        \hhline{~---~---}
        & FID $\downarrow$ & KID ($\times 10^2$)$\downarrow$ & Clip-FID $\downarrow$ & & FID$\downarrow$  & KID ($\times 10^2$)$\downarrow$  & Clip-FID$\downarrow$ \\
        \hhline{----~---}
        Text2Light & 28.3 & 1.45 & 11.5 & & 60.1 & 4.31 & 31.3\\
        PanFusion & 41.7 & 2.85 & 19.8 &  & \cellcolor{tabyellowlight} 30.0 & \cellcolor{tabyellowlight} 1.42 & \cellcolor{tabyellow}\phantom{0}7.8 \\
        OmniDreamer & 71.0 & 5.17 & 23.9 & & 92.3 & 8.89 & 51.7\\
        PanoDiffusion & 58.6 & 4.08 & 26.6 & & 52.9 & 3.51 & 28.9\\
        Diffusion360 & 33.1 & 2.07 & 16.9 & & 45.4 & 3.73 & 18.5 \\
        MVDiffusion & \cellcolor{tabyellowlight}25.7 & 1.11 & 13.5 & & 50.9 & 3.71 & 15.4 \\
        \hhline{----~---}
        \cellcolor{Gray}\textbf{Ours\textsubscript{tiny}} & 27.3 & \cellcolor{tabyellowlight}1.05 &  \cellcolor{tabyellowlight}\phantom{0}8.8 & & 41.7 & 2.99 & 14.7\\
        \cellcolor{Gray}\textbf{Ours\textsubscript{medium}} & \cellcolor{tabyellow}13.8 & \cellcolor{tabyellow}0.66 & \cellcolor{tabyellow}\phantom{0}8.5 & & \cellcolor{tabgreen}\textbf{23.9} & \cellcolor{tabgreen}\textbf{1.28} & \cellcolor{tabyellowlight}10.7 & \\     \cellcolor{Gray}\textbf{Ours\textsubscript{full}} & \cellcolor{tabgreen}\textbf{10.0} & \cellcolor{tabgreen}\textbf{0.35} & \cellcolor{tabgreen}\textbf{\phantom{0}4.1} & & \cellcolor{tabyellow}24.1 & \cellcolor{tabyellow}1.33 & \cellcolor{tabgreen}\textbf{\phantom{0}7.0}\\
\specialrule{0.75pt}{1pt}{1pt}
\specialrule{0.75pt}{1pt}{1pt}
\end{tabular}
}
\caption{\textbf{Quantitative Ablation on the Laval Indoor and SUN360 dataset.} We train a model (Ours\textsubscript{tiny}) on a tiny dataset and another model (Ours\textsubscript{medium}) on a medium dataset}
\label{tab:ablation:datasetsize}
\vspace{-1em}
\end{table}

\begin{figure}[!ht]
 \centering
    \begin{minipage}[c][0.2\linewidth]{0.03\linewidth}
        \centering
        \rotatebox{90}{\textbf{Ours\textsubscript{tiny}}}
    \end{minipage}
    \begin{minipage}[c]{0.45\linewidth}
        \centering
    \includegraphics[width=\linewidth, height=0.5\linewidth]{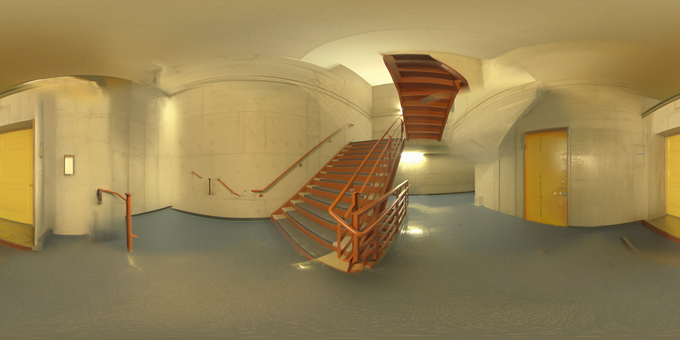}
    \end{minipage}
    \begin{minipage}[c]{0.45\linewidth}
        \centering
    \includegraphics[width=\linewidth, height=0.5\linewidth]{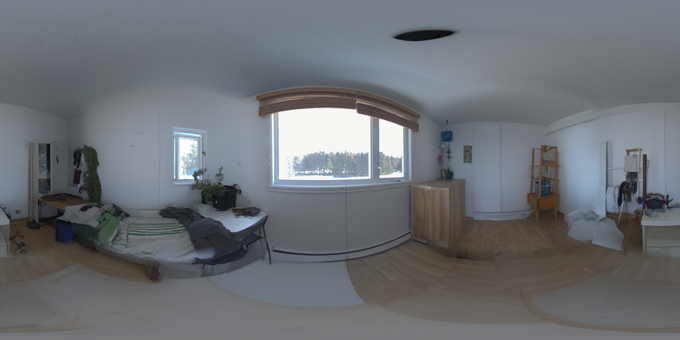}
    \end{minipage}\\
    \begin{minipage}[c][0.2\linewidth]{0.03\linewidth}
        \centering
        \rotatebox{90}{\textbf{Ours\textsubscript{medium}}}
    \end{minipage}
    \begin{minipage}[c]{0.45\linewidth}
        \centering
    \includegraphics[width=\linewidth, height=0.5\linewidth]{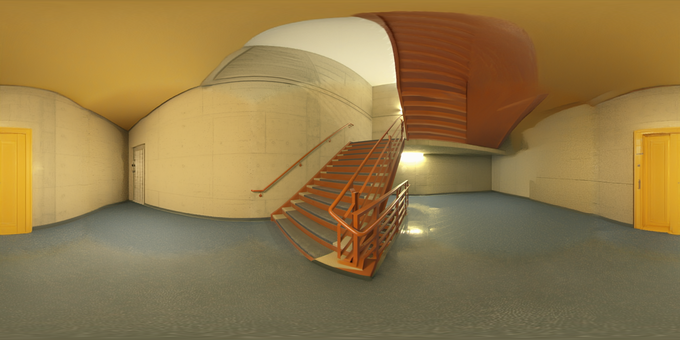}
    \end{minipage}
    \begin{minipage}[c]{0.45\linewidth}
        \centering
    \includegraphics[width=\linewidth, height=0.5\linewidth]{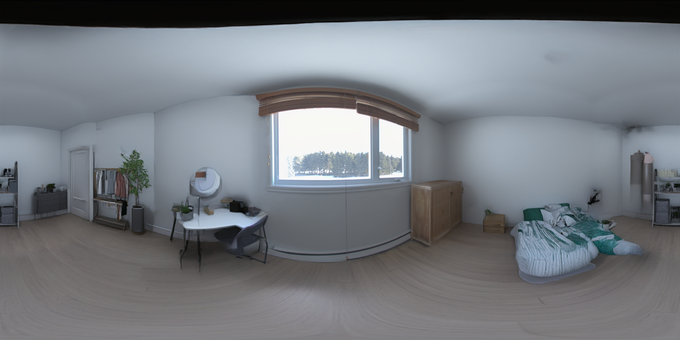}
    \end{minipage}\\
    \begin{minipage}[c][0.2\linewidth]{0.03\linewidth}
        \centering
        \rotatebox{90}{\textbf{Ours\textsubscript{full}}}
    \end{minipage}
    \begin{minipage}[c]{0.45\linewidth}
        \centering
    \includegraphics[width=\linewidth, height=0.5\linewidth]{figures/qualitativecomp/a/ours_multi_9C4A0076-46c51ec2c2.jpg}
    \end{minipage}
    \begin{minipage}[c]{0.45\linewidth}
        \centering
    \includegraphics[width=\linewidth, height=0.5\linewidth]{figures/qualitativecomp/b/ours_multi_9C4A0330-55d4beffc9.jpg}
    \end{minipage}
    \caption{Qualitative results of the ablated models. Both the tiny and the medium model are able to generate consistent panoramas.}\label{fig:appendix:dataablation}
\end{figure}
\newpage
\phantom{a}
\begin{figure}[!t]
    \centering
    \begin{minipage}[c][0.2\linewidth]{0.03\linewidth}
        \centering
        \rotatebox{90}{\textbf{Ours\textsubscript{tiny}}}
    \end{minipage}
    \begin{minipage}[c]{0.65\linewidth}
        \centering
    \includegraphics[width=\linewidth, height=0.5\linewidth]{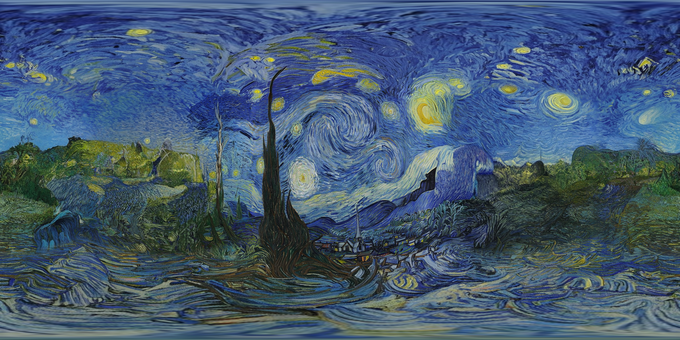}
    \end{minipage}\\
    \begin{minipage}[c][0.2\linewidth]{0.03\linewidth}
        \centering
        \rotatebox{90}{\textbf{Ours\textsubscript{medium}}}
    \end{minipage}
    \begin{minipage}[c]{0.65\linewidth}
        \centering
    \includegraphics[width=\linewidth, height=0.5\linewidth]{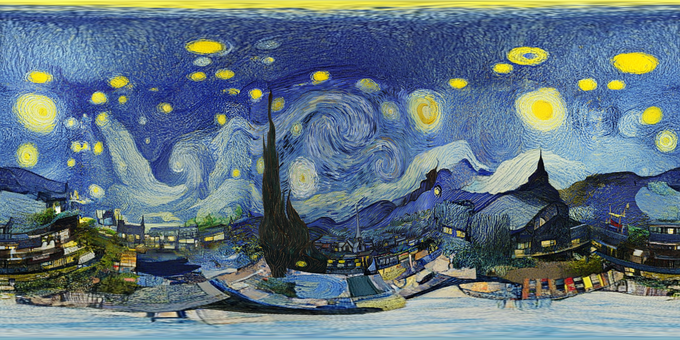}
    \end{minipage}
    \\
    \begin{minipage}[c][0.2\linewidth]{0.03\linewidth}
        \centering
        \rotatebox{90}{\textbf{MVDiffusion}}
    \end{minipage}
   \begin{minipage}[c]{0.65\linewidth}
        \centering
    \includegraphics[width=\linewidth, height=0.5\linewidth]{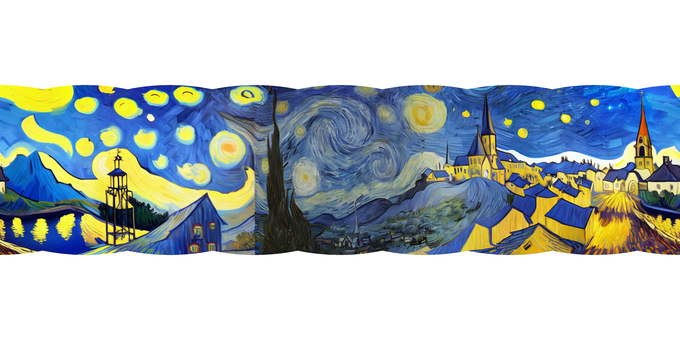}
    \end{minipage}
    \caption{Example of an OOD generation of our tiny (top row) model, medium model (second row) and MVDiffusion (last row).}
    \label{fig:appendix:dataablation:ood}
\end{figure}

%% file: main.bbl
\begin{thebibliography}{46}
\providecommand{\natexlab}[1]{#1}
\providecommand{\url}[1]{\texttt{#1}}
\expandafter\ifx\csname urlstyle\endcsname\relax
  \providecommand{\doi}[1]{doi: #1}\else
  \providecommand{\doi}{doi: \begingroup \urlstyle{rm}\Url}\fi

\bibitem[Akimoto et~al.(2022)Akimoto, Matsuo, and Aoki]{akimoto2022diverse}
Naofumi Akimoto, Yuhi Matsuo, and Yoshimitsu Aoki.
\newblock Diverse plausible 360-degree image outpainting for efficient {3DCG} background creation.
\newblock In \emph{IEEE/CVF Conference on Computer Vision and Pattern Recognition (CVPR)}, 2022.

\bibitem[Baranchuk et~al.(2021)Baranchuk, Rubachev, Voynov, Khrulkov, and Babenko]{baranchuk2021label}
Dmitry Baranchuk, Ivan Rubachev, Andrey Voynov, Valentin Khrulkov, and Artem Babenko.
\newblock Label-efficient semantic segmentation with diffusion models.
\newblock \emph{arXiv preprint arXiv:2112.03126}, 2021.

\bibitem[Bi{\'n}kowski et~al.(2018)Bi{\'n}kowski, Sutherland, Arbel, and Gretton]{binkowski2018demystifying}
Miko{\l}aj Bi{\'n}kowski, Danica~J Sutherland, Michael Arbel, and Arthur Gretton.
\newblock Demystifying {MMD} {GANs}.
\newblock \emph{arXiv preprint arXiv:1801.01401}, 2018.

\bibitem[Chen et~al.(2022)Chen, Wang, and Liu]{chen2022text2light}
Zhaoxi Chen, Guangcong Wang, and Ziwei Liu.
\newblock {Text2Light}: Zero-shot text-driven {HDR} panorama generation.
\newblock \emph{ACM Transactions on Graphics (TOG)}, 41\penalty0 (6), 2022.

\bibitem[Feng et~al.(2023)Feng, Liu, Cui, and Xie]{feng2023diffusion360}
Mengyang Feng, Jinlin Liu, Miaomiao Cui, and Xuansong Xie.
\newblock Diffusion360: Seamless 360 degree panoramic image generation based on diffusion models.
\newblock \emph{arXiv preprint arXiv:2311.13141}, 2023.

\bibitem[Gao et~al.(2024{\natexlab{a}})Gao, Yao, Ye, Wang, Yao, and Wang]{gao2024opa}
Penglei Gao, Kai Yao, Tiandi Ye, Steven Wang, Yuan Yao, and Xiaofeng Wang.
\newblock {OPa-Ma}: Text guided {Mamba} for 360-degree image out-painting.
\newblock \emph{arXiv preprint arXiv:2407.10923}, 2024{\natexlab{a}}.

\bibitem[Gao et~al.(2024{\natexlab{b}})Gao, Holynski, Henzler, Brussee, Martin-Brualla, Srinivasan, Barron, and Poole]{gao2024cat3d}
Ruiqi Gao, Aleksander Holynski, Philipp Henzler, Arthur Brussee, Ricardo Martin-Brualla, Pratul Srinivasan, Jonathan~T Barron, and Ben Poole.
\newblock {CAT3D}: Create anything in 3d with multi-view diffusion models.
\newblock \emph{arXiv preprint arXiv:2405.10314}, 2024{\natexlab{b}}.

\bibitem[Gardner et~al.(2017)Gardner, Sunkavalli, Yumer, Shen, Gambaretto, Gagn{\'e}, and Lalonde]{gardner2017learning}
Marc-Andr{\'e} Gardner, Kalyan Sunkavalli, Ersin Yumer, Xiaohui Shen, Emiliano Gambaretto, Christian Gagn{\'e}, and Jean-Fran{\c{c}}ois Lalonde.
\newblock Learning to predict indoor illumination from a single image.
\newblock \emph{arXiv preprint arXiv:1704.00090}, 2017.

\bibitem[{Gemini Team Google}(2023)]{team2023gemini}
{Gemini Team Google}.
\newblock Gemini: A family of highly capable multimodal models.
\newblock \emph{arXiv preprint arXiv:2312.11805}, 2023.

\bibitem[He et~al.(2023)He, Yang, Chen, Cun, Xia, Zhang, Wang, He, Chen, and Shan]{he2023scalecrafter}
Yingqing He, Shaoshu Yang, Haoxin Chen, Xiaodong Cun, Menghan Xia, Yong Zhang, Xintao Wang, Ran He, Qifeng Chen, and Ying Shan.
\newblock {ScaleCrafter}: Tuning-free higher-resolution visual generation with diffusion models.
\newblock In \emph{International Conference on Learning Representations (ICLR)}, 2023.

\bibitem[Hessel et~al.(2021)Hessel, Holtzman, Forbes, Bras, and Choi]{hessel2021clipscore}
Jack Hessel, Ari Holtzman, Maxwell Forbes, Ronan~Le Bras, and Yejin Choi.
\newblock {CLIPScore}: A reference-free evaluation metric for image captioning.
\newblock \emph{arXiv preprint arXiv:2104.08718}, 2021.

\bibitem[Heusel et~al.(2017)Heusel, Ramsauer, Unterthiner, Nessler, and Hochreiter]{heusel2017gans}
Martin Heusel, Hubert Ramsauer, Thomas Unterthiner, Bernhard Nessler, and Sepp Hochreiter.
\newblock {GANs} trained by a two time-scale update rule converge to a local {Nash} equilibrium.
\newblock \emph{Advances in Neural Information Processing Systems (NeurIPS)}, 30, 2017.

\bibitem[Ho \& Salimans(2022)Ho and Salimans]{ho2022classifier}
Jonathan Ho and Tim Salimans.
\newblock Classifier-free diffusion guidance.
\newblock \emph{arXiv preprint arXiv:2207.12598}, 2022.

\bibitem[Huang et~al.(2024)Huang, Wen, Dong, Wang, Li, Chen, Cao, Liang, Qiao, Dai, et~al.]{huang2024epidiff}
Zehuan Huang, Hao Wen, Junting Dong, Yaohui Wang, Yangguang Li, Xinyuan Chen, Yan-Pei Cao, Ding Liang, Yu~Qiao, Bo~Dai, et~al.
\newblock {EpiDiff}: Enhancing multi-view synthesis via localized epipolar-constrained diffusion.
\newblock In \emph{IEEE/CVF Conference on Computer Vision and Pattern Recognition (CVPR)}, 2024.

\bibitem[Jin et~al.(2024)Jin, Li, Luan, Xiangli, Bi, Zhang, Xu, Sun, and Snavely]{jin2024neural}
Haian Jin, Yuan Li, Fujun Luan, Yuanbo Xiangli, Sai Bi, Kai Zhang, Zexiang Xu, Jin Sun, and Noah Snavely.
\newblock Neural gaffer: Relighting any object via diffusion.
\newblock \emph{arXiv preprint arXiv:2406.07520}, 2024.

\bibitem[Kalischek et~al.(2022)Kalischek, Peters, Wegner, and Schindler]{kalischek2022tetradiffusion}
Nikolai Kalischek, Torben Peters, Jan~D Wegner, and Konrad Schindler.
\newblock {TetraDiffusion}: Tetrahedral diffusion models for 3d shape generation.
\newblock In \emph{European Conference on Computer Vision (ECCV)}, 2022.

\bibitem[Ke et~al.(2024)Ke, Obukhov, Huang, Metzger, Daudt, and Schindler]{ke2024repurposing}
Bingxin Ke, Anton Obukhov, Shengyu Huang, Nando Metzger, Rodrigo~Caye Daudt, and Konrad Schindler.
\newblock Repurposing diffusion-based image generators for monocular depth estimation.
\newblock In \emph{IEEE/CVF Conference on Computer Vision and Pattern Recognition (CVPR)}, 2024.

\bibitem[Kingma \& Ba(2014)Kingma and Ba]{Kingma2014Adam}
Diederik Kingma and Jimmy Ba.
\newblock Adam: A method for stochastic optimization.
\newblock \emph{International Conference on Learning Representations (ICLR)}, 12 2014.

\bibitem[Kocabas et~al.(2021)Kocabas, Huang, Tesch, M{\"u}ller, Hilliges, and Black]{kocabas2021spec}
Muhammed Kocabas, Chun-Hao~P. Huang, Joachim Tesch, Lea M{\"u}ller, Otmar Hilliges, and Michael~J. Black.
\newblock {SPEC}: Seeing people in the wild with an estimated camera.
\newblock In \emph{International Conference on Computer Vision (ICCV)}, 2021.

\bibitem[Kynk{\"a}{\"a}nniemi et~al.(2022)Kynk{\"a}{\"a}nniemi, Karras, Aittala, Aila, and Lehtinen]{kynkaanniemi2022role}
Tuomas Kynk{\"a}{\"a}nniemi, Tero Karras, Miika Aittala, Timo Aila, and Jaakko Lehtinen.
\newblock The role of {ImageNet} classes in {Fr{\'e}chet} {Inception} {Distance}.
\newblock \emph{arXiv preprint arXiv:2203.06026}, 2022.

\bibitem[Lu et~al.(2024)Lu, Hu, Wang, Bai, and Wang]{lu2024autoregressive}
Zhuqiang Lu, Kun Hu, Chaoyue Wang, Lei Bai, and Zhiyong Wang.
\newblock Autoregressive omni-aware outpainting for open-vocabulary 360-degree image generation.
\newblock In \emph{AAAI Conference on Artificial Intelligence (AAAI)}, volume~38, 2024.

\bibitem[Mohammad~Khalid et~al.(2022)Mohammad~Khalid, Xie, Belilovsky, and Popa]{mohammad2022clip}
Nasir Mohammad~Khalid, Tianhao Xie, Eugene Belilovsky, and Tiberiu Popa.
\newblock {CLIP-Mesh}: Generating textured meshes from text using pretrained image-text models.
\newblock In \emph{SIGGRAPH Asia}, 2022.

\bibitem[Persson(accessed 09/2024)]{humus}
Emil Persson.
\newblock Texture from {Humus}.
\newblock \emph{https://www.humus.name/index.php?page=Textures}, accessed 09/2024.

\bibitem[polyhaven.com(accessed 09/2024)]{polyhaven}
polyhaven.com.
\newblock {HDRIs}.
\newblock \emph{https://polyhaven.com/hdris}, accessed 09/2024.

\bibitem[Poole et~al.(2022)Poole, Jain, Barron, and Mildenhall]{poole2022dreamfusion}
Ben Poole, Ajay Jain, Jonathan~T Barron, and Ben Mildenhall.
\newblock {DreamFusion}: Text-to-3d using 2d diffusion.
\newblock \emph{arXiv preprint arXiv:2209.14988}, 2022.

\bibitem[Radford et~al.(2021)Radford, Kim, Hallacy, Ramesh, Goh, Agarwal, Sastry, Askell, Mishkin, Clark, et~al.]{radford2021learning}
Alec Radford, Jong~Wook Kim, Chris Hallacy, Aditya Ramesh, Gabriel Goh, Sandhini Agarwal, Girish Sastry, Amanda Askell, Pamela Mishkin, Jack Clark, et~al.
\newblock Learning transferable visual models from natural language supervision.
\newblock In \emph{International Conference on Machine Learning (ICML)}, 2021.

\bibitem[Rombach et~al.(2022)Rombach, Blattmann, Lorenz, Esser, and Ommer]{Rombach2022Stable}
Robin Rombach, Andreas Blattmann, Dominik Lorenz, Patrick Esser, and Bj\"orn Ommer.
\newblock High-resolution image synthesis with latent diffusion models.
\newblock In \emph{Proceedings of the IEEE/CVF Conference on Computer Vision and Pattern Recognition (CVPR)}, 2022.

\bibitem[Saharia et~al.(2022)Saharia, Chan, Saxena, Li, Whang, Denton, Ghasemipour, Gontijo~Lopes, Karagol~Ayan, Salimans, et~al.]{saharia2022photorealistic}
Chitwan Saharia, William Chan, Saurabh Saxena, Lala Li, Jay Whang, Emily~L Denton, Kamyar Ghasemipour, Raphael Gontijo~Lopes, Burcu Karagol~Ayan, Tim Salimans, et~al.
\newblock Photorealistic text-to-image diffusion models with deep language understanding.
\newblock \emph{Advances in Neural Information Processing Systems (NeurIPS)}, 2022.

\bibitem[Salimans \& Ho(2022)Salimans and Ho]{salimans2022progressive}
Tim Salimans and Jonathan Ho.
\newblock Progressive distillation for fast sampling of diffusion models.
\newblock \emph{arXiv preprint arXiv:2202.00512}, 2022.

\bibitem[Shi et~al.(2023)Shi, Wang, Ye, Long, Li, and Yang]{shi2023mvdream}
Yichun Shi, Peng Wang, Jianglong Ye, Mai Long, Kejie Li, and Xiao Yang.
\newblock {MVDream}: Multi-view diffusion for 3d generation.
\newblock \emph{arXiv preprint arXiv:2308.16512}, 2023.

\bibitem[Somanath \& Kurz(2021)Somanath and Kurz]{somanath2021hdr}
Gowri Somanath and Daniel Kurz.
\newblock {HDR} environment map estimation for real-time augmented reality.
\newblock In \emph{IEEE/CVF Conference on Computer Vision and Pattern Recognition (CVPR)}, 2021.

\bibitem[Song et~al.(2020)Song, Meng, and Ermon]{song2020denoising}
Jiaming Song, Chenlin Meng, and Stefano Ermon.
\newblock Denoising diffusion implicit models.
\newblock \emph{arXiv preprint arXiv:2010.02502}, 2020.

\bibitem[Song et~al.(2023)Song, Cao, Xu, Kang, Tang, Yuan, and Zhao]{song2023roomdreamer}
Liangchen Song, Liangliang Cao, Hongyu Xu, Kai Kang, Feng Tang, Junsong Yuan, and Yang Zhao.
\newblock {RoomDreamer}: Text-driven 3d indoor scene synthesis with coherent geometry and texture.
\newblock \emph{arXiv preprint arXiv:2305.11337}, 2023.

\bibitem[Tang et~al.(2023)Tang, Zhang, Chen, Wang, and Furukawa]{Tang2023mvdiffusion}
Shitao Tang, Fuyang Zhang, Jiacheng Chen, Peng Wang, and Yasutaka Furukawa.
\newblock {MVDiffusion}: Enabling holistic multi-view image generation with correspondence-aware diffusion.
\newblock \emph{arXiv preprint arXiv:2307.01097}, 2023.

\bibitem[Voynov et~al.(2023)Voynov, Hertz, Arar, Fruchter, and Cohen-Or]{voynov2023curved}
Andrey Voynov, Amir Hertz, Moab Arar, Shlomi Fruchter, and Daniel Cohen-Or.
\newblock Curved diffusion: A generative model with optical geometry control.
\newblock \emph{arXiv preprint arxiv:2311.17609}, 2023.

\bibitem[Wang et~al.(2023)Wang, Chen, Ling, Xie, and Song]{wang2023360}
Jionghao Wang, Ziyu Chen, Jun Ling, Rong Xie, and Li~Song.
\newblock 360-degree panorama generation from few unregistered {NFoV} images.
\newblock \emph{arXiv preprint arXiv:2308.14686}, 2023.

\bibitem[Wang et~al.(2024)Wang, Lu, Wang, Bao, Li, Su, and Zhu]{wang2024prolificdreamer}
Zhengyi Wang, Cheng Lu, Yikai Wang, Fan Bao, Chongxuan Li, Hang Su, and Jun Zhu.
\newblock {ProlificDreamer}: High-fidelity and diverse text-to-3d generation with variational score distillation.
\newblock \emph{Advances in Neural Information Processing Systems (NeurIPS)}, 36, 2024.

\bibitem[Wu et~al.(2023)Wu, Zheng, and Cham]{wu2023panodiffusion}
Tianhao Wu, Chuanxia Zheng, and Tat-Jen Cham.
\newblock {PanoDiffusion}: 360-degree panorama outpainting via diffusion.
\newblock In \emph{International Conference on Learning Representations (ICLR)}, 2023.

\bibitem[Xiao et~al.(2018)Xiao, Ehinger, Oliva, and Torralba]{xiao2012sun360}
J.~Xiao, K.~A. Ehinger, A.~Oliva, and A.~Torralba.
\newblock Recognizing scene viewpoint using panoramic place representation.
\newblock In \emph{IEEE Conference on Computer Vision and Pattern Recognition (CVPR)}, 2018.

\bibitem[Yang et~al.(2024)Yang, Dong, Ma, Hu, Liu, Cui, and Ma]{yang2024dreamspace}
Bangbang Yang, Wenqi Dong, Lin Ma, Wenbo Hu, Xiao Liu, Zhaopeng Cui, and Yuewen Ma.
\newblock {DreamSpace}: Dreaming your room space with text-driven panoramic texture propagation.
\newblock In \emph{IEEE Conference Virtual Reality and 3D User Interfaces (VR)}, 2024.

\bibitem[Zeng et~al.(2024)Zeng, Dong, Peers, Kong, Wu, and Tong]{zeng2024dilightnet}
Chong Zeng, Yue Dong, Pieter Peers, Youkang Kong, Hongzhi Wu, and Xin Tong.
\newblock {DiLightNet}: Fine-grained lighting control for diffusion-based image generation.
\newblock In \emph{ACM SIGGRAPH}, 2024.

\bibitem[Zhang et~al.(2024)Zhang, Wu, Gambardella, Huang, Phung, Ouyang, and Cai]{zhang2024taming}
Cheng Zhang, Qianyi Wu, Camilo~Cruz Gambardella, Xiaoshui Huang, Dinh Phung, Wanli Ouyang, and Jianfei Cai.
\newblock Taming stable diffusion for text to 360 $\{$$\backslash$deg$\}$ panorama image generation.
\newblock \emph{arXiv preprint arXiv:2404.07949}, 2024.

\bibitem[Zhang et~al.(2023{\natexlab{a}})Zhang, Rao, and Agrawala]{zhang2023adding}
Lvmin Zhang, Anyi Rao, and Maneesh Agrawala.
\newblock Adding conditional control to text-to-image diffusion models.
\newblock In \emph{IEEE/CVF International Conference on Computer Vision}, 2023{\natexlab{a}}.

\bibitem[Zhang et~al.(2023{\natexlab{b}})Zhang, Song, Huang, Chen, and Liu]{zhang2023diffcollage}
Qinsheng Zhang, Jiaming Song, Xun Huang, Yongxin Chen, and Ming-Yu Liu.
\newblock {DiffCollage}: Parallel generation of large content with diffusion models.
\newblock In \emph{IEEE/CVF Conference on Computer Vision and Pattern Recognition (CVPR)}, 2023{\natexlab{b}}.

\bibitem[Zhao et~al.(2024)Zhao, Srinivasan, Verbin, Park, Brualla, and Henzler]{zhao2024illuminerf}
Xiaoming Zhao, Pratul~P Srinivasan, Dor Verbin, Keunhong Park, Ricardo~Martin Brualla, and Philipp Henzler.
\newblock {IllumiNeRF}: 3d relighting without inverse rendering.
\newblock \emph{arXiv preprint arXiv:2406.06527}, 2024.

\bibitem[Zheng et~al.(2020)Zheng, Zhang, Li, Tang, Gao, and Zhou]{zheng2020structured3d}
Jia Zheng, Junfei Zhang, Jing Li, Rui Tang, Shenghua Gao, and Zihan Zhou.
\newblock Structured3d: A large photo-realistic dataset for structured 3d modeling.
\newblock In \emph{European Conference on Computer Vision (ECCV)}, 2020.

\end{thebibliography}
